\documentclass[conference]{IEEEtran}

\IEEEoverridecommandlockouts
\usepackage{cite}
\usepackage{array}
\usepackage{ragged2e}
\usepackage{amsmath,amssymb,amsfonts}
\usepackage{algorithmic}
\usepackage{graphicx}
\usepackage{tabularx}
\usepackage{textcomp}
\usepackage{textcase}
\usepackage{makecell}
\usepackage{threeparttable}
\usepackage{booktabs}
\usepackage{xcolor}
\colorlet{blue}{black}
\usepackage{bbm}
\usepackage{multirow}
\usepackage{bm}
\usepackage{siunitx}
\usepackage{caption}
\usepackage{graphicx}
\usepackage{subcaption}
\usepackage{dblfloatfix}
\usepackage{placeins}
\usepackage[hidelinks]{hyperref}
\newenvironment{revisionblue}{\color{blue}}{}

\def\BibTeX{{\rm B\kern-.05em{\sc i\kern-.025em b}\kern-.08em
    T\kern-.1667em\lower.7ex\hbox{E}\kern-.125emX}}

\usepackage{etoolbox}
\begin{document}
\setcounter{dbltopnumber}{1}

\newcommand{\jianhua}[1]{\textcolor{red}{XXX JMo: #1}}

\title{AirFM-DDA: Air-Interface Foundation Model in the Delay–Doppler–Angle Domain for AI-Native 6G} 

\author{
    Kejia Bian,
    Meixia Tao,~\IEEEmembership{Fellow,~IEEE,}
    Jianhua Mo,~\IEEEmembership{Senior Member,~IEEE,} 
    Zhiyong Chen,~\IEEEmembership{Senior Member,~IEEE,} \\
    and Leyan Chen
    \thanks{This work is supported by the National Natural Science Foundation of China (NSFC) under Grants 62125108, 62431014, and by the Science and Technology Commission Foundation of Shanghai under Grant 25DP1500100. An earlier version of this work was accepted for presentation at the IEEE ICC Workshop on Wireless Foundation Models for AI-Native 6G and Beyond~\cite{conferenceWork}.
    Kejia Bian is with the School of Information Science and Electronic Engineering, Shanghai Jiao Tong University, Shanghai, 200240, China, and also with the Shanghai Innovation Institute, Shanghai 200231, China (email: kejiabian2024@sjtu.edu.cn).
    Meixia Tao, Jianhua Mo, Zhiyong Chen, and Leyan Chen are with the School of Information Science and Electronic Engineering, Shanghai Jiao Tong University, Shanghai, 200240, China (emails: \{mxtao, mjh, zhiyongchen, alpha0827\}@sjtu.edu.cn). \textit{(Corresponding authors: Meixia Tao; Jianhua Mo.)}}
    \thanks{Data is available on-line at \url{https://github.com/SII-kejia/AirFM-DDA}.}
}
\maketitle

\begin{abstract}
The success of large foundation models is catalyzing a new paradigm for AI-native 6G network design: wireless foundation models (FMs) pre-trained on channel state information (CSI) for air-interface tasks. Existing wireless FMs, however, often operate on CSI in the spatial--temporal--frequency (STF) domain, where multipath components are superimposed and structurally entangled. This hinders the learning of a universal channel representation. Their reliance on global attention also incurs prohibitive complexity for high-dimensional CSI. In this paper, we propose AirFM-DDA, an \underline{Air}-interface \underline{F}oundation \underline{M}odel in the \underline{D}elay--\underline{D}oppler--\underline{A}ngle (DDA) domain. AirFM-DDA reparameterizes CSI into the DDA domain through a four-dimensional Fourier transform to resolve multipath components along physically meaningful axes and employs window-based attention with frame-structure-aware positional encoding. Extensive experiments show that AirFM-DDA learns representations transferable across scenarios, tasks, datasets, and antenna configurations. For channel prediction and estimation, it generalizes zero-shot to unseen cities, achieving average normalized mean-square error (NMSE) gains of 4.9--8.5 dB over the strongest baselines. With only 10\% labeled data, it achieves average gains of 12.0 percentage points in Top-1 accuracy for beam prediction and 3.4 percentage points in F1 score for line-of-sight (LoS) identification. It further transfers across simulated datasets and adapts to real-world measured data and different antenna-array geometries. Compared with global attention, window-based attention reduces training and inference costs by nearly an order of magnitude.
\end{abstract}


\section{Introduction}
\IEEEPARstart{T}{he} physical-layer air interface serves as the technological foundation of modern wireless systems. Traditionally, key air-interface tasks, such as channel estimation, channel prediction, and beam management, have relied on model-based statistical signal processing and task-specific lightweight artificial intelligence (AI) models. However, the transition toward sixth-generation (6G) networks, characterized particularly by extremely large arrays and integrated sensing and communication functionalities, drastically increases system dimensionality and scenario heterogeneity. \textcolor{blue}{Conventional mathematical frameworks struggle with high-dimensional optimization, while task-specific AI models remain constrained by their heavy dependence on data and poor generalization.} Consequently, next-generation AI-native networks require a unified paradigm that provides powerful representational capabilities and enables autonomous adaptation to dynamic environments and diverse tasks.

\textcolor{blue}{Foundation models (FMs) are generally defined as models pre-trained on broad data, typically through self-supervised learning at scale, and adaptable to a wide range of downstream tasks \cite{bommasani2021opportunities}.} In recent years, FMs have demonstrated remarkable generalization and emergent capabilities across diverse domains, most notably in natural language processing \cite{radford2019language} and computer vision \cite{kirillov2023segment}. Inspired by this success, the wireless community has begun tailoring FMs to the physical-layer air interface tasks, leveraging their powerful representation learning for robust, cross-task generalization \cite{11180834, jiang2025towards, chen2025towards, jiang2026comprehensive, 11074348, 11370176, 11449577}. Current efforts in this direction fall into two broad categories: cross-domain adapted FMs and wireless-native FMs. The former repurposes pre-trained backbones from other modalities (e.g., vision or language) through post-training or prompt-based adaptation using limited wireless data~\cite{liu2025llm4wm, zheng2025large, liu2024llm4cp, xie20252dlam, cui2025exploring, guo2025lvm4csi}. In contrast, wireless-native FMs are pre-trained from scratch on large-scale native wireless data, primarily channel state information (CSI) \cite{alikhani2024large, catak2025bert4mimo, liu2025wifo, yang2025wirelessgpt}. This native pre-training paradigm aims to learn a universal representation of physical propagation dynamics so as to enable strong generalization across diverse tasks (e.g., channel estimation, beam prediction, localization) and heterogeneous deployment scenarios.


The design of such wireless-native FMs centers on two key elements: the representation domain of the input wireless data and the network architecture. To date, CSI has become the dominant input modality, as it directly preserves the structure of the wireless propagation channel. Early works typically rely on CSI snapshots to extract underlying channel features within the spatial–frequency (SF) domain \cite{alikhani2024large, catak2025bert4mimo}. More recent approaches extend this to the full spatial–temporal–frequency (STF) domain to capture mobility-induced temporal variations, thereby enabling support for high-mobility communication scenarios \cite{liu2025wifo, yang2025wirelessgpt}. Correspondingly,  model architectures have evolved to handle these increasingly high-dimensional inputs. SF-domain models often employ a Transformer encoder paired with a lightweight task-specific head, while STF-domain models demand fully Transformer-based architectures for joint reasoning over complex, high-dimensional dependencies. A critical technical issue is that tokenizing CSI inherently disrupts its native ordering across space, frequency, and time. To mitigate this, positional encoding is essential to preserve the structural relationships embedded in the SF or STF representations.


However, existing wireless-native FMs exhibit three key limitations. First, these models inevitably suffer from the fact that the CSI observed in the SF or STF domain is inherently a superposition of multiple propagation paths. This aggregated representation obscures the individual path-level characteristics, such as path delay, Doppler shift, and angle, rendering the underlying propagation mechanisms structurally entangled and physically implicit. Second, their network architectures remain dominated by global attention mechanisms with quadratic computational complexity, thereby incurring prohibitive training and inference overhead. Third, current positional encoding schemes are largely confined to discrete grid indices, while overlooking the underlying frame structure of CSI, such as the slot duration and the subcarrier~spacing. 

In this paper, we propose AirFM-DDA, an \textbf{Air}-interface \textbf{F}oundation \textbf{M}odel in the \textbf{D}elay-\textbf{D}oppler-\textbf{A}ngle domain for physical-layer air-interface tasks. Compared with existing wireless-native FMs that operate in SF or STF domains, AirFM-DDA addresses the above three limitations through a co-design of domain representation, architecture, and positional encoding. The main features and advances of AirFM-DDA are summarized as follows:

\begin{itemize}
    \item \textbf{DDA-domain reparameterization:} AirFM-DDA reparameterizes CSI from the STF domain into the DDA domain via a four-dimensional Fourier transform. Unlike the STF representation, in which multipath components are superimposed and thus individual propagation paths become structurally entangled, the DDA domain renders these components sparse and separable along physically meaningful axes (delay, Doppler shift, and angle). This transformation reveals intrinsic local structures for efficient representation learning. With such reparameterization, diverse channel-related tasks, such as channel prediction and estimation, can be unified under a single DDA-domain learning objective, which empowers the FM to extract universal representations of wireless channels.


    \item \textbf{Locality-Aligned Attention:} AirFM-DDA introduces a window-based self-attention mechanism that is both computationally efficient and structurally aligned with DDA-domain CSI. By restricting attention to local windows, this mechanism aligns its receptive field with the locally correlated multipath structures explicitly revealed in the DDA domain. This locality-aware inductive bias not only facilitates precise feature aggregation but also circumvents the prohibitive quadratic complexity of global attention. 

    \item \textbf{Frame-Structure-Aware Positional Encoding:} AirFM-DDA explicitly injects frame-structure priors via a novel frame-structure-aware positional encoding (FS-PE). By conditioning positional embeddings on a resolution-adaptive delay--Doppler grid determined by frame parameters, FS-PE enhances representational capacity with negligible computational overhead. This structure-aware design improves the representation of heterogeneous frame configurations.

    \item \textcolor{blue}{\textbf{Comprehensive empirical validation:} Extensive experiments demonstrate that AirFM-DDA learns transferable wireless representations across scenarios, tasks, datasets, and antenna configurations. For channel prediction and estimation, the pretrained model generalizes zero-shot across unseen city scenarios, yielding average normalized mean-square error (NMSE) gains of approximately 4.9--8.5 dB over the strongest baselines. Beyond these tasks, the learned representation transfers to diverse downstream tasks, such as beam prediction and line-of-sight (LoS) identification, where AirFM-DDA achieves average gains of 12.0 percentage points in Top-1 accuracy and 3.4 percentage points in F1 score, respectively, over the strongest baselines with only 10\% labeled data. Its transferability is further validated through zero-shot evaluation on other simulated datasets, limited-data adaptation to a real-world measured dataset, and adaptation to different antenna array configurations. In addition, the window-based attention backbone improves training throughput by over an order of magnitude and accelerates inference by nearly eightfold over global attention. AirFM-DDA also exhibits consistent scalability and robustness to mobility, delay spread, and noise.}    
        
\end{itemize}

The rest of this paper is organized as follows. Section II reviews the related work. Section III presents the system model and formulates the DDA-domain CSI reconstruction problem. Section IV describes the proposed AirFM-DDA, including data preprocessing, network architecture, and reconstruction-oriented pre-training. Section V formulates the downstream tasks and presents the corresponding task-specific adaptation strategy. Section VI describes the experimental setup, and Section VII reports and analyzes the experimental results. Finally, Section VIII concludes the paper and outlines future research directions.

\textit{Notations:} $\odot$ and $\circ$ denote the Hadamard (element-wise) product and the outer product, respectively; $\circledast$ denotes the normalized one-dimensional circular convolution; $\mathbb{I}(\cdot)$ denotes the indicator function; $\operatorname{vec}(\cdot)$ denotes column-wise matrix vectorization; $\mathbf{1}_n$ denotes the length-$n$ all-one vector; $[x]_N \triangleq x \bmod N$ denotes circular indexing with period $N$; $|\mathcal{S}|$ denotes the cardinality of a set $\mathcal{S}$. $\tilde{(\cdot)}$ and $\hat{(\cdot)}$ denote masked observations and reconstructed quantities, respectively; $\mathcal{R}(\cdot)$ denotes the real-valued transformation that converts a complex-valued tensor into a real-valued representation, and $\mathcal{R}^{-1}(\cdot)$ denotes its inverse that restores the corresponding complex-valued tensor.

\section{Related Work}

FMs in the wireless domain can be broadly classified into two paradigms: cross-domain adaptation and wireless-native pre-training. This section first reviews cross-domain adapted FMs to outline their methodologies and inherent limitations. Subsequently, we systematically investigate wireless-native FMs from two key perspectives: CSI domain and network design. Through this review, we pinpoint the prevailing bottlenecks of the current native FMs. The relevant literature is summarized in~Table~\ref{tab:wireless_native_fm_comparison}.

\nocite{guler2025multi,10971878,jiang2026csi,sheng2025wireless,liu2025foundation,zhang2026hetercsi,wang2026consisformer,zhang2026adaptive,jiang2025mimo,pan2025large,wen2026icwlm,zheng2025muse}

\begin{table*}[!t]
\color{blue}
\centering
\begin{threeparttable}
\caption{Summary of related work on wireless-native FMs.}
\label{tab:wireless_native_fm_comparison}
\scriptsize
\renewcommand{\arraystretch}{1.5}
\setlength{\tabcolsep}{4pt}
\begin{tabular}{c c c c c c c c c}
\toprule
\textbf{Reference} 
& \textbf{CSI Domain} 
& \textbf{Attn. Module} 
& \makecell{\textbf{PE w/ FS}\\ \textbf{Priors}} 
& \makecell{\textbf{Training} \\ \textbf{Dataset}}
& \makecell{\textbf{Training Data} \\ \textbf{Volume}}
& \makecell{\textbf{Params} \\ (M)} 
& \makecell{\textbf{Downstream Tasks}} 
& \makecell{\textbf{Heterogeneous}\\ \textbf{CSI}} \\
\midrule
LWM \cite{alikhani2024large}        & SF       & GA            & $\times$ &  \makecell{DeepMIMO \cite{alkhateeb2019deepmimo}} & 820K & 0.6, 2.5  & \makecell{Beam Prediction\\LoS classification}   & $\times$ \\
\hline
\multirow{2}{*}{\makecell{BERT4MIMO \\ \cite{catak2025bert4mimo}}}
& \multirow{2}{*}{SF}
& \multirow{2}{*}{GA}
& \multirow{2}{*}{$\times$}
& \multirow{2}{*}{\makecell{3GPP TR 38.901 \cite{3gpp38901}}}
& \multirow{2}{*}{6K}
& \multirow{2}{*}{113.8}
& \multirow{2}{*}{CSI reconstruction}
& \multirow{2}{*}{$\checkmark$} \\
& & & & & & & & \\
\hline
\makecell{ContraWiMAE\\ \cite{guler2025multi}}   & SF       & GA            & $\times$ & \makecell{DeepMIMO \cite{alkhateeb2019deepmimo}}  & 2.25M & 0.6  &  \makecell{Beam Prediction\\LoS classification\\Channel Estimation}  & $\times$  \\
\hline
\makecell{Two-tower \\Model \cite{10971878}}         & SF       & GA            & $\times$ & \makecell{WAIR-D \cite{huangfu2022wair}}  & 2.25M & 19.1  & \makecell{CSI Feedback\\Localization\\Beam Prediction\\LoS classification }  &  $\times$ \\
\hline
\makecell{CSI-MAE \\ \cite{jiang2026csi}}         & SF       & GA            & $\times$ & \makecell{Sionna \cite{Sionna}}  & 1.45M & 85.8, 303.3  & \makecell{Channel Prediction\\CSI Feedback\\User Positioning}  &  $\checkmark$ \\
\hline
\makecell{ICWLM\\ \cite{wen2026icwlm}}           & S or ST  & Causal GA     & $\times$ & \makecell{3GPP TR 38.901 \cite{3gpp38901}}  & 240K & 10.6  & \makecell{Multi-user precoding\\Channel Prediction}  &  $\checkmark$ \\
\hline
\makecell{MUSE-FM\\ \cite{zheng2025muse} }       & S or TF  & GA            & $\times$ & \makecell{Sionna \cite{Sionna}}  & 100K & 134.4  & \makecell[c]{Channel Estimation\\MIMO Detection\\Channel Decoding\\Multi-User Precoding \\ Localization}  & $\checkmark$ \\
\hline
\makecell{WiFo\\ \cite{liu2025wifo} }            & STF      & GA            & $\times$ & \makecell{3GPP TR 38.901 \cite{3gpp38901}}  & 160K & 0.3-86.1  & Channel Prediction  & $\checkmark$  \\
\hline
\makecell{WirelessGPT\\ \cite{yang2025wirelessgpt}} & STF   & Axis-Wise GA  & $\times$ & \makecell{Traciverse \cite{yang2025wirelessgpt} \\ Sensiverse \cite{luo2023sensiverse} \\ DeepMIMO  \cite{alkhateeb2019deepmimo}}  & -- & 79.6  & \makecell[c]{Channel Estimation\\Channel Prediction\\Activity Recognition\\Scene Reconstruction \\ Object Tracking}  & $\checkmark$  \\
\hline
\makecell{FM for\\Multi-task \\Prediction\cite{sheng2025wireless}}  & STF  & Causal GA & $\times$ & \makecell{3GPP TR 38.901 \cite{3gpp38901} \\ Telecom Italia \cite{Barlacchi2015}}  & 25.69M & --  & \makecell{Channel Prediction\\Angle Prediction\\Traffic Prediction}  & $\checkmark$  \\
\hline
\makecell{WiFo-2\\ \cite{liu2025foundation}}     & STF      & GA            & $\times$ & Hybrid dataset\tnote{†}  & 567K & 8.3-100.9  & Multiple tasks\tnote{‡}  & $\checkmark$  \\
\hline
\makecell{HeterCSI\\ \cite{zhang2026hetercsi}}   & STF      & GA            & $\times$ & \makecell{3GPP TR 38.901 \cite{3gpp38901}}  & 360K & 69.9  & \makecell{Channel Reconstruction\\Channel Prediction}  & $\checkmark$  \\
\hline
\makecell{ConsisFormer\\ \cite{wang2026consisformer}} & STF & GA & $\times$ & DeepMIMO \cite{alkhateeb2019deepmimo} & 96K & 9.6 & \makecell{Channel Prediction\\Beam Prediction\\Localization\\LoS Classification} & $\times$ \\
\hline
\makecell{Adaptive 3D-RoPE\\ \cite{zhang2026adaptive}} & STF & GA & $\times$ & 3GPP TR 38.901 \cite{3gpp38901} & 144K & 71.3 & \makecell{CSI Reconstruction\\Channel Prediction} & $\checkmark$ \\
\midrule
\textbf{AirFM-DDA} & \textbf{DDA} & \textbf{Window-based Attn.} & $\checkmark$ & \makecell{DeepMIMO \cite{alkhateeb2019deepmimo}}  & 2.38M & 62.6-140.5 & \makecell{Channel Estimation\\Channel Prediction\\Beam Prediction\\LoS classification}  & $\checkmark$ \\
\bottomrule
\end{tabular}
\begin{tablenotes}
\footnotesize
\item Abbreviations: S, ST, and TF denote Spatial, Spatial--Temporal, and Temporal--Frequency, respectively. GA denotes Global Attention. FS and PE denote frame-structure and positional encoding. 
\item[†] WiFo-2 \cite{liu2025foundation} utilizes eight datasets across three categories: statistical channel modeling, real-world measurements, and ray-tracing.
\item[‡] WiFo-2 \cite{liu2025foundation} considers a broad spectrum of CSI-related tasks: CSI prediction, CSI estimation, scenario classification, beam prediction, localization, CSI feedback, angle estimation, and signal detection.

\end{tablenotes}
\end{threeparttable}
\vspace{-0.4cm}
\end{table*}

\subsection{Cross-Domain Adapted FMs}
Cross-domain adapted FMs repurpose pre-trained models from other fields, such as large language models (LLMs) and large vision models (LVMs), for the wireless domain via post-training. A common strategy in this line of work is to reshape the input wireless data into representations that better align with the structure of the adopted backbone, such as sequential token streams for LLMs or image-like tensors for LVMs \cite{liu2025llm4wm, zheng2025large, liu2024llm4cp, xie20252dlam, cui2025exploring, guo2025lvm4csi}. For instance, the authors of \cite{liu2025llm4wm} and \cite{zheng2025large} employ task-specific encoders to transform heterogeneous wireless data inputs from different physical-layer tasks into sequential embeddings, enabling shared LLM-based multi-task processing. Other studies further enrich the input modality with propagation-aware channel representations beyond CSI in the STF domain, such as delay-domain \cite{xie20252dlam, liu2024llm4cp} and angle-domain \cite{cui2025exploring} features. Meanwhile, LVM4CSI \cite{guo2025lvm4csi} transforms CSI from the SF domain to the angular-delay (AD) domain to improve compatibility with pre-trained LVMs. However, the central objective of cross-domain adapted FMs is still to fit wireless data inputs into backbones originally designed for text or images. As a result, they do not fully resolve the fundamental mismatch between wireless CSI data, which are inherently complex-valued and structured across the STF domain, and the inductive biases of pre-trained LLMs and~LVMs.

\subsection{Wireless-Native FMs}
\subsubsection{CSI Domain of Wireless-Native FMs}

In contrast to cross-domain adaptation, wireless-native FMs address this mismatch by pre-training exclusively on native wireless data. This pre-training aims to capture a transferable representation of physical propagation dynamics. Recent studies predominantly adopt CSI as their input modality, with the CSI domain progressively shifting from the SF domain to the more comprehensive STF domain. Early efforts, such as LWM \cite{alikhani2024large} and BERT4MIMO \cite{catak2025bert4mimo}, leverage masked reconstruction of SF-domain CSI to learn fundamental channel features, which can subsequently be used for multiple downstream tasks. This approach is further enhanced via contrastive learning \cite{guler2025multi, 10971878} and larger-scale dataset training \cite{jiang2026csi}. To capture the temporal dynamics arising from user mobility and environmental changes, several works \cite{liu2025wifo, sheng2025wireless, yang2025wirelessgpt, liu2025foundation, zhang2026hetercsi, wang2026consisformer, zhang2026adaptive} extend the CSI domain to the full STF domain. For instance, WiFo \cite{liu2025wifo} performs self-supervised masked pre-training across heterogeneous STF-domain CSI to capture generalized representations for diverse channel-related tasks. Nevertheless, the CSI representations used by these native FMs remain multipath-superimposed, thereby leaving the underlying propagation mechanisms structurally entangled and physically~implicit.

To alleviate this multipath superposition, preliminary studies have attempted to incorporate physical priors into the learning process. CSI-CLIP \cite{jiang2025mimo} utilizes the Fourier transform to construct auxiliary views of CSI for contrastive pre-training. Concurrently, LWLM \cite{pan2025large} learns the domain-transformation invariance from the consistency across SF and AD domains. However, these alignments operate only in the latent representation space, with the input CSI remaining in the SF domain. Consequently, the backbone is still forced to directly process multipath-superimposed CSI.

\begin{figure*}[!t]
    \centering
    \begin{subfigure}[t]{0.47\linewidth}
        \centering
        \includegraphics[width=0.90\linewidth]{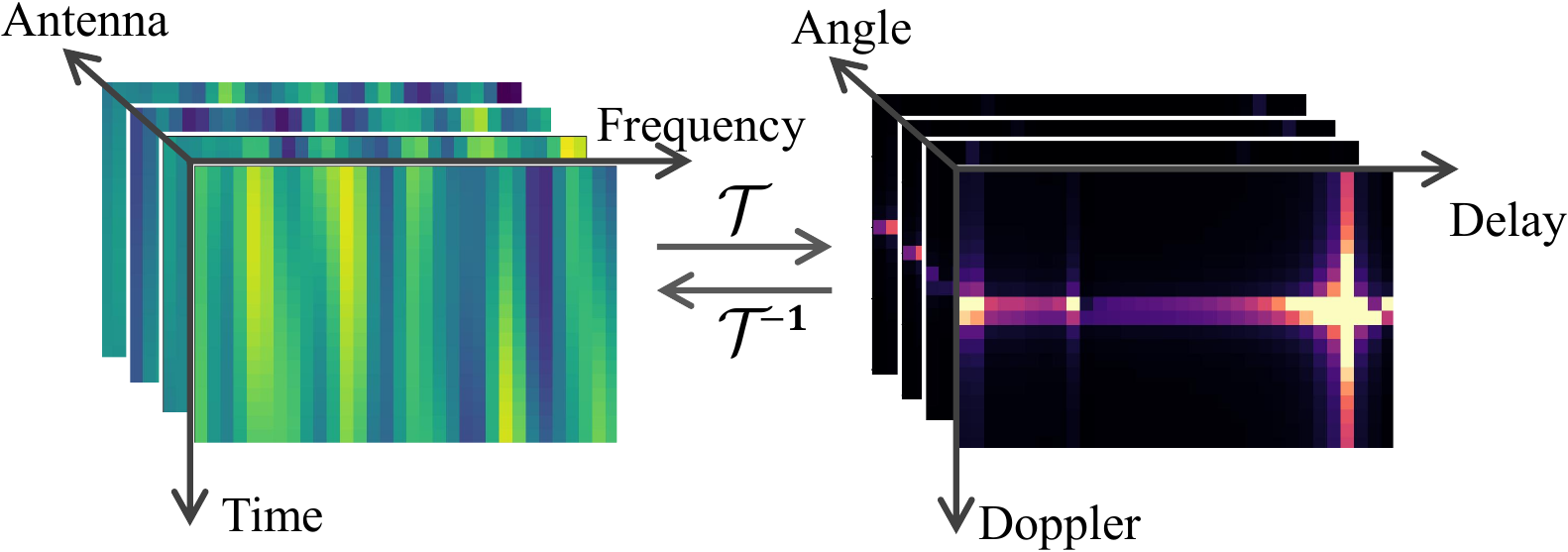}
        \caption{CSI target.}
        \label{fig:xxx_a}
    \end{subfigure}\hfill
    \begin{subfigure}[t]{0.47\linewidth}
        \centering
        \includegraphics[width=0.90\linewidth]{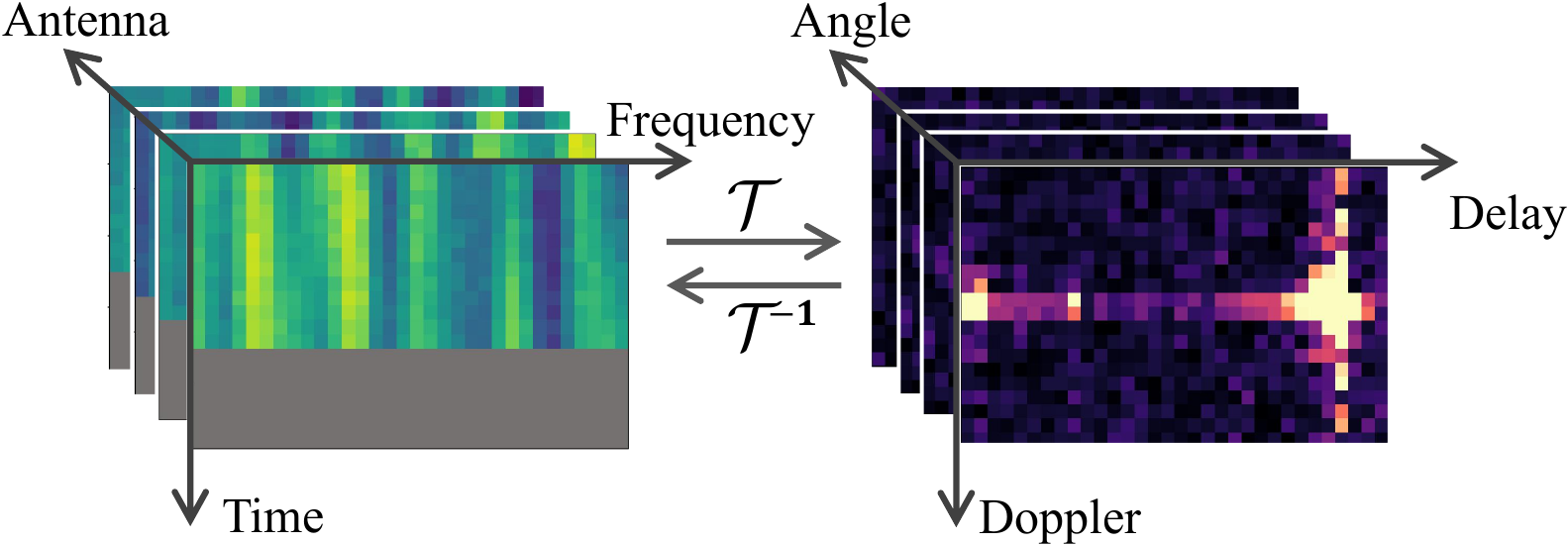}
        \caption{CSI input of TP task (observation ratio $x=0.65$).}
        \label{fig:xxx_b}
    \end{subfigure}

    \vspace{0.6em}

    \begin{subfigure}[t]{0.47\linewidth}
        \centering
        \includegraphics[width=0.90\linewidth]{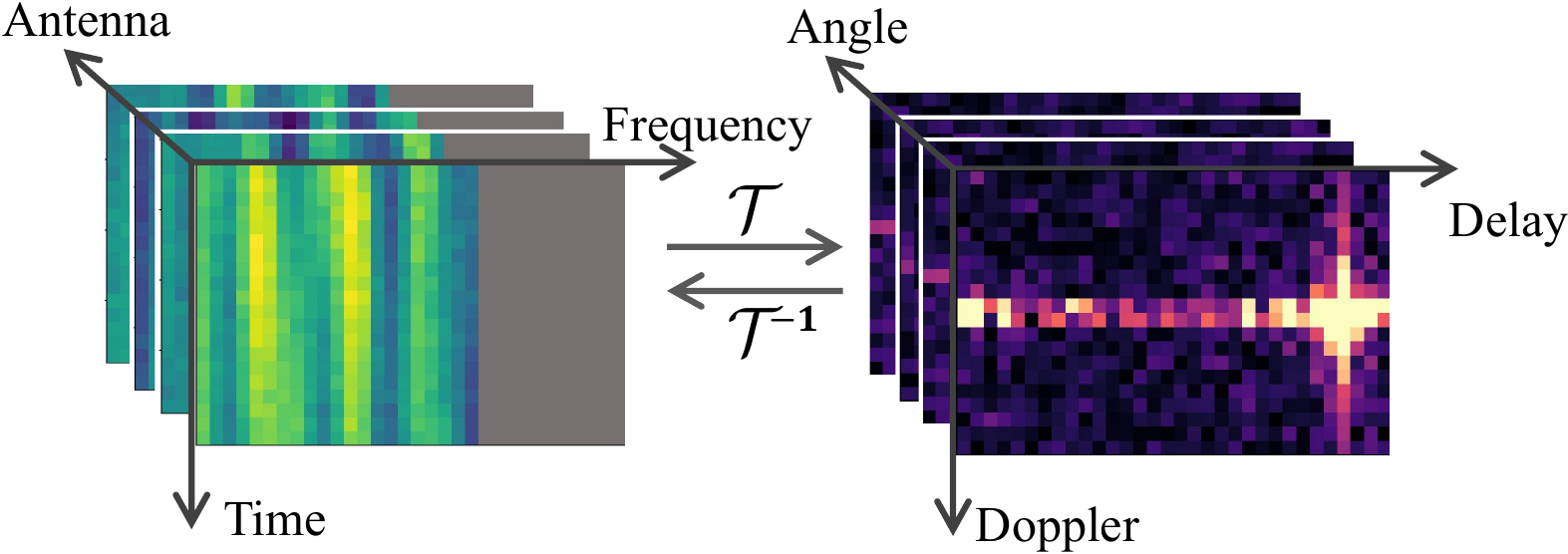}
        \caption{CSI input of FP task ($x=0.65$).}
        \label{fig:xxx_c}
    \end{subfigure}\hfill
    \begin{subfigure}[t]{0.47\linewidth}
        \centering
        \includegraphics[width=0.90\linewidth]{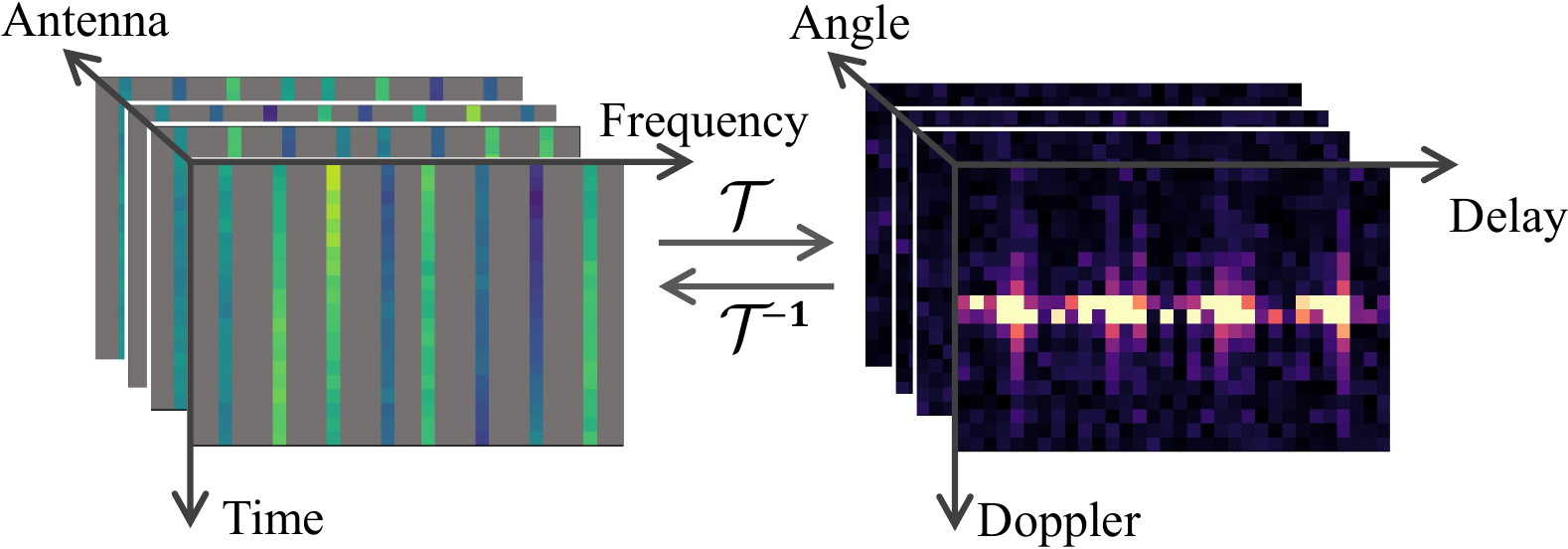}
        \caption{CSI input of CE task (pilot spacing $D=4$).}
        \label{fig:xxx_d}
    \end{subfigure}
    \caption{An illustration of the TP, FP, and CE tasks in the DDA domain. $\mathcal{T}$ denotes the Fourier transform defined in~Eq.~\eqref{eq:stf_to_dda}.}
    \label{fig:illustration diverse tasks}
\vspace{-0.35cm}
\end{figure*}

\setcounter{equation}{1}
\begin{figure*}[!b]
\hrulefill
\begin{equation}
\mathbf{H}_{\mathrm{dda}}[\nu,\tau,q_1,q_2]
=
\frac{1}{\sqrt{N_t N_f N_{\mathrm{rx}}}}
\sum_{s=0}^{N_t-1}
\sum_{k=0}^{N_f-1}
\sum_{n_1=0}^{N_{\mathrm{rx},1}-1}
\sum_{n_2=0}^{N_{\mathrm{rx},2}-1}
\mathbf{H}_{\mathrm{stf}}[s,k,n_1,n_2]\,
\exp\!\left[
-j2\pi\left(
\frac{\nu s}{N_t}
-
\frac{\tau k}{N_f}
+
\frac{q_1 n_1}{N_{\mathrm{rx},1}}
+
\frac{q_2 n_2}{N_{\mathrm{rx},2}}
\right)
\right].
\label{eq:stf_to_dda}
\end{equation}
\end{figure*}
\setcounter{equation}{0}

\subsubsection{Network Design of Wireless-Native FMs}

While the CSI domain determines the structural clarity of multipath propagation, the network design defines the inductive biases required to efficiently capture these domain-specific structures. Driven by powerful attention mechanisms, Transformers dominate the architectural design of native FMs. 
Early efforts such as LWM \cite{alikhani2024large} and BERT4MIMO \cite{catak2025bert4mimo} employ a Transformer encoder paired with a lightweight multilayer perceptron (MLP) or convolutional neural network (CNN) adaptation head. To strengthen the mapping from latent representations to channel-specific objectives, subsequent studies replace these lightweight heads with Transformer-based masked reconstruction decoders \cite{liu2025wifo, zhang2026hetercsi}. For in-context-learning-oriented models such as ICWLM \cite{wen2026icwlm}, the architecture further shifts toward a causal, autoregressive Transformer framework. Beyond these backbone-level adaptations, key Transformer components have also been refined. For example, WirelessGPT \cite{yang2025wirelessgpt} models the STF domain through tri-domain embeddings and cross-domain self-attention, while WiFo-2 \cite{liu2025foundation} replaces standard feed-forward blocks with sparse mixture-of-experts blocks to reduce inference costs. \textcolor{blue}{Complementing these efforts, ConsisFormer \cite{wang2026consisformer} improves computational efficiency through consistency-guided aggregation of neighboring CSI tokens before global self-attention.}

Since Transformers are inherently order-agnostic, positional information is typically introduced into the tokenized input. LWM \cite{alikhani2024large} and MUSE-FM \cite{zheng2025muse} employ one-dimensional positional encodings to preserve the sequential relations among serialized tokens. Extending this line, \cite{sheng2025wireless} incorporates granularity encoding to accommodate varying temporal sampling intervals. To capture the inherent STF dependencies in tokenized wireless representations, WiFo \cite{liu2025wifo} and WirelessGPT \cite{yang2025wirelessgpt} adopt absolute three-dimensional positional encodings across different physical dimensions. \textcolor{blue}{Beyond these absolute encodings, Adaptive 3D-RoPE \cite{zhang2026adaptive} employs axis-decoupled, channel-conditioned rotary frequencies to model relative dependencies across temporal, frequency, and spatial dimensions.}

Despite these advancements, most existing network designs remain constrained by two critical bottlenecks. First, they predominantly rely on global attention with quadratic complexity. This leads to prohibitive training costs, substantial inference latency, and frequent out-of-memory failures when processing high-dimensional CSI or fine-grained patch partitioning. Second, their positional encodings remain largely insensitive to the underlying frame structure of CSI representations. By parameterizing positions only through grid indices, these models overlook frame-structure parameters, thereby failing to capture the actual physical scales of CSI.

\section{System Model and Problem Formulation}

\subsection{System Model}

We consider an orthogonal frequency division multiplexing (OFDM)-based communication system consisting of single-antenna users (UEs) and a base station (BS) equipped with a uniform planar array (UPA). The CSI in the STF domain between a UE and the BS is denoted by the channel tensor, $\mathbf{H}_{\mathrm{stf}} \in \mathbb{C}^{N_t \times N_f \times N_{\mathrm{rx},1} \times N_{\mathrm{rx},2}}$, where $N_t$ and $N_f$ represent the total numbers of samples along the time and frequency axes, respectively. The parameters $N_{\mathrm{rx},1}$ and $N_{\mathrm{rx},2}$ denote the numbers of antenna elements along the horizontal and vertical dimensions of the UPA, with a total of $N_{\mathrm{rx}} = N_{\mathrm{rx},1} N_{\mathrm{rx},2}$ receive antennas.

Specifically, at the $s$-th time sample and the $k$-th frequency sample, the spatial channel response $\mathbf{H}_{\mathrm{stf}}[s,k] \in \mathbb{C}^{N_{\mathrm{rx},1} \times N_{\mathrm{rx},2}}$ is given by
\begin{equation}
\mathbf{H}_{\mathrm{stf}}[s,k]
=
\sum_{p=1}^{N_{\text{path}}}
\beta_p
e^{j2\pi (\nu_p s\Delta t - \tau_p k\Delta f)}\,
\mathbf{A}_{\mathrm{rx}}\!\left(\theta_{p},\phi_{p}\right).
\label{spatial channel response}
\end{equation}
\setcounter{equation}{2}
\par\noindent Here, $N_{\text{path}}$ denotes the number of propagation paths. For the $p$-th path, $\beta_p$ is the complex path gain, while $\nu_p$ and $\tau_p$ represent the Doppler shift and propagation delay, respectively. The parameters $\Delta t$ and $\Delta f$ denote the sampling intervals along the time and frequency axes, respectively. The matrix $\mathbf{A}_{\mathrm{rx}}(\theta_p,\phi_p)\in\mathbb{C}^{N_{\mathrm{rx},1}\times N_{\mathrm{rx},2}}$ denotes the array response matrix of the UPA corresponding to the azimuth angle of arrival (AoA) $\theta$ and the elevation AoA $\phi$.

Given the STF-domain CSI, we proceed to derive the DDA-domain expression. The DDA-domain CSI is obtained via a four-dimensional transformation of $\mathbf{H}_{\mathrm{stf}}$, which consists of a symplectic finite Fourier transform (SFFT) applied to the time-frequency grid and a two-dimensional discrete Fourier transform (DFT) applied across the spatial dimensions of the UPA. Let $\mathbf{H}_{\mathrm{dda}} \in \mathbb{C}^{N_\nu \times N_\tau \times N_{\mathrm{rx},1} \times N_{\mathrm{rx},2}}$ denote the resulting DDA-domain CSI, where $N_\nu$ and $N_\tau$ correspond to the numbers of Doppler and delay bins, respectively. \textcolor{blue}{In this work, we employ a full-sized Fourier transform with $N_{\nu}=N_t$ and $N_{\tau}=N_f$. Since no dimensional truncation is involved, the transformation is invertible and preserves all CSI information. The STF-to-DDA transformation is given in~\eqref{eq:stf_to_dda}.}

\begin{figure*}[!t]
    \captionsetup{labelfont={color=blue},textfont={color=blue}}
    \captionsetup[subfigure]{labelfont={color=blue},textfont={color=blue}}
    \centering
    \begin{minipage}{0.95\textwidth}
    \centering
    \begin{subfigure}[t]{\linewidth}
        \centering
        \includegraphics[width=\linewidth]{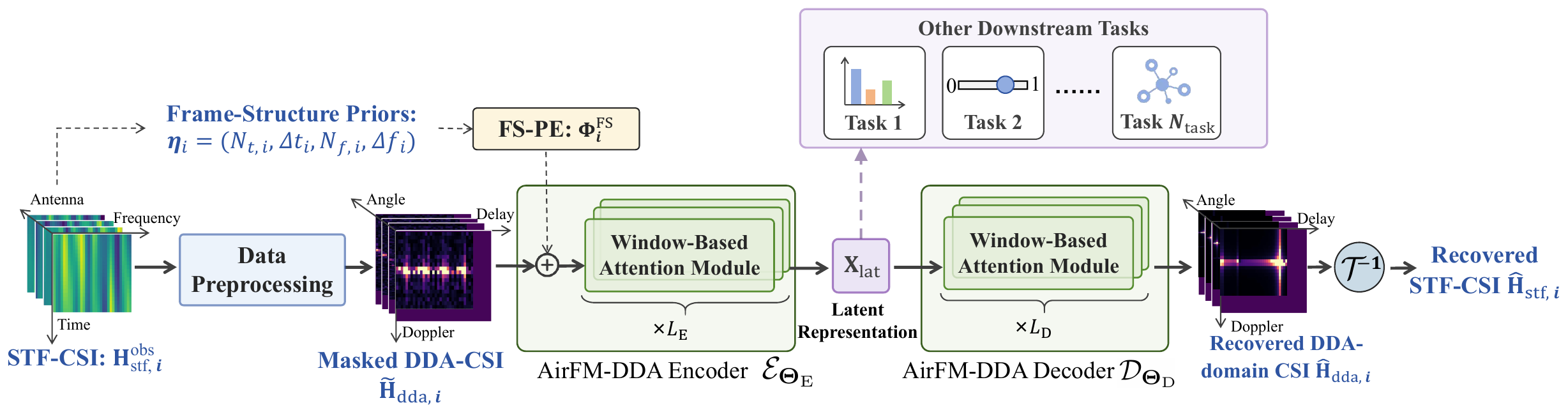}
        \caption{Overall Framework.}
        \label{fig:airfm-overall-framework}
    \end{subfigure}

    \vspace{0.5em}

    \begin{subfigure}[t]{0.49\linewidth}
        \vspace{0pt}
        \centering
        \includegraphics[height=0.31\linewidth]{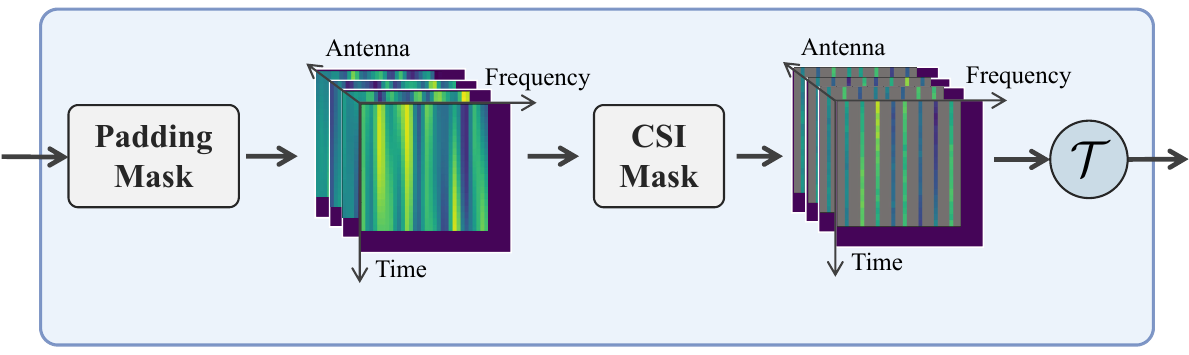}
        \caption{Data Preprocessing.}
        \label{fig:airfm-data-preprocessing}
    \end{subfigure}\hfill
    \begin{subfigure}[t]{0.49\linewidth}
        \vspace{0pt}
        \centering
        \includegraphics[height=0.31\linewidth]{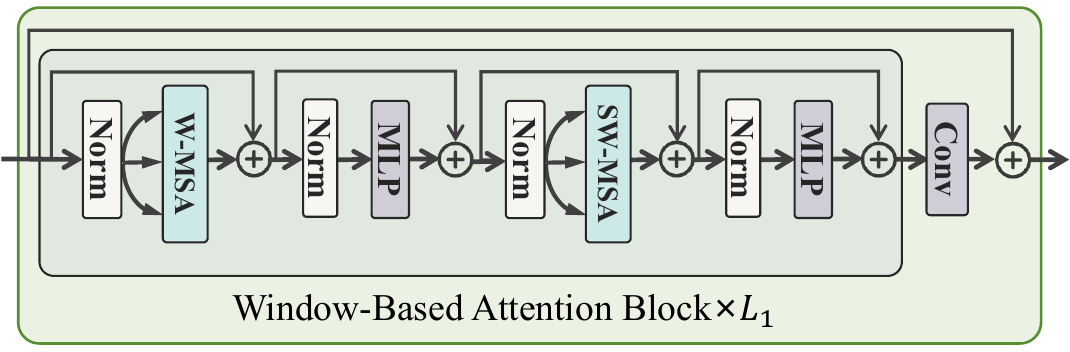}
        \caption{Window-Based Attention Module.}
        \label{fig:airfm-window-attention-module}
    \end{subfigure}
    \end{minipage}
    \caption{Overview of AirFM-DDA.
    $\mathcal{E}_{\boldsymbol{\Theta}_{\mathrm{E}}}$ and
    $\mathcal{D}_{\boldsymbol{\Theta}_{\mathrm{D}}}$ denote the shared encoder
    and reconstruction decoder, respectively.
    $\mathcal{T}$ and $\mathcal{T}^{-1}$ denote the Fourier transform defined
    in~Eq.~\eqref{eq:stf_to_dda} and its inverse, respectively.
    $\oplus$ denotes element-wise addition.}
    \label{fig:AirMIND Framework}
\vspace{-0.4cm}
\end{figure*}

\subsection{Problem Formulation}

CSI reconstruction is widely adopted in wireless-native FMs to learn generalizable CSI representations. In this subsection, we formulate CSI reconstruction from partial observations along the time–frequency dimensions in the STF domain and derive the corresponding DDA-domain formulation. Specifically, we consider three representative tasks: time-domain channel prediction (TP), frequency-domain channel prediction (FP), and channel estimation (CE).

Let $\mathbf{H}_{\mathrm{stf}}$ denote the ground-truth STF-domain CSI.
We model the observation as a binary masking operation along a single axis $d\in\{t,f\}$, formulated as

\begin{equation}
\widetilde{\mathbf{H}}_{\mathrm{stf}}
=
\mathbf{M}_d \odot \mathbf{H}_{\mathrm{stf}},
\label{eq:mask_stf}
\end{equation}
where $\mathbf{M}_d \in \{0,1\}^{N_t \times N_f \times N_{\text{rx,1}} \times N_{\text{rx,2}}}$ is defined from a one-dimensional sampling pattern
$\mathbf{m}_d \in \{0,1\}^{N_d}$ as
\begin{equation}
\mathbf{M}_d = 
\begin{cases}
\mathbf{m}_t \circ \mathbf{1}_{N_f} \circ \mathbf{1}_{N_{\mathrm{rx},1}} \circ \mathbf{1}_{N_{\mathrm{rx},2}}, & d=t, \\
\mathbf{1}_{N_t} \circ \mathbf{m}_f \circ \mathbf{1}_{N_{\mathrm{rx},1}} \circ \mathbf{1}_{N_{\mathrm{rx},2}}, & d=f.
\end{cases}
\label{eq:mask_def}
\end{equation}
Different realizations of $\mathbf{m}_d$ correspond to different reconstruction tasks.

By applying~\eqref{eq:stf_to_dda} to the masked observation in~\eqref{eq:mask_stf}, we obtain the DDA-domain expression
\begin{equation}
\widetilde{\mathbf{H}}_{\mathrm{dda}}[\nu, \tau,q_1,q_2]
=
\begin{cases}
\big(\mathbf{H}_{\mathrm{dda}}[\cdot,\tau,q_1,q_2]\circledast \mathbf{w}_t\big)[\nu], & d=t,\\[2pt]
\big(\mathbf{H}_{\mathrm{dda}}[\nu,\cdot,q_1,q_2]\circledast \mathbf{w}_f\big)[\tau], & d=f.
\end{cases}
\label{eq:mask_conv_unified}
\end{equation}
The kernels $\mathbf{w}_t$ and $\mathbf{w}_f$ are defined as the unnormalized one-dimensional DFTs of $\mathbf{m}_t$ and $\mathbf{m}_f$, respectively, with transform signs consistent with~\eqref{eq:stf_to_dda}. Based on the unified characterization in~\eqref{eq:mask_conv_unified}, we proceed to detail the specific mask definitions and their corresponding DDA-domain dual expression for each task. Notably, Fig.~\ref{fig:illustration diverse tasks} provides an intuitive illustration of the TP, FP, and CE tasks in the DDA domain.


\subsubsection{Time-Domain Prediction}
The TP task aims to extrapolate the CSI of future time samples from a contiguous history. Given an observation ratio $x\in(0,1)$, we employ a causal temporal mask vector $\mathbf{m}_t \in \{0, 1\}^{N_t}$, defined as
\begin{equation}
\mathbf{m}_t[s]=\mathbb{I}\!\left(0\le s< N_t^{\mathrm{obs}}\right), \qquad N_t^{\mathrm{obs}}=\lfloor xN_t\rfloor.
\label{eq:mask_time_prefix}
\end{equation}
The objective is to recover the unobserved suffix $\mathbf{H}_{\mathrm{stf}}[s,\cdot]$ for $s\ge N_t^{\mathrm{obs}}$. In the DDA domain, equation~\eqref{eq:mask_conv_unified} implies that this temporal truncation results in Doppler-domain spreading. Specifically, the observation $\widetilde{\mathbf{H}}_{\mathrm{dda}}$ is obtained by convolving $\mathbf{H}_{\mathrm{dda}}$ along the Doppler axis with the kernel $\mathbf{w}_t$, acting as a Dirichlet-type window that blurs the Doppler resolution.

\subsubsection{Frequency-Domain Prediction}
The FP task targets extrapolation across subcarriers from a partially observed band. Given an observation ratio $x\in(0,1)$, we observe a contiguous subset of subcarriers using mask vector $\mathbf{m}_f \in \{0, 1\}^{N_f}$, defined as
\begin{equation}
\mathbf{m}_f[k]=\mathbb{I}\!\left(0\le k< N_f^{\mathrm{obs}}\right), \qquad N_f^{\mathrm{obs}}=\lfloor xN_f\rfloor.
\label{eq:mask_freq_prefix}
\end{equation}
The objective is to recover the missing subcarriers $\mathbf{H}_{\mathrm{stf}}[\cdot,k,\cdot]$ for $k\ge N_f^{\mathrm{obs}}$. By~\eqref{eq:mask_conv_unified}, truncation along the frequency axis corresponds to circular convolution along the delay axis in the DDA domain. Consequently, the kernel $\mathbf{w}_f$ smears the sparse delay components of $\mathbf{H}_{\mathrm{dda}}$ and induces delay spreading.

\subsubsection{Frequency-Domain Channel Estimation}
The CE task considers recovering the full CSI from discrete pilots. Given a pilot spacing $D$ that divides $N_f$, we employ an equispaced (comb-type) sampling pattern with the mask vector $\mathbf{m}_f \in \{0, 1\}^{N_f}$, defined as
\begin{equation}
\mathbf{m}_f[k]=\sum_{r=0}^{N_f/D-1}\delta[k-rD],
\label{eq:mask_comb}
\end{equation}
where $\delta[\cdot]$ is the Kronecker delta. In contrast to the contiguous masking adopted for prediction, the periodic sampling in~\eqref{eq:mask_comb} yields a comb-shaped response in the delay domain, denoted~as
\begin{equation}
\mathbf{w}_f[\tau]=\frac{N_f}{D}\sum_{s=0}^{D-1}\delta\!\left[\tau-s\frac{N_f}{D}\right].
\label{eq:delay_comb}
\end{equation}
{The comb-shaped delay-domain response in~\eqref{eq:delay_comb} indicates that comb-type pilots produce structured periodic aliasing. Accordingly, the CE task can be viewed as a structured de-aliasing problem.}

In summary, although the aforementioned tasks adopt distinct sampling patterns defined by $\mathbf{M}_d$, they share the common objective of recovering the complete CSI from partial observations. Consequently, we formulate channel reconstruction as a unified learning problem within the DDA domain. Let $\mathcal{F}_{\boldsymbol{\Theta}}$ denote the unified reconstruction network parameterized by $\boldsymbol{\Theta}$. The network takes the real-valued masked DDA expression $\mathcal{R}\left(\tilde{\mathbf{H}}_{\mathrm{dda}}\right)$ as input. The reconstructed STF-domain CSI is obtained as
\begin{equation}
\widehat{\mathbf{H}}_{\mathrm{stf}}
=
\mathcal{T}^{-1}\!\left(
\mathcal{R}^{-1}\!\left\{
\mathcal{F}_{\boldsymbol{\Theta}}\!\left({\mathcal{R}\left(\tilde{\mathbf{H}}_{\mathrm{dda}}\right)}\right)
\right\}
\right),
\label{eq:recon_process}
\end{equation}
where $\mathcal{T}^{-1}(\cdot)$ denotes the inverse operator of the transform defined in~\eqref{eq:stf_to_dda}. The learning objective is to identify the optimal parameters $\boldsymbol{\Theta}^\star$ that minimize the expected reconstruction error over the joint distribution of channel realizations and task-specific masking patterns. This optimization problem is formulated as
\begin{equation}
\boldsymbol{\Theta}^\star
=
\operatorname*{arg\,min}_{\boldsymbol{\Theta}}\,
\mathbb{E}_{\mathbf{H}_{\mathrm{stf}}, \mathbf{M}_d}
\!\left[
\mathcal{L}\!\left(
\mathbf{H}_{\mathrm{stf}},
\widehat{\mathbf{H}}_{\mathrm{stf}}
\right)
\right],
\label{eq:optimization_goal_general}
\end{equation}
where $\mathcal{L}(\cdot,\cdot)$ denotes a reconstruction loss function.

\section{Proposed AirFM-DDA}
In this section, we instantiate the unified DDA-domain reconstruction mapping $\mathcal{F}_{\boldsymbol{\Theta}}$ as AirFM-DDA and present its complete reconstruction-oriented pre-training pipeline, encompassing CSI preprocessing, network architecture, and training strategy. As illustrated in Fig.~\ref{fig:AirMIND Framework}, AirFM-DDA first maps the preprocessed DDA-domain CSI into a shared latent representation through a common encoder $\mathcal{E}_{\boldsymbol{\Theta}_{\mathrm{E}}}$ and then recovers the target CSI through a reconstruction decoder $\mathcal{D}_{\boldsymbol{\Theta}_{\mathrm{D}}}$. The resulting shared representation also provides a common interface for the downstream-task adaptation introduced in Section V.

\subsection{Data Preprocessing}
As a cornerstone of AirFM-DDA, the data preprocessing stage serves a dual purpose: (i) aligning CSI samples with heterogeneous dimensions induced by different frame structures, and (ii) transforming the CSI from the multipath-superimposed STF domain into the DDA domain. The preprocessing pipeline is summarized below.

We consider a dataset $\mathcal{S}$ consisting of $|\mathcal{S}|$ paired samples, denoted as 
\begin{equation}
    \mathcal{S}=\left\{ (\mathbf{H}_{\mathrm{stf}, i}^{\mathrm{obs}}, \mathbf{H}_{\mathrm{stf}, i}^{\mathrm{gt}}) \right\}_{i=1}^{|\mathcal{S}|}, 
    \label{eq: Training Dataset}
\end{equation}
where, for the $i$-th sample, $\mathbf{H}_{\mathrm{stf}, i}^{\mathrm{obs}}$ represents the noisy observation of the full CSI, and $\mathbf{H}_{\mathrm{stf}, i}^{\mathrm{gt}}$ denotes the corresponding ground-truth STF-domain CSI. Each sample pair is associated with a frame structure $\boldsymbol{\eta}_i=(N_{t,i}, \Delta t_i, N_{f,i}, \Delta f_i)$, where $N_{t,i}$ and $N_{f,i}$ specify the temporal and frequency dimensions, respectively, while $\Delta t_i$ and $\Delta f_i$ denote the corresponding sampling intervals. Let $N_{t,\max}= \max_i N_{t,i}$ and $N_{f,\max}= \max_i N_{f,i}$ denote the global maximum temporal and frequency dimensions across the dataset, respectively.

First, we employ a mask-based padding mechanism that maps each sample to a uniform size while preserving its effective region. For the $i$-th sample, we define a binary validity mask $\mathbf{V}_{i} \in \{0,1\}^{N_{t,\max}\times N_{f,\max}}$ as
\begin{equation}
\mathbf{V}_{i}[s,k] = \mathbb{I}\!\left(0 \le s < N_{t,i}\right) \cdot \mathbb{I}\!\left(0 \le k < N_{f,i}\right).
\end{equation}
Based on this definition, the padded tensor $\mathbf{H}_{\mathrm{pad},i}\in\mathbb{C}^{N_{t,\max}\times N_{f,\max}\times N_{\mathrm{rx},1}\times N_{\mathrm{rx},2}}$ is constructed via the element-wise operation $\mathbf{H}_{\mathrm{pad},i}[s,k,\cdot] = \mathbf{V}_{i}[s,k] \cdot \mathbf{H}_{\mathrm{stf},i}^{\mathrm{obs}}[s,k,\cdot]$. All subsequent operations are governed by $\mathbf{V}_{i}$ to ensure that padded elements remain excluded from feature aggregation and loss computation. For brevity, the following derivation focuses on the valid, unpadded tensor $\mathbf{H}_{\mathrm{stf},i}^{\mathrm{obs}}$ at its native size.

Subsequently, AirFM-DDA employs the mask $\mathbf{M}_d$ defined in~\eqref{eq:mask_def}, termed the CSI mask. This mask is intrinsically linked to the target task. Specifically, we determine the masking axis based on the task type and configure the sampling parameters, such as the observation ratio $x$ or the interval $D$. The application of this mask to the noisy observation $\mathbf{H}_{\mathrm{stf}, i}^{\mathrm{obs}}$ yields the partially observed tensor, denoted as $\tilde{\mathbf{H}}_{\mathrm{stf}, i}^{\mathrm{obs}}$.

Finally, $\tilde{\mathbf{H}}_{\mathrm{stf}, i}^{\mathrm{obs}}$ is converted into the DDA domain via the Fourier transform in~\eqref{eq:stf_to_dda}, yielding the features $\tilde{\mathbf{H}}_{\mathrm{dda}, i} = \mathcal{T}\big(\tilde{\mathbf{H}}_{\mathrm{stf}, i}^{\mathrm{obs}}\big) \in \mathbb{C}^{N_{\nu,i} \times N_{\tau,i} \times N_{\text{rx},1} \times N_{\text{rx},2}}$. Before feeding this tensor into the backbone network, its two-dimensional angle dimensions are flattened into a unified angle axis $N_{\text{rx}}$, followed by the concatenation of its real and imaginary components along this axis to yield the real-valued input $\mathbf{H}_{\mathrm{in}, i} \in \mathbb{R}^{N_{\nu,i} \times N_{\tau,i} \times 2N_{\text{rx}}}$.

\subsection{Network Architecture}
\subsubsection{Window-Based Attention Module}
In the conventional STF domain, global-attention-based backbones have been widely adopted for wireless representation learning. However, directly applying them to the DDA domain presents critical challenges in both computational feasibility and structural~alignment.

Given the input DDA-domain CSI features partitioned along the Doppler-delay dimensions into non-overlapping patches of size $P \times P$, we assume that both $N_\nu$ and $N_\tau$ are divisible by $P$. The flattened sequence then contains $L_{\text{seq}} = {N_\nu N_\tau}/{P^2}$ patches, which are projected into an embedding sequence $\mathbf{X}_{0, i}\in \mathbb{R}^{L_{\text{seq}} \times C}$, where $C$ denotes the embedding dimension. Standard multi-head self-attention (MSA) exhaustively evaluates pairwise interactions across all $L_{\text{seq}}$ patches, which inherently dictates a prohibitive quadratic computational complexity of $\mathcal{O}(L_{\text{seq}}^2 C)$. Furthermore, as derived in Section III-B, STF-domain masking translates to a global convolutional blurring or aliasing in the DDA domain. This fundamental property precludes the use of patch-dropping strategies\footnote{The authors in \cite{liu2025wifo, liu2025foundation, zhang2026hetercsi} adopt a masked autoencoder-style pre-training paradigm. By removing the masked CSI patches at the encoder input, they shorten the effective token sequence, thereby partially alleviating the prohibitive quadratic complexity induced by global attention.}. Consequently, the network is forced to process the complete, high-dimensional DDA grid, which makes the $\mathcal{O}(L_{\text{seq}}^2 C)$ overhead computationally impractical at scale. Beyond efficiency, global attention is misaligned with DDA-domain physics: multipath components are locally clustered rather than globally correlated.

Inspired by the Swin Transformer architecture~\cite{liu2021swin}, we introduce a window-based attention mechanism to address these issues. Specifically, the DDA grid is partitioned into non-overlapping local windows, each containing \(M = P_{\text{w}}\times P_{\text{w}}\) patches. Consider a single attention head with dimension \(d_h = C/h\), where \(h\) denotes the number of heads. The window-based multi-head self-attention (W-MSA) is then performed independently within each window and is formulated as
\begin{equation}
\text{Attention}_{\text{W}}(Q_{\text{w}},K_{\text{w}},V_{\text{w}})
=\text{softmax}\!\left(\frac{Q_{\text{w}}K_{\text{w}}^{\top}}{\sqrt{d_h}}+B_{\mathrm{rel}}\right)V_{\text{w}},
\label{W-MSA}
\end{equation}
where \(Q_{\text{w}},K_{\text{w}},V_{\text{w}}\in\mathbb{R}^{M\times d_h}\) are the query, key, and value projections of the DDA patch tokens within a window, and \(B_{\mathrm{rel}}\in\mathbb{R}^{M\times M}\) is the relative position bias. Restricting attention to \(M\) patches per window reduces the attention complexity to \(\mathcal{O}(L_{\text{seq}}MC)\), which is linear in \(L_{\text{seq}}\) for fixed \(M\). The block also includes linear projections, an MLP, and convolutions. Crucially, the local receptive field matches the clustered DDA-domain multipath structure, enabling targeted feature aggregation.

\begin{figure}[t]
    \centering
    \includegraphics[width=0.8\linewidth]{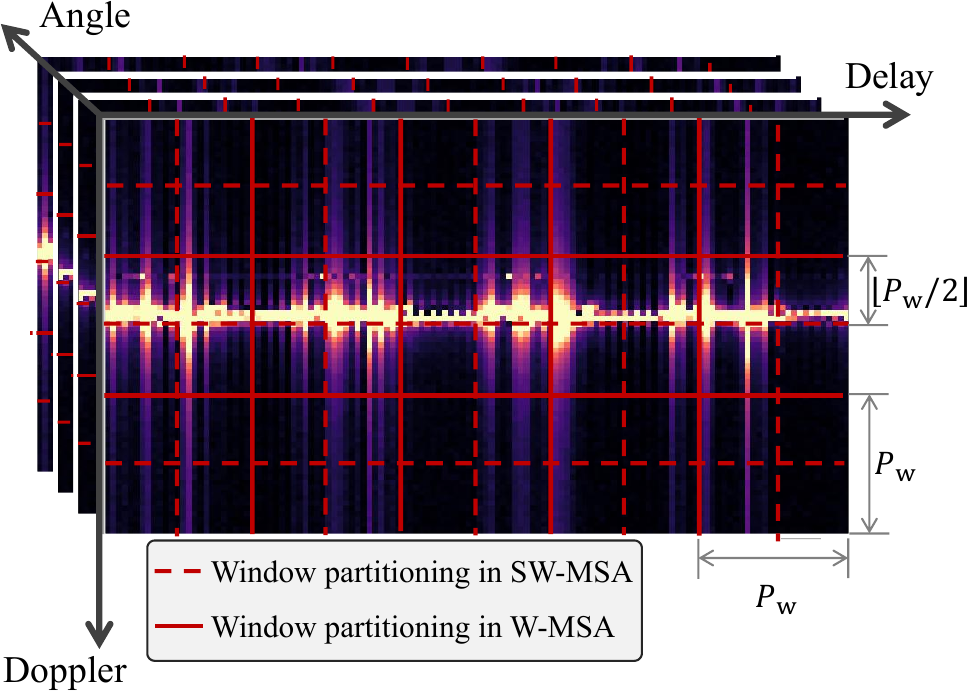}
    \caption{Window partitioning in SW-MSA and W-MSA for DDA-domain CSI.}
    \label{fig:Window Partitioning}
\vspace{-0.6cm}
\end{figure}

While W-MSA ensures efficiency and local structural alignment, isolated windows disrupt DDA-domain continuity, as multipath clusters and delay-Doppler spreads may cross fixed window boundaries. Cross-window interactions are therefore needed to preserve multipath continuity. To achieve this, we adopt shifted window-based multi-head self-attention (SW-MSA), as shown in Fig.~\ref{fig:Window Partitioning}. By shifting the windows by $\lfloor P_\text{w}/2 \rfloor$ patches along the Doppler and delay axes, SW-MSA bridges adjacent W-MSA windows and applies the attention in \eqref{W-MSA} to propagate information across the DDA grid.

To integrate localized and cross-window attention, we construct a generic window-based attention module following the residual grouping strategy of SwinIR \cite{liang2021swinir}. Each module comprises $L_1$ structurally identical attention blocks and a module-level residual connection. Let $\mathcal{G}_{m,\ell}$ denote the $\ell$-th block of the $m$-th module, and let $\mathcal{M}_m$ denote the overall mapping of that module, where $\ell=1,\ldots,L_1$ and $m$ serves as a generic module index. Given the input $\mathbf{X}_{m-1}$ to the $m$-th module, with the aforementioned embedded sequence $\mathbf{X}_{0,i}$ abbreviated as $\mathbf{X}_0$, we define $\mathbf{Z}_{m,0}=\mathbf{X}_{m-1}$ and recursively compute
\begin{equation}
\mathbf{Z}_{m,\ell}=\mathcal{G}_{m,\ell}\!\left(\mathbf{Z}_{m,\ell-1}\right), \quad \ell=1,\ldots,L_1.
\end{equation}
Within each block, feature extraction is first performed by W-MSA to capture localized multipath features, and is then followed by SW-MSA to enable cross-window interactions. After each attention operation, an MLP together with residual connections is applied to enhance representation capacity and stabilize training. Based on the stacked block outputs, the $m$-th window-based attention module is formulated as
\begin{equation}
\mathbf{X}_m=\mathcal{M}_m(\mathbf{X}_{m-1})=\mathbf{X}_{m-1}+\mathrm{Conv}_m\!\left(\mathbf{Z}_{m,L_1}\right),
\label{eq:window_module_mapping}
\end{equation}
where $\mathrm{Conv}_m(\cdot)$ denotes the convolutional layer applied to the output of the last block in the $m$-th module.
It is worth noting that throughout these operations, the feature resolution is strictly maintained at $N_\nu \times N_\tau$. This design ensures the preservation of the fine-grained multipath structures encoded in the DDA domain.

\subsubsection{FS-PE}
To explicitly inject frame-structure priors, we propose the FS-PE, which encodes the frame parameters $\boldsymbol{\eta}_i=(N_{t,i}, \Delta t_i, N_{f,i}, \Delta f_i)$ into the positional embeddings. Unlike standard positional embeddings that rely solely on integer indices, FS-PE constructs continuous coordinates based on the physical resolutions of each sample.
First, for the $i$-th sample, the delay resolution $r_{\tau,i}$ and the Doppler resolution $r_{\nu,i}$ are computed as
\begin{equation}
    r_{\tau,i} = \frac{1}{N_{f,i}\Delta f_i}, \quad
    r_{\nu,i} = \frac{1}{ N_{t,i}\Delta t_i}.
    \label{eq:resolutions}
\end{equation}
To ensure scale consistency across different frame structures, these resolutions are normalized by pre-defined global reference resolutions $(r^{\mathrm{ref}}_\tau, r^{\mathrm{ref}}_\nu)$, yielding the sample-specific coordinate step sizes, denoted as
\begin{equation}
    s_{\tau,i}= \frac{r_{\tau,i}}{r^{\mathrm{ref}}_\tau}, \quad s_{\nu,i}= \frac{r_{\nu,i}}{r^{\mathrm{ref}}_\nu}.
    \label{eq:step_sizes}
\end{equation}
Next, we map these step sizes to the 2D patch grid in the Doppler-delay plane. Assuming the input is divided into patches with grid indices $n_\nu\in\{0,\dots,N_{\nu,i}/P-1\}$ and $n_\tau\in\{0,\dots,N_{\tau,i}/P-1\}$, each patch token covers $P$ bins along both axes. Consequently, the effective step sizes on the patch grid become $P s_{\nu,i}$ and $P s_{\tau,i}$. The frame-structure-conditioned continuous coordinates for a patch at $(n_\nu, n_\tau)$ are thus defined as
\begin{equation}
    u_i[n_\nu]=n_\nu P s_{\nu,i}, \qquad v_i[n_\tau]=n_\tau P s_{\tau,i}.
\end{equation}
Finally, we project these continuous coordinates into a high-dimensional feature space. Using $L_{\text{pos}}=C/4$ frequency pairs, the sinusoidal sub-vectors for the Doppler and delay axes are formulated as
\begin{equation}
    \begin{aligned}
        \bm{\psi}^{\nu}_i(n_\nu)
        &=
        \big[\sin(u_i[n_\nu]\omega_\ell),\, \cos(u_i[n_\nu]\omega_\ell)\big]_{\ell=0}^{L_{\text{pos}}-1},\\
        \bm{\psi}^{\tau}_i(n_\tau)
        &=
        \big[\sin(v_i[n_\tau]\omega_\ell),\, \cos(v_i[n_\tau]\omega_\ell)\big]_{\ell=0}^{L_{\text{pos}}-1},
    \end{aligned}
    \label{eq:fs_pe_axis}
\end{equation}
where $\omega_\ell = \sigma^{-\ell/L_{\text{pos}}}$ and $\sigma$ defaults to $10000$. The overall FS-PE tensor $\bm{\Phi}^{\mathrm{FS}}_{i} \in \mathbb{R}^{\frac{N_{\nu, i}}{P} \times \frac{N_{\tau, i}}{P} \times C}$ is then formed by concatenating these sub-vectors along the feature channel dimension at each grid index, which is formulated as
\begin{equation}
    \bm{\Phi}^{\mathrm{FS}}_{i}[n_\nu, n_\tau, \cdot]
    =
    [\bm{\psi}^{\nu}_i(n_\nu) ; \bm{\psi}^{\tau}_i(n_\tau)].
    \label{eq:fs_pe}
\end{equation}

To construct the end-to-end forward pass of AirFM-DDA, the aforementioned window-based attention modules and FS-PE are integrated into a unified reconstruction network $\mathcal{F}_{\boldsymbol{\Theta}}$, which comprises a shared encoder $\mathcal{E}_{\boldsymbol{\Theta}_{\mathrm{E}}}$ and a reconstruction decoder $\mathcal{D}_{\boldsymbol{\Theta}_{\mathrm{D}}}$. The forward pass initiates with a convolutional embedding layer $f_{\mathrm{emb}}(\cdot)$, which projects the preprocessed real-valued DDA-domain CSI $\mathbf{H}_{\mathrm{in},i}$ into a $C$-dimensional feature space. Subsequently, the frame-structure priors are injected through element-wise addition with the FS-PE tensor, yielding the initialized encoder feature
\begin{equation}
\mathbf{X}^{\mathrm{E}}_{0,i}
=
f_{\mathrm{emb}}\!\left(\mathbf{H}_{\mathrm{in},i}\right)
+
\bm{\Phi}^{\mathrm{FS}}_{i}.
\label{eq:encoder_initialization}
\end{equation}

To distinguish their roles in the two network components, $\mathcal{M}^{\mathrm{E}}_{m}$ and $\mathcal{M}^{\mathrm{D}}_{m}$ denote the $m$-th window-based attention modules in the shared encoder and reconstruction decoder, respectively. Both follow the generic module mapping $\mathcal{M}_{m}$ defined in~\eqref{eq:window_module_mapping}, but employ separate learnable parameters. Starting from this initialization, the feature is successively refined by $L_{\mathrm{E}}$ cascaded window-based attention modules, producing the shared latent representation
\begin{equation}
\begin{aligned}
\mathbf{X}_{\mathrm{lat},i}
&=
\mathcal{E}_{\boldsymbol{\Theta}_{\mathrm{E}}}
\!\left(\mathbf{H}_{\mathrm{in},i};\bm{\Phi}^{\mathrm{FS}}_{i}\right)\\
&=
\left(
\mathcal{M}^{\mathrm{E}}_{L_{\mathrm{E}}}
\circ\cdots\circ
\mathcal{M}^{\mathrm{E}}_{1}
\right)
\!\left(\mathbf{X}^{\mathrm{E}}_{0,i}\right),
\end{aligned}
\label{eq:shared_encoder_output}
\end{equation}
where $\mathbf{X}_{\mathrm{lat},i}\in\mathbb{R}^{L_{\mathrm{seq}}\times C}$ serves as a transferable representation shared by CSI reconstruction and downstream tasks.

For CSI reconstruction, this representation is further processed by the decoder, which consists of $L_{\mathrm{D}}$ additional window-based attention modules followed by a convolutional projection layer $f_{\mathrm{proj}}(\cdot)$. After mapping the real-valued decoder output back to its complex form, the recovered DDA-domain CSI is expressed as
\begin{equation}
\begin{aligned}
\widehat{\mathbf{H}}_{\mathrm{dda},i}
&=
\mathcal{R}^{-1}\!\left\{
\mathcal{D}_{\boldsymbol{\Theta}_{\mathrm{D}}}
\!\left(\mathbf{X}_{\mathrm{lat},i}\right)
\right\}\\
&=
\mathcal{R}^{-1}\!\Bigg\{
f_{\mathrm{proj}}
\!\left[
\left(
\mathcal{M}^{\mathrm{D}}_{L_{\mathrm{D}}}
\circ\cdots\circ
\mathcal{M}^{\mathrm{D}}_{1}
\right)
\!\left(\mathbf{X}_{\mathrm{lat},i}\right)
\right]
\Bigg\}.
\end{aligned}
\label{eq:reconstruction_decoder}
\end{equation}

Accordingly, the unified reconstruction network introduced in Section~III-B is formulated as
\begin{equation}
\mathcal{F}_{\boldsymbol{\Theta}}
=
\mathcal{D}_{\boldsymbol{\Theta}_{\mathrm{D}}}
\circ
\mathcal{E}_{\boldsymbol{\Theta}_{\mathrm{E}}},
\qquad
\boldsymbol{\Theta}
=
\{\boldsymbol{\Theta}_{\mathrm{E}},\boldsymbol{\Theta}_{\mathrm{D}}\}.
\label{eq:encoder_decoder_composition}
\end{equation}

Finally, outside the network forward pass, the recovered DDA-domain CSI is transformed back to the STF domain through the inverse operator of \eqref{eq:stf_to_dda}, yielding the final reconstruction $\widehat{\mathbf{H}}_{\mathrm{stf},i}=\mathcal{T}^{-1}(\widehat{\mathbf{H}}_{\mathrm{dda},i})$. In contrast, the downstream tasks introduced in Section V directly reuse $\mathbf{X}_{\mathrm{lat},i}$ and replace the reconstruction decoder with lightweight task-specific adaptation heads.

\begin{revisionblue}
\subsection{Heterogeneity-Aware Progressive Training Strategy}

The reconstruction-oriented pre-training of AirFM-DDA involves two forms of heterogeneity. First, CSI samples may follow different frame structures and therefore have different temporal and frequency dimensions as well as different physical resolutions. Second, TP, FP, and CE correspond to distinct observation patterns and reconstruction difficulties. Training these configurations separately may cause the learned representation to become biased toward a particular task or frame structure, whereas exposing a randomly initialized model to all difficult configurations from the beginning may result in unstable or slow convergence. We therefore develop a heterogeneity-aware and progressive training strategy.

The first design is task-consistent heterogeneous training. Unlike conventional masked pre-training that randomly removes patches from the model input, AirFM-DDA constructs the corruption according to the physical observation pattern of each reconstruction task. For every training sample, a task is independently and uniformly selected from $\{\mathrm{TP},\mathrm{FP},\mathrm{CE}\}$, and its task-specific masking parameter is sampled from the predefined range associated with the current training stage. The corresponding STF-domain mask is then generated and applied to the CSI observation. The partially observed CSI is subsequently transformed into the DDA domain. As derived in Section~III, the STF-domain masks induce convolutional spreading or structured aliasing over the DDA grid. AirFM-DDA therefore retains the complete DDA grid and learns to reverse these physically meaningful degradations rather than reconstructing randomly removed DDA patches.

To jointly learn from heterogeneous tasks and frame structures, each mini-batch aggregates samples with different original CSI dimensions and task configurations. For each sample, the frame-dependent FS-PE is generated from its physical sampling parameters, while the validity mask excludes padded entries from feature aggregation and loss computation. The shared encoder and reconstruction decoder are jointly optimized using the average per-sample NMSE,
\begin{equation}
\mathcal{L}
=
\frac{1}{\left|\mathcal{B}_b\right|}
\sum_{i\in\mathcal{B}_b}
\frac{
\left\|
\mathbf{V}_i \odot
\left(
\mathbf{H}^{\mathrm{gt}}_{\mathrm{stf},i}
-
\widehat{\mathbf{H}}_{\mathrm{stf},i}
\right)
\right\|_F^2
}{
\left\|
\mathbf{V}_i \odot
\mathbf{H}^{\mathrm{gt}}_{\mathrm{stf},i}
\right\|_F^2
}.
\end{equation}
Here, $\mathbf{V}_i$ is broadcast over the antenna dimensions. Across training, this sampling exposes the model to diverse mixtures of frame structures, reconstruction objectives, and observation conditions.

The second design adopts a two-stage difficulty curriculum, consistent with the broader use of staged optimization in learned signal-recovery networks \cite{zhu2021deep}. Inspired by curriculum learning~\cite{9392296}, Stage~1 retains the complete set of reconstruction tasks but trains the model using a limited subset of the pre-training data and relatively mild observation configurations. This warm-up stage allows the encoder and decoder to first establish a stable representation of the common delay, Doppler, and angular structures. Stage~2 is initialized from the Stage-1 model and expands the training to the complete pre-training dataset with a broader range of more challenging observation configurations. Both stages jointly optimize the shared encoder and reconstruction decoder; no network component is frozen during reconstruction-oriented pre-training. The concrete task-parameter distributions, data sizes, optimization schedules, and training durations for the two stages are provided in Section~VI-B.2 and summarized in Table~\ref{table: network optimization}.
\end{revisionblue}

\begin{revisionblue}

\section{AirFM-DDA-Enabled Downstream Tasks}
\label{sec:downstream_tasks}

Through the reconstruction-oriented pre-training described in Section IV, the shared encoder learns a transferable representation of the delay, Doppler, and angular structures of wireless channel propagation. Beyond CSI reconstruction, this representation provides a common basis for adapting AirFM-DDA to diverse downstream air-interface tasks. In this work, we focus on two downstream tasks: sub-6 GHz-aided millimeter-wave (mmWave) beam prediction (BP) and CSI-based LoS identification. Unlike CSI reconstruction, both tasks yield discrete task outputs rather than high-dimensional channel tensors. This section first formulates their system models and learning objectives, and then presents the task-specific adaptation heads and training strategy.

\subsection{System Models and Task Formulation}

Let $u\in\mathcal{U}_{\mathrm{down}}=\{\mathrm{BP},\mathrm{LoS}\}$ denote a downstream task, and let $\mathbf{H}_{\mathrm{stf},i}^{u}$ be its CSI input. Unlike in reconstruction pre-training, no reconstruction-oriented CSI mask is applied to the downstream-task inputs. Following Section~IV-A, the encoder input is obtained as $\mathbf{H}_{\mathrm{in},i}^{u}=\mathcal{R}\!\left(\mathcal{T}\!\left(\mathbf{H}_{\mathrm{stf},i}^{u}\right)\right)$.
The frozen shared encoder and a task-specific adaptation head then~perform
\begin{equation}
\begin{aligned}
\mathbf{X}_{\mathrm{lat},i}^{u}
&=
\mathcal{E}_{\boldsymbol{\Theta}_{\mathrm{E}}^{\star}}
\!\left(\mathbf{H}_{\mathrm{in},i}^{u};\bm{\Phi}_{i}^{\mathrm{FS}}\right),\\
\widehat{\mathbf{Y}}_{u,i}
&=
\mathcal{A}^{u}_{\boldsymbol{\Theta}_{u}}
\!\left(\mathbf{X}_{\mathrm{lat},i}^{u}\right),
\qquad u\in\mathcal{U}_{\mathrm{down}},
\end{aligned}
\label{eq:downstream_adaptation}
\end{equation}
where $\mathcal{A}^{u}_{\boldsymbol{\Theta}_{u}}$ denotes the adaptation head parameterized by $\boldsymbol{\Theta}_{u}$.

\subsubsection{Sub-6 GHz-Aided mmWave Beam Prediction}
BP exploits readily acquired sub-6 GHz CSI to infer the optimal beam in the mmWave band. Although the two bands experience different small-scale fading, their delay and angular signatures share geometry-dependent information that is useful for beam alignment~\cite{9121328}. We consider a dual-band system in which the BS acquires the sub-6 GHz uplink CSI and selects a downlink mmWave beam from a DFT codebook $\mathcal{C}_{\mathrm{mm}}=\{\mathbf{f}_b\}_{b=1}^{B}$, where $B$ is the codebook size, $N_{\mathrm{mm}}$ is the number of antenna elements at the mmWave BS, and each $\mathbf{f}_b\in\mathbb{C}^{N_{\mathrm{mm}}\times1}$ is a normalized analog beamforming vector.

Let $\mathbf{H}_{\mathrm{stf}}^{\mathrm{sub}}$ and $\mathbf{H}_{\mathrm{stf}}^{\mathrm{mm}}$ denote the paired sub-6 GHz and mmWave CSI, respectively. For the mmWave channel matrix $\mathbf{H}_{\mathrm{stf}}^{\mathrm{mm}}[s,k]$, define $\mathbf{h}_{s,k}^{\mathrm{mm}}=\operatorname{vec}(\mathbf{H}_{\mathrm{stf}}^{\mathrm{mm}}[s,k])$. The received signal using beam $\mathbf{f}_b$ is
\begin{equation}
y_{b,s,k}^{\mathrm{mm}}
=
\left(\mathbf{h}_{s,k}^{\mathrm{mm}}\right)^{H}
\mathbf{f}_b x_{s,k}^{\mathrm{mm}}
+n_{s,k}^{\mathrm{mm}},
\label{eq:mmwave_received_signal}
\end{equation}
where $x_{s,k}^{\mathrm{mm}}$ is the transmitted symbol and $n_{s,k}^{\mathrm{mm}}$ is additive white Gaussian noise. Let $N_f^{\mathrm{mm}}$ denote the number of mmWave subcarriers. The beam index that maximizes the average beamforming gain across these subcarriers is given~by
\begin{equation}
b^{\star}
=
\operatorname*{arg\,max}_{b\in\{1,\ldots,B\}}
\frac{1}{N_f^{\mathrm{mm}}}
\sum_{k=0}^{N_f^{\mathrm{mm}}-1}
\left|
\left(\mathbf{h}_{s,k}^{\mathrm{mm}}\right)^{H}\mathbf{f}_b
\right|^2.
\label{eq:optimal_mmwave_beam}
\end{equation}
The encoder input is $\mathbf{H}_{\mathrm{stf}}^{\mathrm{BP}}=\mathbf{H}_{\mathrm{stf}}^{\mathrm{sub}}$, while $b^{\star}$ obtained through mmWave beam sweeping serves as the target label. The BP head outputs a probability vector $\widehat{\mathbf{p}}_{\mathrm{BP}}\in[0,1]^B$, whose $b$-th entry $\widehat{p}_{\mathrm{BP},b}$ is the predicted probability of beam $b$. The predicted beam and cross-entropy loss are
\begin{equation}
\widehat{b}
=\operatorname*{arg\,max}_{b\in\{1,\ldots,B\}}
\widehat{p}_{\mathrm{BP},b},
\qquad
\ell_{\mathrm{BP}}
=-\log\widehat{p}_{\mathrm{BP},b^{\star}}.
\label{eq:bp_decision_loss}
\end{equation}

\subsubsection{CSI-Based LoS Identification}
This task determines whether a direct propagation path exists between the BS and UE. Let $c\in\{0,1\}$ denote the link-state label, where $c=1$ represents LoS and $c=0$ represents non-line-of-sight (NLoS). Under the geometric model in~\eqref{spatial channel response}, a LoS channel contains a direct component whose delay and angular signature are consistent with the BS-UE geometry. An NLoS channel is instead dominated by reflected, scattered, or diffracted paths, generally exhibiting larger excess delay and stronger angular dispersion. Transforming the input $\mathbf{H}_{\mathrm{stf}}^{\mathrm{LoS}}$ into the DDA domain makes these propagation signatures explicit.

The LoS adaptation head outputs $\widehat{p}_{\mathrm{LoS}}\in[0,1]$. Given a decision threshold $\gamma_{\mathrm{LoS}}$, the estimated link state and binary cross-entropy loss are
\begin{equation}
\begin{aligned}
\widehat{c}
&=\mathbb{I}\!\left(\widehat{p}_{\mathrm{LoS}}\geq\gamma_{\mathrm{LoS}}\right),\\
\ell_{\mathrm{LoS}}
&=-c\log\widehat{p}_{\mathrm{LoS}}
-(1-c)\log\!\left(1-\widehat{p}_{\mathrm{LoS}}\right).
\end{aligned}
\label{eq:los_decision_loss}
\end{equation}

\subsection{Task-Specific Adaptation}
\label{subsec:downstream_adaptation_strategy}

For both BP and LoS identification, the task-specific adaptation head is implemented as a lightweight CNN that maps the shared latent representation $\mathbf{X}_{\mathrm{lat},i}^{u}$ to the corresponding task~output.

During downstream adaptation, the reconstruction decoder is removed and the pre-trained encoder parameters $\boldsymbol{\Theta}_{\mathrm{E}}^{\star}$ are frozen. Only the parameters $\boldsymbol{\Theta}_{u}$ of the corresponding adaptation head are updated. Let $\mathcal{D}_u$ denote the labeled data distribution for task $u$ over input-label pairs $(\mathbf{H}_{\mathrm{stf}}^{u},y_u)$, where $y_{\mathrm{BP}}=b^{\star}$ for BP and $y_{\mathrm{LoS}}=c$ for LoS identification. The head-only optimization is formulated as
\begin{equation}
\boldsymbol{\Theta}_{u}^{\star}
=
\operatorname*{arg\,min}_{\boldsymbol{\Theta}_{u}}
\mathbb{E}_{(\mathbf{H}_{\mathrm{stf}}^{u},y_u)\sim\mathcal{D}_u}
\!\left[
\ell_u\!\left(
y_u,
\mathcal{A}^{u}_{\boldsymbol{\Theta}_{u}}
\!\left(\mathbf{X}_{\mathrm{lat}}^{u}\right)
\right)
\right],
\label{eq:downstream_training_objective}
\end{equation}
where $\mathbf{X}_{\mathrm{lat}}^{u}$ is produced by the fixed encoder in~\eqref{eq:downstream_adaptation}, and $\ell_u$ is the cross-entropy loss in~\eqref{eq:bp_decision_loss} for BP or the binary cross-entropy loss in~\eqref{eq:los_decision_loss} for LoS identification. By learning only the lightweight task-specific mapping, this strategy preserves the pre-trained channel representation and reduces the amount of labeled data required for adaptation.
\end{revisionblue}

\section{Experimental Setup}
\label{sec:experiment_setup}

This section establishes the dataset and experimental protocols used to evaluate AirFM-DDA. We first describe the generation and partitioning of the heterogeneous CSI dataset. We then specify the model configurations, pre-training and downstream adaptation settings, comparison baselines, and evaluation metrics. These settings are used consistently throughout the experimental results reported in Section~\ref{sec:experiment_results}.

\subsection{Dataset Generation}
\label{sec:dataset_generation}

In this paper, we construct a large-scale, heterogeneous, high-dimensional CSI dataset in the STF domain. Specifically, we leverage multipath channel data from 22 DeepMIMO cities (indexed 0--19, 23, and 27) to generate 2.85 million (2.85M) CSI samples with 384.88 billion CSI points at a carrier frequency of $f_c=3.5$~GHz~\cite{alkhateeb2019deepmimo}. For the BP task, each sub-6 GHz CSI sample is paired with the corresponding mmWave CSI generated at a carrier frequency of $f_c^{\mathrm{mm}}=28$~GHz. For each city, we further process the same multipath data into two categories, coarse-grained (Frame Structure-A, FS-A) and fine-grained (FS-B), to capture different frequency sampling granularities. 
We adopt the 30~kHz 5G NR numerology~\cite{3gpp38211}, with FS-A and FS-B sampling the resulting OFDM grid at effective frequency intervals of 1.44~MHz and 0.36~MHz, respectively.
The BS uses an $8\mathrm{H}\times4\mathrm{V}$ half-wavelength-spaced UPA, and the UE is equipped with a single omnidirectional antenna. To cover mobility from quasi-static to highly dynamic regimes, UE speeds are uniformly sampled over the range of 0--120 km/h, and the UEs move along linear trajectories.

Following the data generation, the raw CSI samples are normalized by their $\ell_2$-norm to ensure unit energy. To simulate realistic channel conditions, complex Gaussian noise is injected into the normalized CSI with a randomly sampled signal-to-noise ratio (SNR) in the range from 5 to 20 dB. For the inference phase, a default SNR of 10 dB is applied unless otherwise specified. After preprocessing, the dataset is stratified by city indices. Specifically, cities 0--17 serve as the pre-training set (with 10\% withheld for validation), whereas cities 18, 19, 23, and 27 comprise the test-domain dataset. For the latter, 10\% of samples are allocated to final evaluation, and the remaining 90\% are employed to train baseline models. Table~\ref{table: Dataset Settings} summarizes the specific configurations for the test set.

\begin{table}[!tb]
\centering
\begin{threeparttable}
\scriptsize
\setlength{\tabcolsep}{4.5pt}
    \caption{Frame-structure parameters for the test set of the proposed dataset.}

\begin{tabular}{>{\centering\arraybackslash}m{1.6cm}
                >{\centering\arraybackslash}m{1.25cm}
                >{\centering\arraybackslash}m{0.6cm}
                >{\centering\arraybackslash}m{0.4cm}
                >{\centering\arraybackslash}m{0.6cm}
                >{\centering\arraybackslash}m{0.4cm}
                >{\centering\arraybackslash}m{1.2cm}}
\toprule[0.8pt]
City & \makecell[c]{Frame\\Structure*} & $\Delta f$ (MHz) & $N_{f}$ & $\Delta t$ (ms) & $N_t$ & Sample Count \\
\toprule[0.8pt]
\multirow{2}{*}{Denver (18)}
 & FS-A & $1.44$ & 32 & 0.50 & 80 & 88631 \\
 & FS-B & $0.36$ & 64 & 0.50 & 80 & 88631 \\
\hline
\multirow{2}{*}{Oklahoma City (19)}
 & FS-A & $1.44$ & 32 & 0.50 & 80 & 82222 \\
 & FS-B & $0.36$ & 72 & 0.50 & 80 & 82222 \\
\hline
\multirow{2}{*}{Beijing (23)}
 & FS-A & $1.44$ & 32 & 0.50 & 40 & 45706 \\
 & FS-B & $0.36$ & 32 & 0.50 & 40 & 45706 \\
\hline
\multirow{2}{*}{Rio de Janeiro (27)}
 & FS-A & $1.44$ & 64 & 0.50 & 80 & 19739 \\
 & FS-B & $0.36$ & 128 & 0.50 & 80 & 19739 \\
\bottomrule[0.8pt]
\end{tabular}
\label{table: Dataset Settings}
\begin{tablenotes}
\footnotesize
\item[*] FS-A and FS-B denote two frequency sampling settings for the same multipath data in each city.
\end{tablenotes}
    
\end{threeparttable}
\vspace{-0.2cm}
\end{table}

\subsection{Implementation and Evaluation Protocols}
\label{subsec:implementation_evaluation_protocols}

This subsection describes the model configurations, training and adaptation settings, baselines, and evaluation metrics used in the experiments.

\subsubsection{Network Architecture}
AirFM-DDA comprises Small, Base, and Large variants that share a patch size of $P=1$, a window size of $P_{\mathrm{w}}=8$, $L_{\mathrm{E}}=3$ encoder modules, and $L_{\mathrm{D}}=1$ reconstruction decoder module, with $L_1=6$ blocks per module. Their scale-specific embedding dimensions, attention-head counts, and parameter counts are summarized in Tables~\ref{tab:backbone_hparams} and~\ref{table:Swin_Scaling}. The encoder output serves as the shared representation for downstream tasks; the lightweight adaptation heads are detailed in Section~\ref{subsec:downstream_setup}.

\begin{table}[!t]
\centering
\begin{threeparttable}
\caption{Architectural configurations of AirFM-DDA.}
\label{tab:backbone_hparams}
\setlength{\tabcolsep}{2.5pt}
\renewcommand{\arraystretch}{1.15}
\begin{tabular}{>{\RaggedRight\arraybackslash}p{5.2cm} | >{\Centering\arraybackslash}p{2.5cm}}
\toprule
\textbf{Hyperparameter} & \textbf{Value} \\
\midrule
Max. time samples $N_{t,\max}$ & $80$ \\
Max. frequency samples $N_{f,\max}$ & $128$ \\
Input channels & $64$ \\
Patch size $P$ & $1$ \\
Window size $P_\text{w}$ & $8$ \\
Embedding dimension $C$ & varied* \\
Window-based attention block count $L_1$  & $6$ \\
Window-based attention modules in shared encoder $L_{\mathrm{E}}$ & $3$ \\
Window-based attention modules in reconstruction decoder $L_{\mathrm{D}}$ & $1$ \\
Number of attention heads $h$ & varied* \\
MLP expansion ratio & $2$ \\
FS-PE reference resolution $(r_\tau^{\text{ref}}, r_\nu^{\text{ref}})$ & $(1.36 \;\text{ns},\,3.15\;\text{Hz})$ \\
Normalization & LayerNorm \\
\bottomrule
\end{tabular}
\begin{tablenotes}
\footnotesize
\item[*] Detailed parameters are listed in Table~\ref{table:Swin_Scaling}.
\end{tablenotes}
\end{threeparttable}
\vspace{-0.2cm}
\end{table}

\begin{table}[!t]
\centering
\caption{AirFM-DDA scaling configurations and parameter counts.}
\label{table:Swin_Scaling}
\begin{tabular}{>{\raggedright\arraybackslash}m{2.4cm}
                >{\centering\arraybackslash}m{1.3cm}
                >{\centering\arraybackslash}m{1.9cm}
                >{\centering\arraybackslash}m{1.3cm}}
\toprule
Model & Embedding Dimension & Number of Attention Heads & Parameters (M) \\
\midrule
AirFM-DDA-Small  & 512 & 4 & 62.62 \\
AirFM-DDA-Base & 640 & 5 & 97.69 \\
AirFM-DDA-Large  & 768 & 6 & 140.52 \\
\bottomrule
\end{tabular}
\vspace{-0.5cm}
\end{table}

\subsubsection{Pre-Training Strategy}
Following the two-stage strategy described in Section~IV-C, Stage I trains on 10\% of the pre-training data for 10 epochs using the easier configurations $x=0.75$ and $D=2$, whereas Stage II uses the full dataset for 30 epochs with $x\in[0.5,0.75]$ and $D\in\{2,4\}$. The complete optimization and learning-rate schedules are summarized in Table~\ref{table: network optimization}, and all experiments use bfloat16 mixed precision. All experiments are conducted with BF16 mixed precision on a multi-GPU server providing approximately 7.9 PFLOPS of theoretical BF16 Tensor Core compute, together with dual Intel Xeon Platinum 8558 CPUs.

\begin{table}[t]
\centering
\caption{Optimization and learning rate schedules for AirFM-DDA pre-training.}
\label{table: network optimization}
\setlength{\tabcolsep}{8pt}
\renewcommand{\arraystretch}{1.15}
\begin{tabular}{>{\raggedright\arraybackslash}m{3.1cm}
                >{\centering\arraybackslash}m{1.7cm}
                >{\centering\arraybackslash}m{1.7cm}}
\toprule
\textbf{Setting} & \textbf{Stage I} & \textbf{Stage II} \\
\midrule
Optimizer & Adam & Adam \\
LR scheduler & StepLR & StepLR \\
Initial learning rate & $6\times10^{-5}$ & $3\times10^{-5}$ \\
Minimum learning rate & $3\times10^{-5}$ & $5\times10^{-6}$ \\
LR decay factor & $0.8$ & $0.9$ \\
LR decay interval (steps) & $400$ & $5000$ \\
Number of epochs & $10$ & $30$ \\
Global batch size & $176$ & $176$ \\
Steps per epoch & $1{,}215$ & $12{,}173$ \\
\midrule 
Observation ratio $x$ & $0.75$ & $0.5 \text{--} 0.75$ \\
Pilot spacing $D$ & $2$ & $2, 4$ \\
\bottomrule
\end{tabular}
\vspace{-0.45cm}
\end{table}

\begin{revisionblue}
\subsubsection{Downstream Adaptation Setup}
\label{subsec:downstream_setup}
For both downstream tasks, the shared encoder is initialized with the reconstruction-pretrained weights and kept frozen, while only the corresponding lightweight adaptation head is updated. The downstream adaptation is configured as follows. Each head uses a lightweight CNN classifier consisting of an initial one-by-one feature projection, four residual blocks based on depthwise-separable convolutions, global average pooling, and a two-layer classifier. In each residual block, a depthwise convolution extracts local patterns separately within each feature channel, a one-by-one pointwise convolution mixes information across channels, and a shortcut connection adds the block input to the transformed features. The BP head outputs $B=64$ beam logits, whereas the LoS head outputs a binary LoS probability. Both heads are trained for 50 epochs with learning rate $10^{-3}$.

For BP, we use a critically sampled DFT codebook with $B=64$, and obtain each optimal beam label by exhaustive beam sweeping according to~\eqref{eq:optimal_mmwave_beam}. For both BP and LoS identification, City 18 in Table~\ref{table: Dataset Settings} is used as the target scenario. We allocate 90\% of its samples to the training split and hold out the remaining 10\% for evaluation. To evaluate label efficiency, we sample 10\%, 25\%, 50\%, and 100\% of the training split as labeled data, while all methods are evaluated on the same held-out 10\% split.
\end{revisionblue}

\subsubsection{Baselines}
\textcolor{blue}{For quantitative evaluation, we select reproducible baselines that are directly aligned with the evaluated tasks and span wireless-native FMs, cross-domain adapted FMs, task-specific networks, and conventional model-based methods. Because these methods differ in task scope and CSI-input compatibility, we adopt task-matched comparisons.}

\begin{itemize}
    \item \textcolor{blue}{\textbf{WiFo (Zero-Shot)}~\cite{liu2025wifo}: We select WiFo as the principal wireless-native FM baseline for TP and FP because, like AirFM-DDA, it performs reconstruction-oriented pre-training on heterogeneous STF-domain CSI and supports zero-shot channel reconstruction. WiFo-Large is initialized with its publicly released weights, further trained on our constructed pre-training dataset for 30 epochs with an initial learning rate of $8\times 10^{-5}$, and evaluated zero-shot on the same unseen test scenarios as AirFM-DDA.}

    \item \textcolor{blue}{\textbf{LWM}~\cite{alikhani2024large}: We include LWM as a wireless-native FM baseline for BP and LoS identification. LWM learns general wireless representations from SF-domain CSI and adapts them using task-specific heads. For a fair comparison, LWM is pre-trained on our dataset, and its original BP and LoS heads are replaced with the same lightweight CNN architectures used by AirFM-DDA.}

    \item \textbf{LLM4WM (Full-Shot)}~\cite{liu2025llm4wm}: We include LLM4WM as a cross-domain adapted FM baseline that supports TP, FP, CE, BP, and LoS identification. Its network architecture and training hyperparameters follow the original setup in~\cite{liu2025llm4wm}. For the reconstruction experiments, LLM4WM is evaluated under the full-shot setting; for downstream adaptation, it uses the labeled-data ratios specified above.
        
    \item \textbf{Transformer (Full-Shot)}~\cite{jiang2022transformer}: We adopt this method as a baseline for channel prediction tasks. While the original work focuses solely on the TP task, we extend it to the FP task by adapting the masking strategy to align with our evaluation protocol. This model is trained for 50 epochs using the Adam optimizer, with a learning rate annealed from $10^{-4}$ to $10^{-6}$ via a cosine decay schedule. For evaluation, we report its performance under the full-shot~configuration.
    
    \item \textbf{U-Net (Full-Shot)}: We employ a customized U-Net as the task-specific baseline for the CE task. Implemented as a fully convolutional network, it features a four-level encoder-decoder architecture. The model is trained for 100 epochs using the Adam optimizer, guided by a cosine annealing schedule that decays the learning rate from $10^{-3}$ to $10^{-5}$. The evaluation is conducted under a full-shot setting.

    \item \textcolor{blue}{\textbf{Autoregressive Moving-Average (ARMA) Predictor}~\cite{barbieri2009arma}: We employ a complex ARMA$(4,2)$ predictor as a conventional baseline for TP. For each sample, shared AR and MA coefficients are estimated from the mean-centered visible CSI prefix, and the masked suffix is recursively predicted with the MA component initialized using the visible-prefix residuals.}

    \item \textcolor{blue}{\textbf{Linear Minimum Mean-Square Error (LMMSE) Estimator}~\cite{edfors1998ofdm}: We employ covariance-aided LMMSE as a conventional baseline for FP and CE. For each frame configuration, the frequency-domain mean and covariance are estimated exclusively from the target-domain calibration split. LMMSE reconstructs the masked frequency suffix from the visible prefix for FP and the complete channel from noisy comb pilots for CE.}

    \item \textcolor{blue}{\textbf{DeepBP}~\cite{9121328}: We adopt DeepBP as a task-specific baseline for BP. It directly maps the observed sub-6 GHz CSI to the mmWave beam label and is trained only on labeled BP data from the target scenario. It is implemented as a six-hidden-layer MLP classifier.}

    \item \textcolor{blue}{\textbf{ST-CNN}~\cite{sun2023channel}: We adopt ST-CNN as a task-specific baseline for LoS identification. The channel observation is converted into a $224\times224$ time-frequency image through the Stockwell transform and then processed by a compact classifier with four convolutional layers and two fully connected layers. Its original output layer is replaced with a binary LoS/NLoS classifier.}

    \item \textcolor{blue}{\textbf{Raw CSI}: This direct supervised baseline feeds the original CSI representation into the same lightweight classifier used by AirFM-DDA, without reconstruction-oriented pre-training or shared DDA-domain encoding. It isolates the benefit of the learned foundation representation.}

    \item \textcolor{blue}{\textbf{KNN}: We include K-nearest neighbors as a classical non-parametric baseline for BP and LoS identification.}

\end{itemize}

\textcolor{blue}{For TP and FP, WiFo serves as the zero-shot learned baseline. ARMA serves as an observation-only conventional reference for TP, whereas LMMSE serves as a target-calibrated conventional reference for FP and CE. According to task applicability, LLM4WM, Transformer, U-Net, and other methods are trained separately for each target configuration using matched target-domain data, and their full-shot results are reported to provide in-domain reference performance.}

\subsubsection{Evaluation Metrics}
{We report reconstruction and downstream-task accuracy together with computational cost.}

\paragraph{Reconstruction accuracy (Recon. Acc.)}
{Reconstruction accuracy is measured by NMSE in dB, computed over the masked regions for TP and FP and over the complete CSI tensor for CE.}

\begin{figure*}[!t]
    \centering
    \begin{subfigure}[t]{0.47\linewidth}
        \centering
\includegraphics[width=0.90\linewidth]{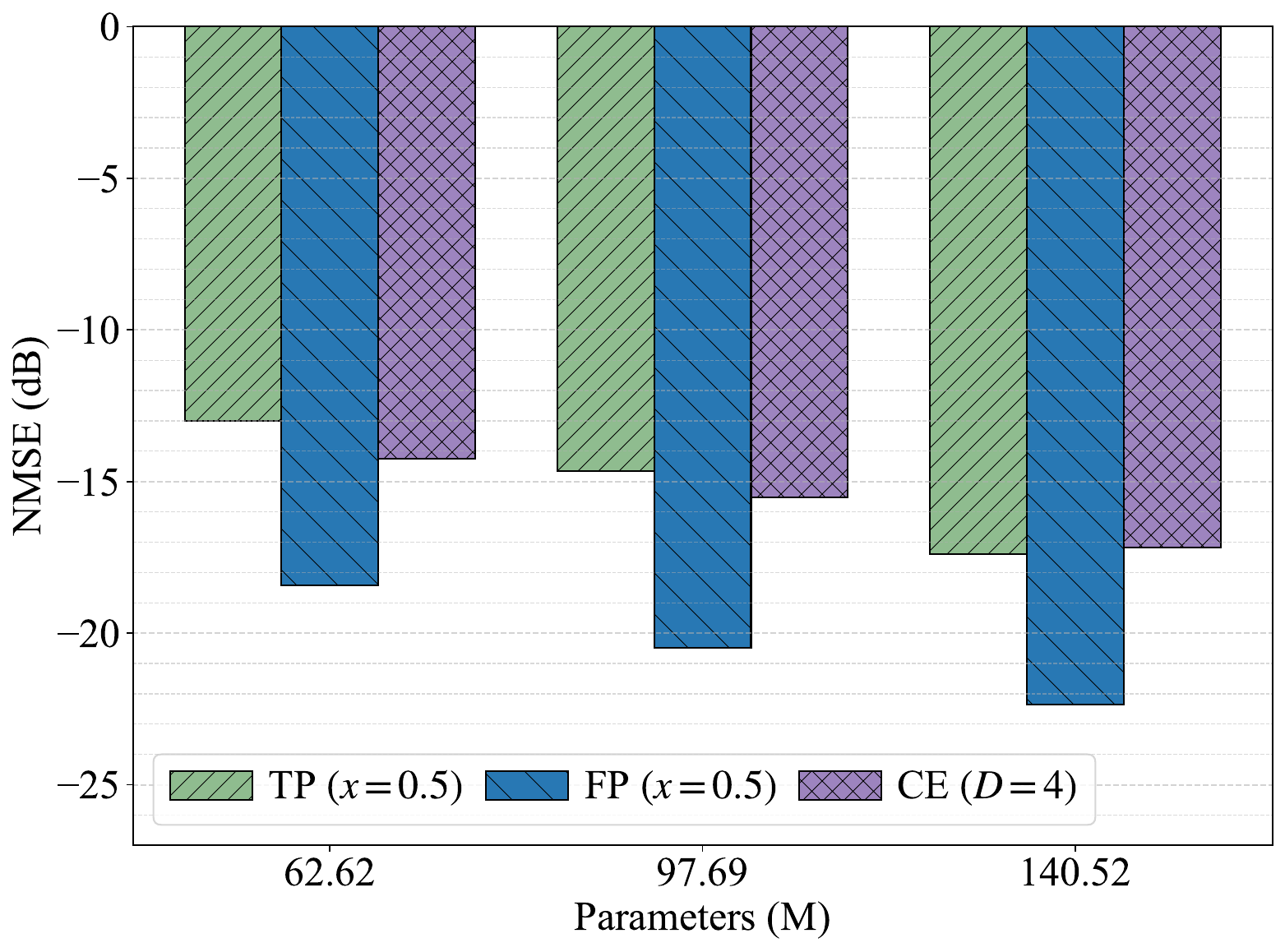}
        \caption{Backbone scale ablation.}
        \label{fig:backbone_scale_ablation_sub}
    \end{subfigure}
    \hfill
    \begin{subfigure}[t]{0.47\linewidth}
        \centering
\includegraphics[width=0.90\linewidth]{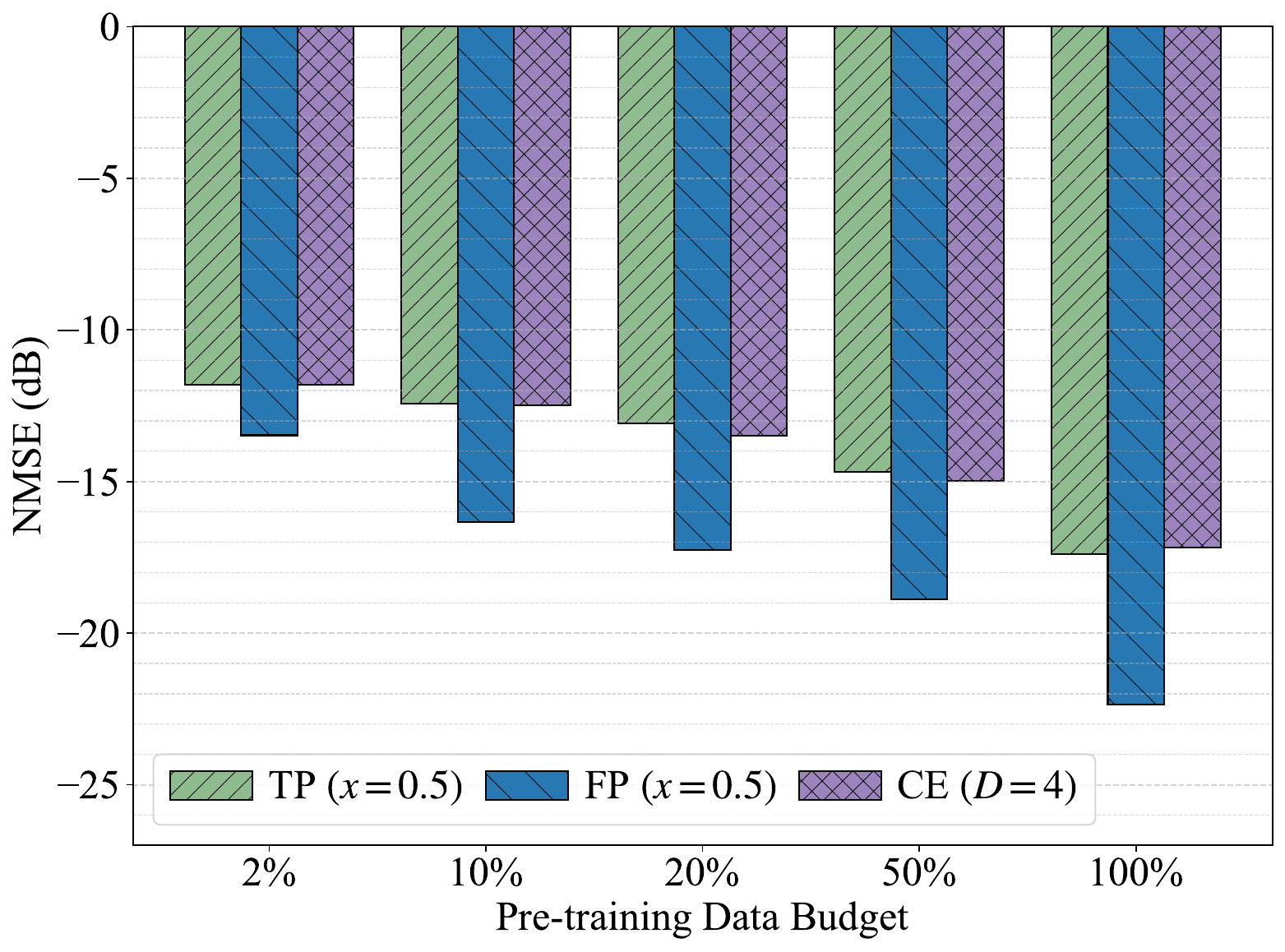}
        \caption{Pre-training data budget ablation for AirFM-DDA-Large.}
        \label{fig:pretraining_data_ablation_sub}
    \end{subfigure}

    \vspace{-0.15cm}
    \caption{AirFM-DDA ablation results.}
    \label{fig:scaling_ablations}
\vspace{-0.45cm}
\end{figure*}

\begin{revisionblue}
\paragraph{Beam prediction}
For BP, we use Top-1 accuracy, where a prediction is correct if the highest-probability beam index matches the ground-truth optimal beam index.

\paragraph{LoS identification}
For LoS identification, we use the F1 score, i.e., the harmonic mean of precision and recall.
\end{revisionblue}

\paragraph{Computational cost (Comp. Cost)}
Computational cost is evaluated using five metrics: (i) \textit{Params}, the total number of parameters; (ii) \textit{FLOPs}, the floating-point operations per single-sample forward pass; (iii) \textit{Train Thpt}, the training throughput (samples/s under full GPU utilization); (iv) \textit{Infer Latency}, the single-request inference latency at batch size one; and (v) \textit{Peak GPU Mem}, the peak GPU memory for single-sample inference.

\section{Experimental Results}
\label{sec:experiment_results}

With the dataset and evaluation protocols established in Section~\ref{sec:experiment_setup}, this section presents a comprehensive empirical evaluation of AirFM-DDA. We first conduct ablation studies on the backbone architecture, model scaling, and pre-training data size. We then report zero-shot CSI reconstruction and few-shot downstream adaptation results, followed by an efficiency evaluation and an assessment of cross-dataset generalization.

\begin{table}[!t]
    \centering
    \begin{threeparttable}
    \caption{Backbone architecture ablation on the validation set and computational efficiency comparison.}
    \label{table:backbone_ablation_efficiency}
    \footnotesize
    \setlength{\tabcolsep}{4pt}
    \renewcommand{\arraystretch}{1.2}
    \begin{tabular*}{\columnwidth}{@{\extracolsep{\fill}}l|l|ccc}
    \toprule

    \multicolumn{1}{c}{}
    & \multicolumn{1}{c|}{}
    & \makecell{\textbf{AirFM-DDA} \\ \textbf{w/ ViT}\tnote{*}}
    & \makecell{\textbf{AirFM-DDA}\tnote{†} \\ \textbf{w/o FS-PE}}
    & \makecell{\textbf{AirFM}\\\textbf{-DDA}\tnote{†}} \\
    \midrule

    \multirow{3}{*}{\makecell[c]{\textit{Recon.} \\ \textit{Acc.}}}
    & TP $(x=0.5)$ & -12.65 & -16.74 & \textbf{-17.34} \\
    & FP $(x=0.5)$ & -15.33 & -20.09 & \textbf{-22.01} \\
    & CE $(D=4)$   & -10.72 & -14.14 & \textbf{-17.47} \\
    \midrule[0.7pt]

    \multirow{5}{*}{\makecell[c]{\textit{Comp.} \\ \textit{Cost}}}
    & Params {\footnotesize (M)}                  & 140.52  & 140.52  & 140.52 \\
    & FLOPs\tnote{‡} {\footnotesize(G)}          & 5306.92 & 1462.64 & 1462.65 \\
    & Train Thpt {\footnotesize(samples/s)}      & 7.25    & 83.95   & 82.93 \\
    & Infer Latency\tnote{‡} {\footnotesize(ms)} & 246.32  & 31.61   & 32.25 \\
    & Peak GPU Mem\tnote{‡} {\footnotesize(GiB)} & 11.13   & 2.01    & 2.01 \\
    \bottomrule
    \end{tabular*}

    \begin{tablenotes}[flushleft]
    \footnotesize
    \item[*] ``ViT'' denotes the global-attention ablation.
    \item[†] ``AirFM-DDA'' refers to the AirFM-DDA-Large configuration.
    \item[‡] Inference uses batch size 1; the CSI sample dimensions along the spatial, temporal, and frequency axes are $32$, $80$, and $128$, respectively.
    \end{tablenotes}
    \end{threeparttable}
    \vspace{-0.6cm}
\end{table}

\subsection{Ablation Studies}
We ablate the backbone architecture, model scale, and pre-training data size to assess their contributions to AirFM-DDA.

\begin{figure*}[t]

    \captionsetup{labelfont={color=blue},textfont={color=blue}}
    \captionsetup[subfigure]{labelfont={color=blue},textfont={color=blue}}
    \centering

    \begin{subfigure}[t]{0.47\linewidth}
        \centering
        \includegraphics[width=0.90\linewidth]{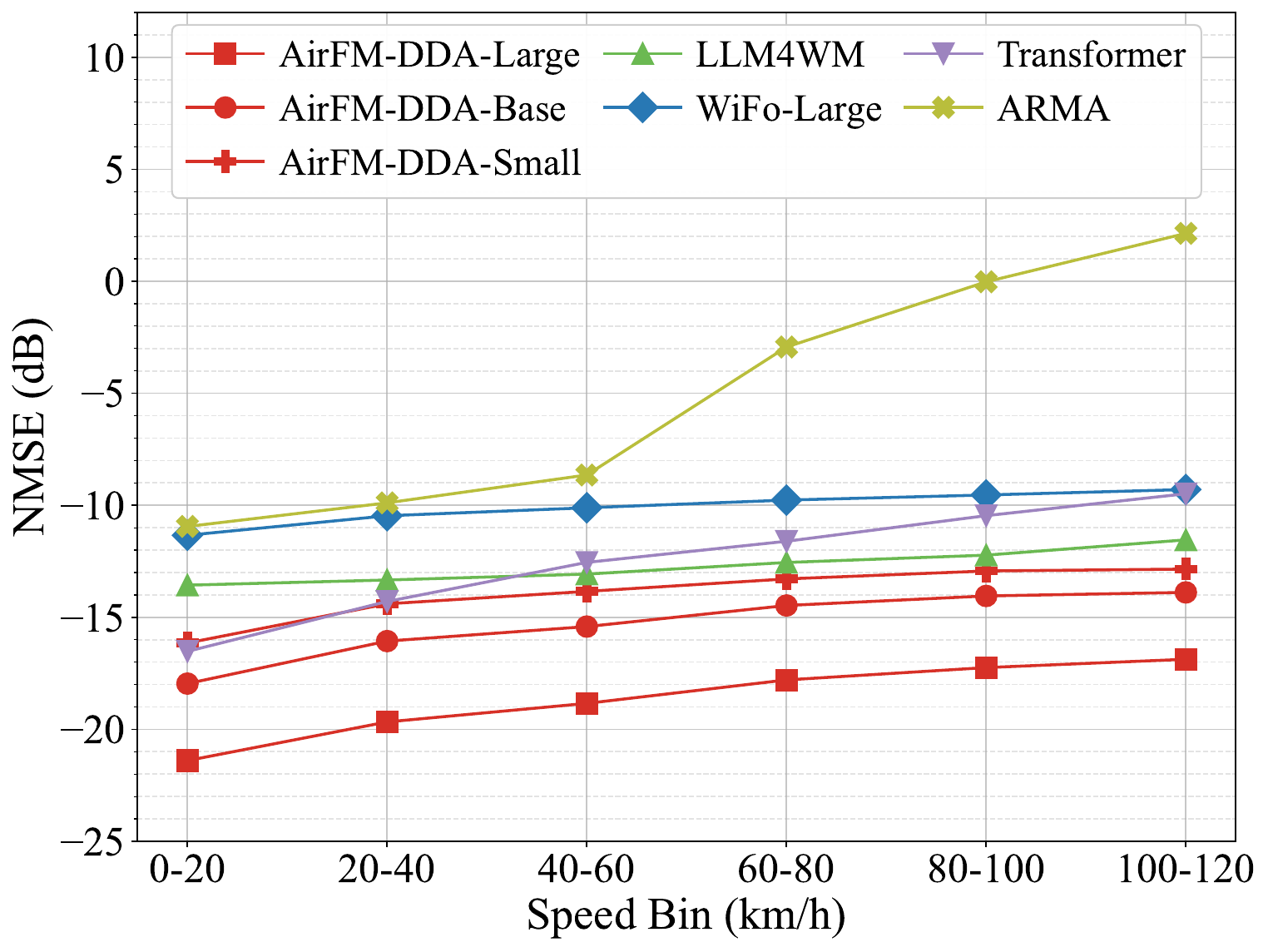}
        \caption{TP performance with $x=0.75$.}
        \label{fig:tp_x075_sub}
    \end{subfigure}
    \hfill
    \begin{subfigure}[t]{0.47\linewidth}
        \centering
        \includegraphics[width=0.90\linewidth]{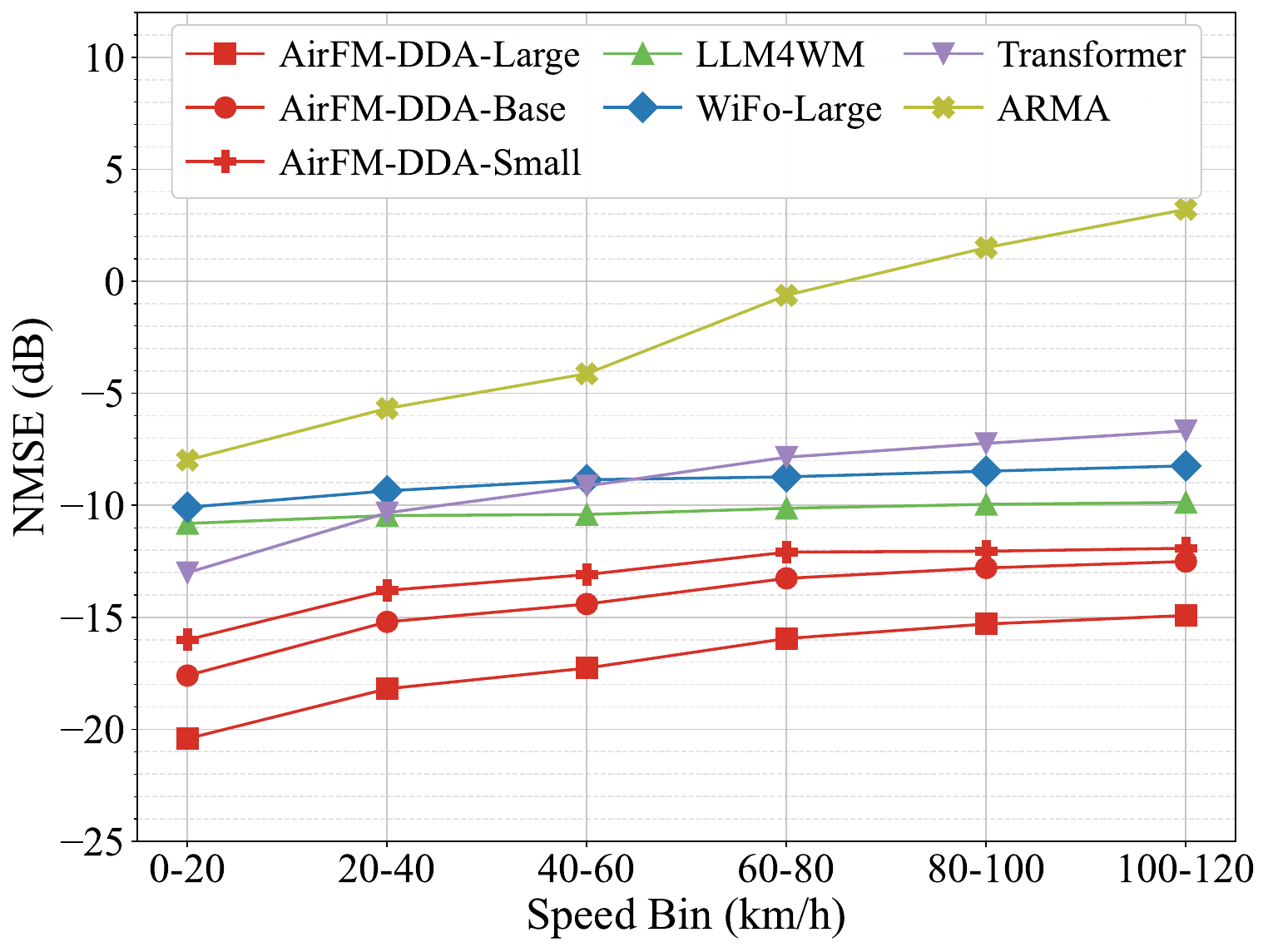}
        \caption{TP performance with $x=0.5$.}
        \label{fig:tp_x05_sub}
    \end{subfigure}

    \vspace{-0.15cm}
    \caption{NMSE performance of the TP task on the test set across various user speeds.}
    \label{fig:TP Task Performance}

    \vspace{0.15cm}

    \begin{subfigure}[t]{0.47\linewidth}
        \centering
        \includegraphics[width=0.90\linewidth]{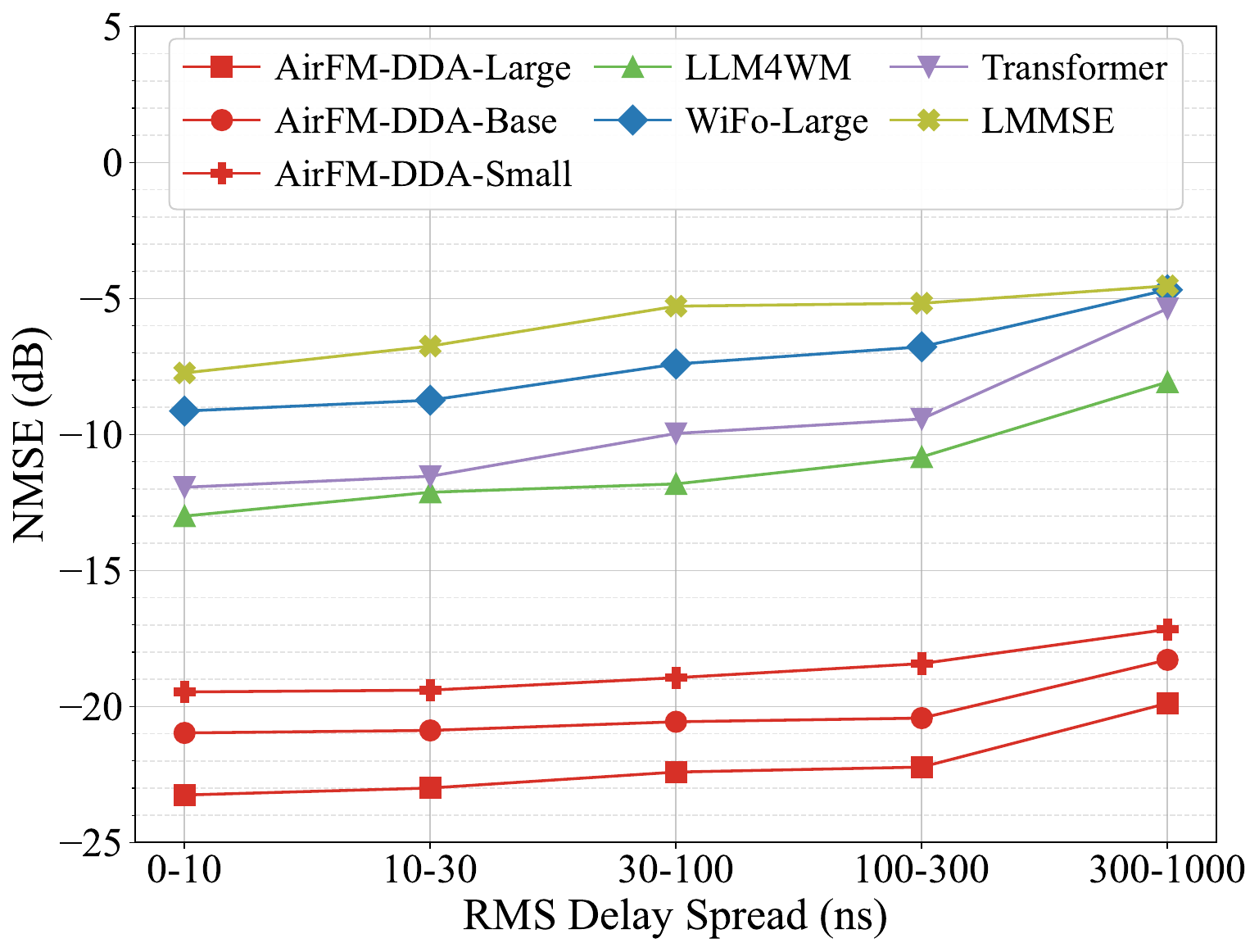}
        \caption{FP performance with $x=0.75$.}
        \label{fig:fp_x075_sub}
    \end{subfigure}
    \hfill
    \begin{subfigure}[t]{0.47\linewidth}
        \centering
        \includegraphics[width=0.90\linewidth]{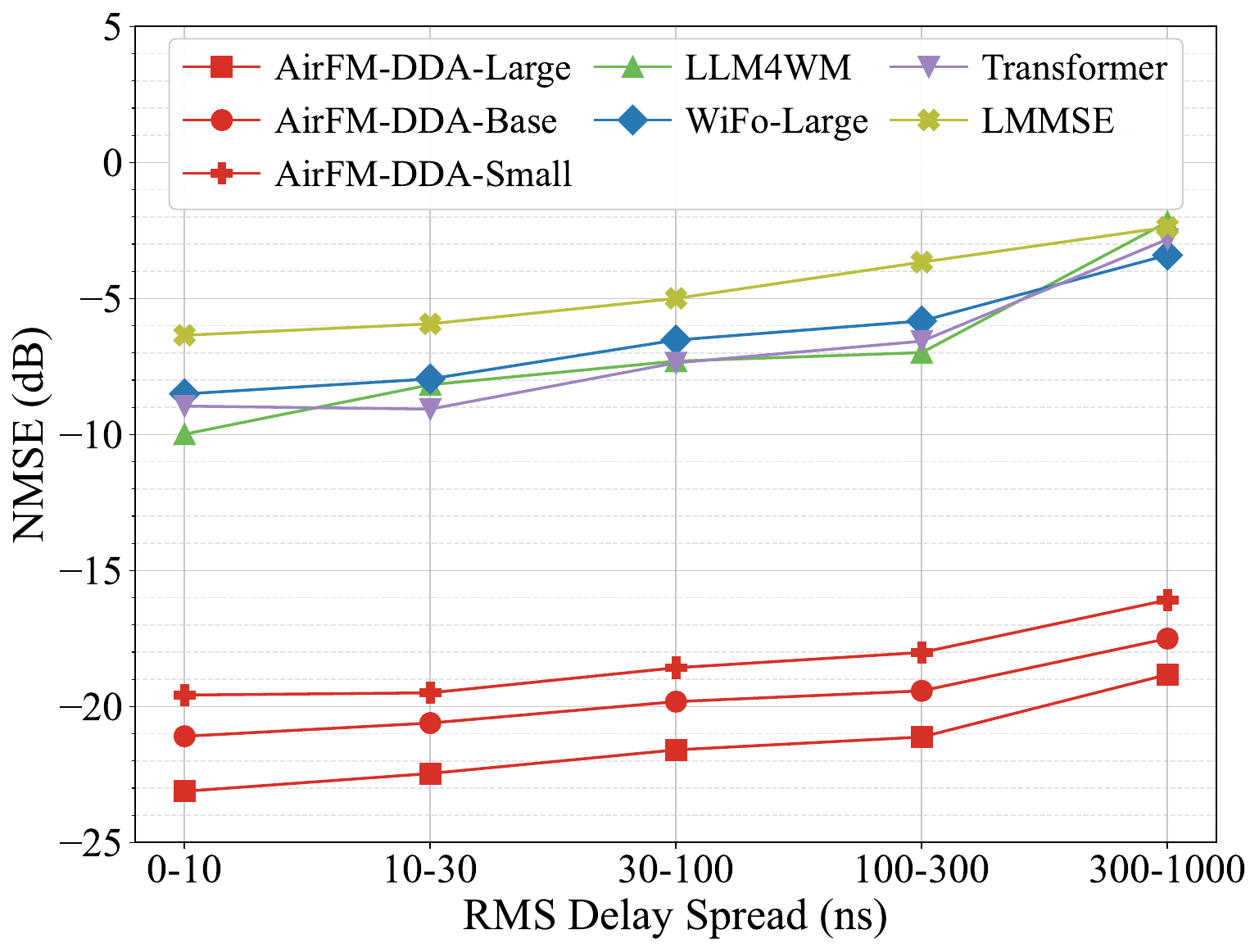}
        \caption{FP performance with $x=0.5$.}
        \label{fig:fp_x05_sub}
    \end{subfigure}

    \vspace{-0.15cm}
    \caption{NMSE performance of the FP task on the test set across different RMS delay-spread ranges.}
    \label{fig:FP Task Performance}

    \vspace{-0.5cm}
\end{figure*}

\begin{figure*}[t]

    \vspace{-0.1cm}

    \captionsetup{labelfont={color=blue},textfont={color=blue}}
    \captionsetup[subfigure]{labelfont={color=blue},textfont={color=blue}}
    \centering

    \begin{subfigure}[t]{0.47\linewidth}
        \centering
        \includegraphics[width=0.90\linewidth]{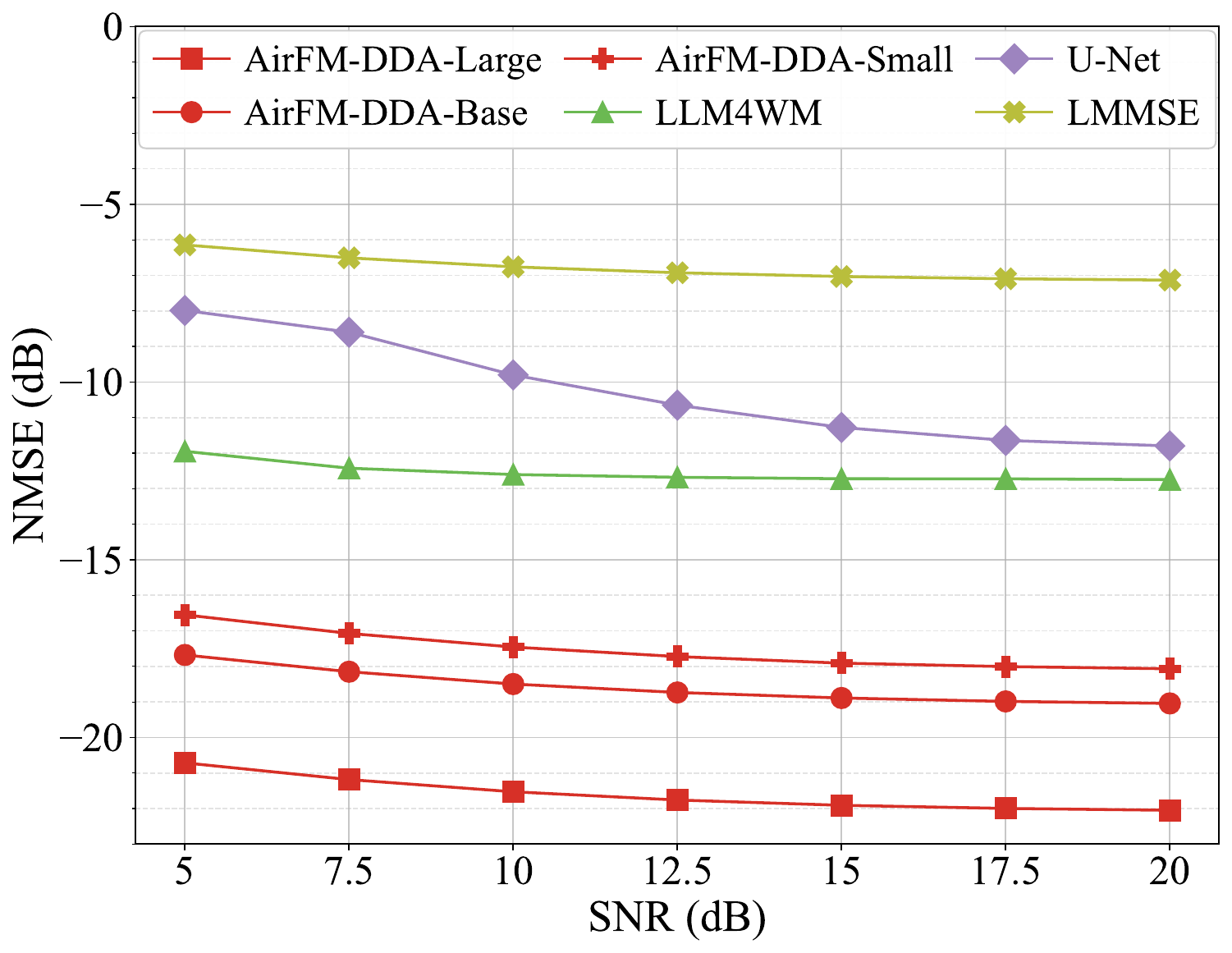}
        \caption{CE performance with $D=2$.}
        \label{fig:ce_d2_sub}
    \end{subfigure}
    \hfill
    \begin{subfigure}[t]{0.47\linewidth}
        \centering
        \includegraphics[width=0.90\linewidth]{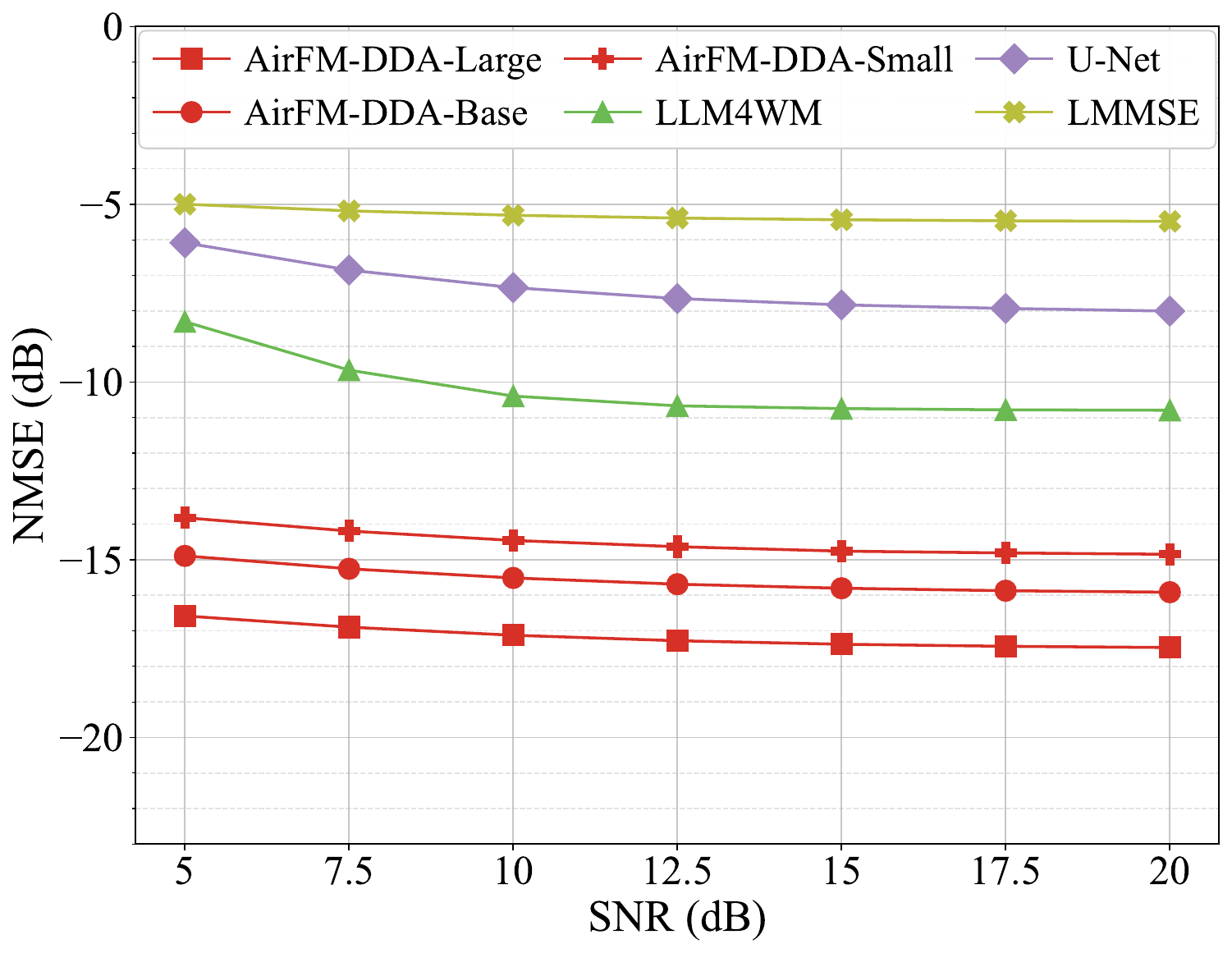}
        \caption{CE performance with $D=4$.}
        \label{fig:ce_d4_sub}
    \end{subfigure}

    \vspace{-0.15cm}
    \caption{NMSE performance of the CE task on the test set under different SNR conditions.}
    \label{fig:CE Task Performance}

    \vspace{0.15cm}

    \begin{minipage}{\textwidth}
    \centering
    \captionsetup{type=table}

    \begin{threeparttable}
    \caption{Few-shot adaptation performance on LoS identification and BP over the target scenario.}
    \label{tab:bp_los_adaptation}

    \scriptsize
    \setlength{\tabcolsep}{2.2pt}
    \renewcommand{\arraystretch}{1.08}

    \begin{tabular*}{\textwidth}{@{\extracolsep{\fill}}llccccccccc}
    \toprule
    Task & Metric & \makecell{Training\\Ratio}
    & \makecell{AirFM-DDA\\-Large}
    & \makecell{AirFM-DDA\\-Base}
    & \makecell{AirFM-DDA\\-Small}
    & LLM4WM
    & LWM
    & \makecell{Raw\\CSI}
    & \makecell{Task\\Specific}
    & KNN \\
    \midrule

    \multirow{4}{*}{{LoS Identification}}
    & \multirow{4}{*}{\makecell{F1 Score\\(\%)}}
    & 10\%  & \textbf{96.39} & 95.28 & 94.24 & 88.30 & 91.86 & 84.06 & 84.76 & 65.10 \\
    & & 25\%  & \textbf{96.65} & 95.82 & 94.66 & 91.00 & 92.18 & 88.04 & 89.93 & 69.84 \\
    & & 50\%  & \textbf{97.02} & 96.23 & 95.26 & 92.40 & 93.24 & 90.77 & 91.70 & 70.80 \\
    & & 100\% & \textbf{97.59} & 97.09 & 95.52 & 93.50 & 94.24 & 92.25 & 93.77 & 75.00 \\

    \midrule

    \multirow{4}{*}{BP}
    & \multirow{4}{*}{\makecell{Top-1 Acc.\\(\%)}}
    & 10\%  & \textbf{65.95} & 61.43 & 57.41 & 49.62 & 44.75 & 32.44 & 35.18 & 21.08 \\
    & & 25\%  & \textbf{70.24} & 67.00 & 62.25 & 55.97 & 52.16 & 38.41 & 45.83 & 28.81 \\
    & & 50\%  & \textbf{76.86} & 72.68 & 67.50 & 60.60 & 58.04 & 43.44 & 51.57 & 34.66 \\
    & & 100\% & \textbf{82.82} & 79.51 & 77.16 & 66.38 & 64.59 & 47.31 & 56.84 & 38.41 \\

    \bottomrule
    \end{tabular*}

    \begin{tablenotes}
    \footnotesize
    \item[*] The task-specific baseline is ST-CNN for LoS identification and DeepBP for BP.
    \end{tablenotes}

    \end{threeparttable}
    \end{minipage}

    \vspace{-0.5cm}

\end{figure*}

\begin{figure*}[t]

\begin{minipage}{\textwidth}
\color{blue}
\centering
\captionsetup{type=table}
\begin{threeparttable}
\caption{Computational cost comparison of different models.}
\label{tab:response_efficiency_reviewerone}
\scriptsize
\setlength{\tabcolsep}{1.2pt}
\renewcommand{\arraystretch}{1.15}
\begin{tabular*}{\textwidth}{@{\extracolsep{\fill}}lccccccccc@{}}
\toprule
Metric
& \makecell{AirFM-DDA\\-Small}
& \makecell{AirFM-DDA\\-Base}
& \makecell{AirFM-DDA\\-Large}
& WiFo-Large
& LLM4WM
& U-Net
& Transformer
& ARMA
& LMMSE \\
\midrule
Params (M) & 62.62 & 97.69 & 140.52 & 85.88 & 95.21 & 11.71 & 9.79 & -- & -- \\
FLOPs\tnote{*} (G) & 657.02 & 1020.07 & 1462.65 & 1550.00 & 2.33 & 2.15 & 38.55 & 0.15 & 0.085 \\
Infer Latency\tnote{*} (ms) & 25.06 & 28.28 & 32.25 & 253.21 & 5.08 & 3.48 & 10.78 & 2.11 & 0.063 \\
Peak GPU Mem (GiB) & 1.44 & 1.68 & 2.01 & 13.66 & 1.21 & 0.84 & 0.90 & 0.027 & 0.014 \\
\bottomrule
\end{tabular*}
\begin{tablenotes}
\footnotesize
\item[*] For AirFM-DDA, the reported FLOPs and inference latency include the forward and inverse STF–DDA transforms.
\end{tablenotes}
\end{threeparttable}
\end{minipage}

\par\vspace{0.8em}
\setcounter{subfigure}{0}

    \captionsetup{labelfont={color=blue},textfont={color=blue}}
    \captionsetup[subfigure]{labelfont={color=blue},textfont={color=blue}}
    \centering
    \begin{subfigure}[t]{0.47\linewidth}
        \centering
\includegraphics[width=0.90\linewidth]{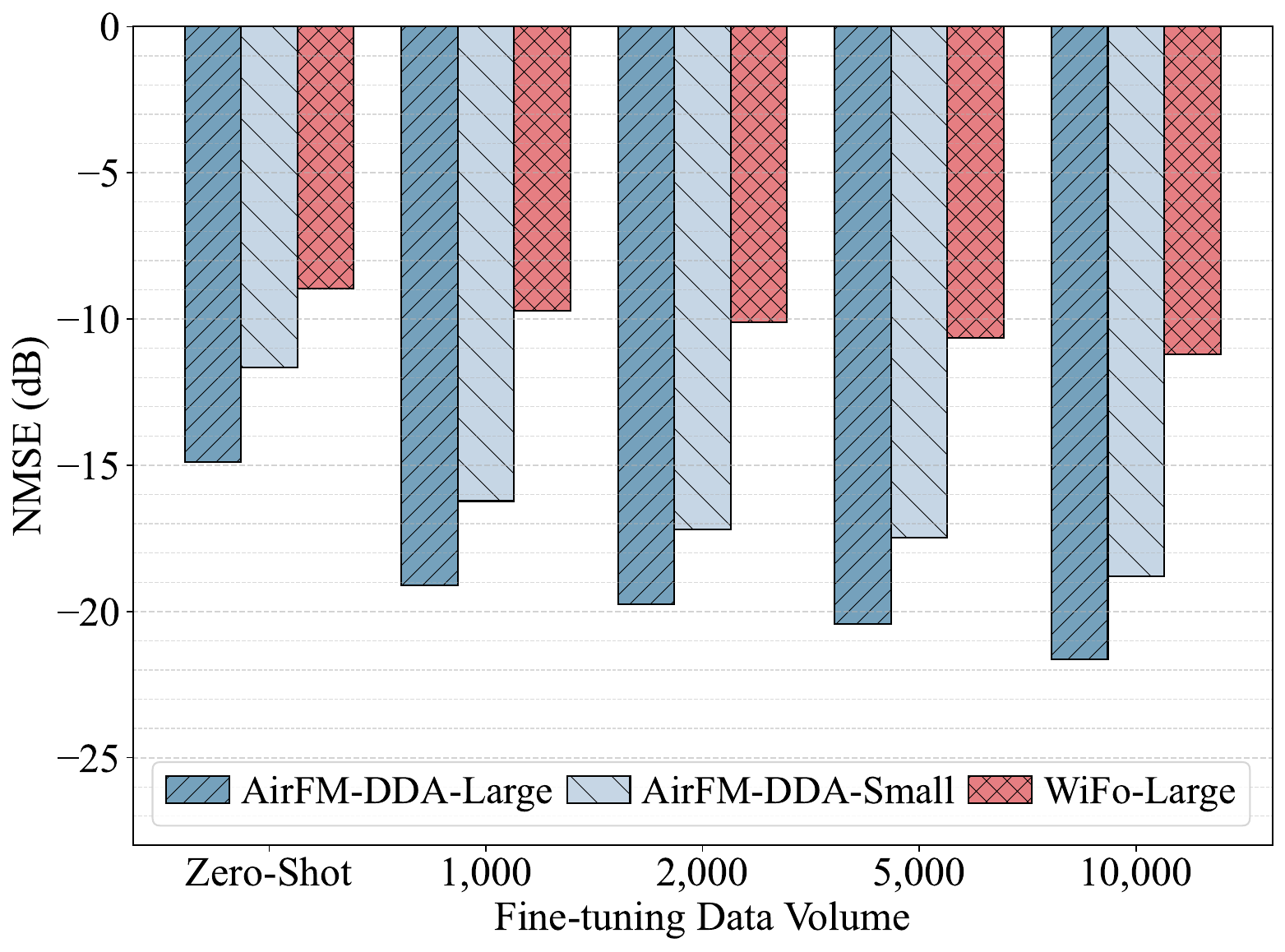}
        \caption{TP performance with $x=0.5$.}
        \label{fig:wair_d_tp_sub}
    \end{subfigure}
    \hfill
    \begin{subfigure}[t]{0.47\linewidth}
        \centering
\includegraphics[width=0.90\linewidth]{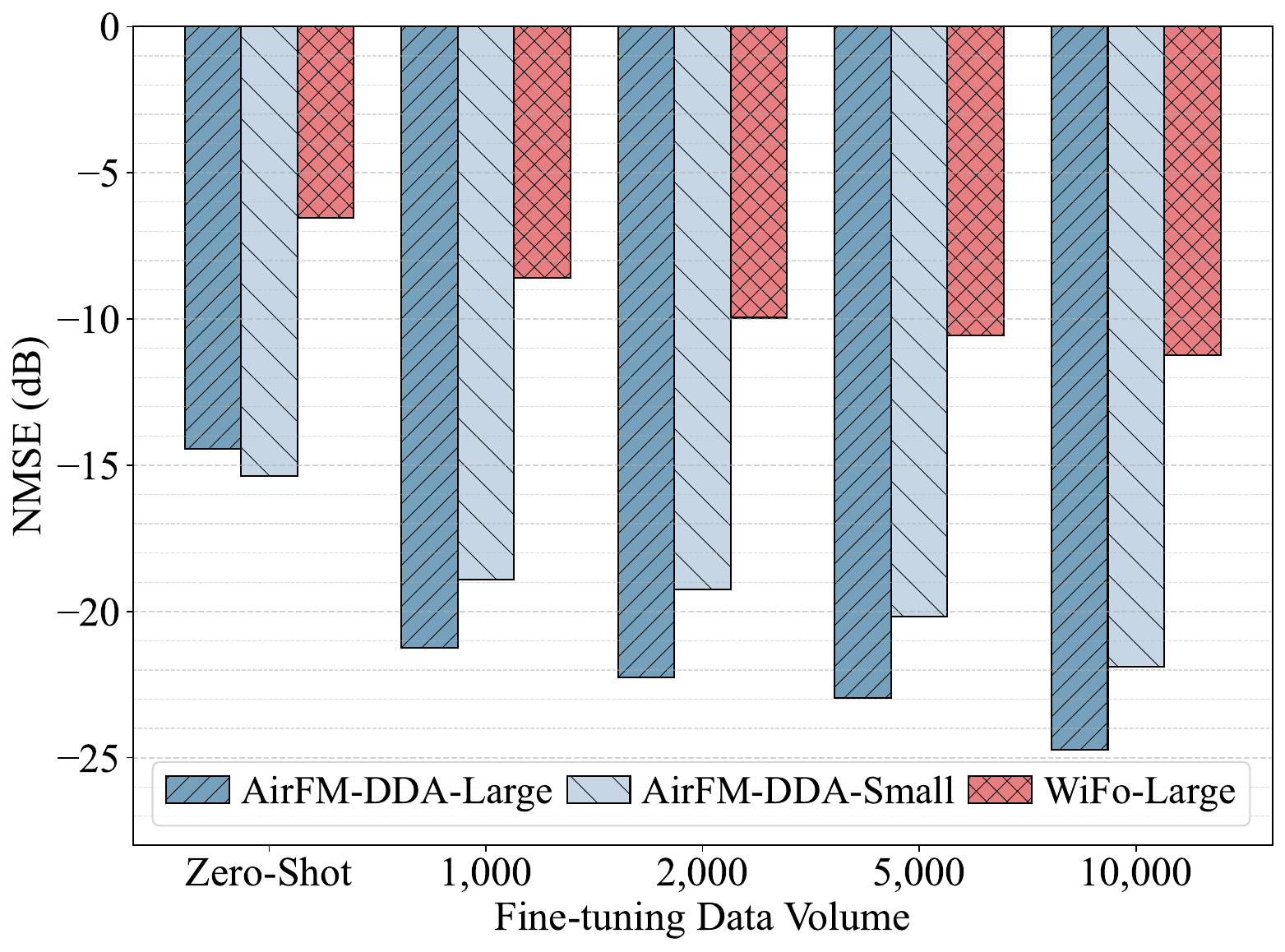}
        \caption{FP performance with $x=0.5$.}
        \label{fig:wair_d_fp_sub}
    \end{subfigure}

    \vspace{-0.15cm}
    \caption{NMSE performance on the WAIR-D-based dataset.}
    \label{fig:WAIR-D Generalization}

\vspace{-0.35cm}
\end{figure*}

\subsubsection{Backbone Architecture}
To assess the contributions of window-based attention and FS-PE, we conduct two ablation studies on the validation set: first, we replace W-MSA/SW-MSA with global self-attention, making the model equivalent to a vanilla ViT; second, we remove the FS-PE. As summarized in Table~\ref{table:backbone_ablation_efficiency}, the window-based attention backbone (AirFM-DDA w/o FS-PE) outperforms the global-attention variant (AirFM-DDA w/ ViT) by 3.42--4.76 dB across the TP, FP, and CE tasks, confirming the strong alignment between windowed attention and the localized multipath structures in the DDA domain. Furthermore, incorporating FS-PE brings an additional gain of around 1.95 dB, which validates the benefit of explicitly embedding frame-structure priors. Table~\ref{table:backbone_ablation_efficiency} also reports the computational efficiency of these backbone architectures. Compared with the global-attention variant, the window-based attention backbone slashes training costs by an order of magnitude and accelerates inference nearly eightfold, all while maintaining an identical parameter budget of 140.52~M. Moreover, the FS-PE incurs nearly zero overhead in terms of parameters and FLOPs, offering a highly cost-effective performance enhancement.

\subsubsection{Backbone Scaling}
We next investigate the impact of backbone scaling on task performance by evaluating the three model variants detailed in Table~\ref{table:Swin_Scaling}. To ensure a fair comparison, all variants are trained using the same strategy outlined in Table~\ref{table: network optimization}. Fig.~\ref{fig:backbone_scale_ablation_sub} presents the validation-set performance of AirFM-DDA-Small, -Base, and -Large across TP, FP, and CE tasks. As observed, performance consistently improves as the backbone parameter count increases, yielding an average performance gap of approximately 3.75~dB between the largest and smallest variants across all tasks. This monotonic improvement underscores the strong scalability of the AirFM-DDA architecture and confirms that increased model capacity directly translates into superior representation capability.

\subsubsection{Pre-training Data Size}
Beyond model dimensions, the scale of pre-training data provides another axis for examining empirical scaling trends. To systematically investigate this, we vary the data budget by uniformly subsampling the complete pre-training dataset at ratios of 2\%, 10\%, 20\%, 50\%, and 100\%, while maintaining a consistent training strategy. Fig.~\ref{fig:pretraining_data_ablation_sub} illustrates the validation-set performance of AirFM-DDA-Large on TP, FP, and CE tasks when trained with these varying fractions of pre-training data. The results indicate that the model's representation capability improves rapidly with the expansion of pre-training data volume. Specifically, the average performance gap between models trained on 2\% and 100\% of the data is approximately 6.61~dB. These findings highlight the importance of large-scale pre-training for realizing the backbone's representation potential.
\subsection{Task-Level Performance Evaluation}

{We evaluate zero-shot CSI reconstruction on TP, FP, and CE and few-shot adaptation on BP and LoS identification.}

\begin{revisionblue}
\subsubsection{Zero-Shot CSI Reconstruction}
Zero-shot evaluation refers to directly applying the pre-trained AirFM-DDA to a target test domain without using any target-domain training samples for fine-tuning. Specifically, AirFM-DDA is pre-trained on DeepMIMO cities 0--17 and directly evaluated on four test city layouts that are completely excluded from pre-training and summarized in Table~\ref{table: Dataset Settings}: Denver (18), Oklahoma City (19), Beijing (23), and Rio de Janeiro (27). Under this strict protocol, AirFM-DDA is evaluated on the TP, FP, and CE tasks. For TP and FP, WiFo is evaluated under the same zero-shot setting and therefore serves as the direct zero-shot baseline. ARMA is included as an observation-only conventional reference for TP, whereas LMMSE is included as a target-calibrated conventional reference for FP and CE. To provide additional performance references, LLM4WM~\cite{liu2025llm4wm}, Transformer~\cite{jiang2022transformer}, and U-Net are fully trained on the corresponding target scenarios; their results are thus regarded as full-shot, in-distribution references rather than zero-shot~comparisons.

For the TP task, Figs.~\ref{fig:tp_x075_sub} and~\ref{fig:tp_x05_sub} evaluate future-CSI prediction over user speeds from 0 to 120~km/h at $x=0.75$ and $x=0.5$, respectively. All AirFM-DDA variants outperform the baselines over nearly all speed bins and retain a clear margin in the most difficult regime of 100--120~km/h with only half of the temporal samples observed. In contrast, the observation-only ARMA predictor deteriorates rapidly with mobility, while WiFo-Large, the only directly comparable zero-shot learned baseline, remains substantially weaker. Even AirFM-DDA-Small, with 62.62 million parameters, is competitive with or superior to the full-shot references, showing that the gain is not explained by model size alone. This behavior supports the DDA-domain design: temporal truncation appears as Doppler-domain spreading, and locality-aligned window attention recovers clustered Doppler--angle structures across unseen layouts. Scaling further strengthens this representation, with AirFM-DDA-Large reaching an average NMSE of $-17.84$~dB.

For the FP task, Figs.~\ref{fig:fp_x075_sub} and~\ref{fig:fp_x05_sub} stratify the unseen-city samples by RMS delay spread and evaluate two observation ratios. The performance gap becomes most informative as propagation grows more dispersive. In the 300--1000~ns regime, the learned baselines degrade to approximately $-8$ to $-2$~dB, while the target-calibrated LMMSE reference reaches $-4.53$~dB at $x=0.75$ and $-2.38$~dB at $x=0.5$; every AirFM-DDA variant preserves an NMSE below $-15$~dB. The advantage also persists when the observed bandwidth is reduced from $x=0.75$ to $x=0.5$.

These results support the central motivation: frequency truncation induces delay-domain spreading, while the DDA representation exposes multipath delays as sparse local clusters rather than entangled STF-domain superpositions. Window-based attention captures this local structure, while shifted windows preserve continuity across partition boundaries. The consistent zero-shot gains over WiFo and full-shot references demonstrate robust generalization to unseen scenarios.

For the CE task, Figs.~\ref{fig:ce_d2_sub} and~\ref{fig:ce_d4_sub} evaluate reconstruction from comb-type pilots at SNRs of 5--20~dB with $D=2$ and $4$. AirFM-DDA achieves the lowest NMSE for every SNR and pilot spacing. Even AirFM-DDA-Small averages about $-17.5$~dB for $D=2$ and $-14.5$~dB for $D=4$, providing an average gain of about 4.7~dB over the strongest baseline. The sparser pilot pattern increases the error, yet the AirFM-DDA variants vary by at most 1.6~dB across the SNR range, indicating limited noise sensitivity. This comparison requires care: LLM4WM and U-Net are full-shot references and LMMSE uses target-domain calibration, whereas AirFM-DDA uses no target-domain samples. The result supports the DDA formulation: comb pilots induce periodic delay-bin superposition, turning CE into structured de-aliasing. By exposing this geometry and encoding frame-dependent resolutions via FS-PE, AirFM-DDA reuses a pre-trained physical representation rather than learning scenario-specific interpolation.

Overall, the AirFM-DDA family achieves the best reconstruction performance across TP, FP, and CE and exhibits strong robustness to variations in user mobility, temporal dispersion, and noise levels. Across these tasks, it achieves average NMSE gains of 4.9--8.5~dB over the strongest baselines. These results demonstrate that the DDA domain provides a promising CSI representation for wireless-native FMs and verify the effectiveness of the proposed backbone in capturing wireless propagation patterns.
\end{revisionblue}

\begin{revisionblue}
\subsubsection{Few-Shot Adaptation for BP and LoS Identification}
We next evaluate downstream adaptation on BP and LoS identification under the settings in Section~\ref{subsec:downstream_setup}. The shared encoder is initialized with the reconstruction-pretrained weights and kept frozen, while only the corresponding lightweight task-specific head is updated.

Table~\ref{tab:bp_los_adaptation} shows that AirFM-DDA consistently achieves the best few-shot adaptation performance on both downstream tasks. With only 10\% labeled data, the three AirFM-DDA variants achieve an average Top-1 accuracy of 61.6\% for BP and an average F1 score of 95.3\% for LoS identification, outperforming the strongest baselines by 12.0 and 3.4 percentage points, respectively. For LoS identification, AirFM-DDA-Large reaches an F1 score of 96.39\% with only 10\% labeled data and improves to 97.59\% with the full training split, outperforming the strongest baseline, LWM, by 4.53 and 3.35 percentage points, respectively. The advantage is more pronounced for BP, where accurate beam selection requires fine-grained geometric discrimination. AirFM-DDA-Large achieves 65.95\% Top-1 accuracy with 10\% labeled data and 82.82\% with the full training split, exceeding LLM4WM by 16.33 and 16.44 percentage points, respectively. Moreover, all AirFM-DDA variants outperform the task-specific DeepBP and ST-CNN baselines, demonstrating the transferability of reconstruction-oriented DDA-domain pre-training.
\end{revisionblue}

\begin{figure*}[!t]
    \centering
    \vspace{-0.15cm}
    \begin{minipage}[t]{0.485\textwidth}
    \color{blue}
    \vspace{0pt}
    \centering
    \captionof{table}{Zero-shot NMSE (dB) on the LAMBDA dataset at an SNR of 10~dB.}
    \label{tab:lambda_zero_shot_full}
    \footnotesize
    \setlength{\tabcolsep}{2.5pt}
    \renewcommand{\arraystretch}{1.08}
    \begin{tabular}{lcccc}
    \toprule
    Task & \makecell{WiFo\\-Large} & \makecell{AirFM-DDA\\-Small} & \makecell{AirFM-DDA\\-Base} & \makecell{AirFM-DDA\\-Large} \\
    \midrule
TP ($x=0.5$) & -14.46 & -18.25 & -18.70 & \textbf{-19.19} \\
    FP ($x=0.5$) & -8.38  & -17.92 & -18.54 & \textbf{-19.48} \\
    \bottomrule
    \end{tabular}
    \end{minipage}
    \hfill
    \begin{minipage}[t]{0.485\textwidth}
    \color{blue}
    \vspace{0pt}
    \centering
    \captionof{table}{Fine-tuned NMSE (dB) on the LuViRA evaluation set.}
    \label{tab:luvira_transfer_full}
    \footnotesize
    \setlength{\tabcolsep}{2.5pt}
    \renewcommand{\arraystretch}{1.08}
    \begin{tabular}{lcccc}
    \toprule
    Task & \makecell{WiFo\\-Large} & \makecell{AirFM-DDA\\-Small} & \makecell{AirFM-DDA\\-Base} & \makecell{AirFM-DDA\\-Large} \\
    \midrule
    TP ($x=0.5$) & -7.17 & -8.99  & -9.38  & \textbf{-10.00} \\
    FP ($x=0.5$) & -7.45 & -10.09 & -10.35 & \textbf{-11.11} \\
    \bottomrule
    \end{tabular}
    \end{minipage}
\vspace{0.2cm}
\end{figure*}

{The above results evaluate AirFM-DDA across both zero-shot CSI reconstruction and few-shot downstream adaptation. Since CSI reconstruction is the pre-training objective and directly probes the quality of the DDA-domain channel representation, the following subsections further examine computational efficiency, cross-dataset generalization, and antenna-array configuration adaptation.}

\begin{revisionblue}
\subsection{Efficiency Evaluation}

Table~\ref{tab:response_efficiency_reviewerone} compares computational efficiency under the channel-prediction setting with an observation ratio $x$ of $0.5$. The comparison mainly focuses on AirFM-DDA and WiFo, as both are wireless-native FMs and are configured with the same sequence length of 10,240 patches. Although WiFo discards masked patches before encoding, its global-attention backbone still incurs substantial overhead. In contrast, AirFM-DDA processes the complete DDA grid using locality-aligned window attention. AirFM-DDA-Small requires only $657.02$~GFLOPs, $25.06$~ms inference latency, and $1.44$~GiB peak memory, compared with $1550$~GFLOPs, $253.21$~ms, and $13.66$~GiB for WiFo-Large. Even AirFM-DDA-Large, despite having more parameters, reduces latency by nearly eightfold and memory consumption by about sevenfold, while retaining slightly lower FLOPs. These results demonstrate that window-based attention provides a markedly more efficient foundation-model backbone than the global-attention design. Notably, the reported AirFM-DDA costs already include FFT preprocessing and IFFT postprocessing.
\end{revisionblue}

\begin{table*}[!t]
\centering
\caption{NMSE (dB) comparison of AirFM-DDA-Large and WiFo-Large under different antenna-array configurations. 
For AirFM-DDA-Large, results before and after applying the shared LoRA adapter are reported separately.}
\label{tab:upa_adaptation}

\footnotesize
\setlength{\tabcolsep}{3.0pt}
\renewcommand{\arraystretch}{1.08}

\begin{tabular}{ccccccc}
\toprule
\multirow{2}{*}[-1.0ex]{Antenna Array}
& \multicolumn{3}{c}{TP ($x=0.5$)}
& \multicolumn{3}{c}{FP ($x=0.5$)} \\
\cmidrule(lr){2-4}
\cmidrule(lr){5-7}

& \makecell{WiFo-Large}
& \makecell{AirFM-DDA-Large \\ w/o LoRA}
& \makecell{AirFM-DDA-Large \\ w/ LoRA}
& \makecell{WiFo-Large}
& \makecell{AirFM-DDA-Large \\ w/o LoRA}
& \makecell{AirFM-DDA-Large \\ w/ LoRA} \\
\midrule

$4\mathrm{H}\times4\mathrm{V}$
& -10.68
& -14.20
& \textbf{-16.62}
& -11.78
& -11.33
& \textbf{-17.91} \\

$16\mathrm{H}\times2\mathrm{V}$
& -11.36
& -13.20
& \textbf{-16.43}
& -12.65
& -10.75
& \textbf{-17.28} \\

$8\mathrm{H}\times6\mathrm{V}$
& -8.13
& -13.12
& \textbf{-15.23}
& -7.54
& -10.62
& \textbf{-16.43} \\

\bottomrule
\end{tabular}

\vspace{-0.5cm}
\end{table*}

\begin{revisionblue}
\subsection{Cross-Dataset Generalization Evaluation}

To assess whether the representations learned from the DeepMIMO-based pre-training corpus can generalize beyond the source distribution, we conduct cross-dataset evaluations on two additional simulated channel datasets, WAIR-D \cite{huangfu2022wair} and LAMBDA \cite{zhou2026lambda}, as well as the real-world measured LuViRA dataset \cite{yaman2024luvira}. Such cross-distribution generalization is also a key challenge in wireless signal recognition under varying channel conditions \cite{li2026open}. These experiments jointly examine zero-shot simulation-to-simulation transfer, limited-data adaptation, and simulation-to-measurement transfer.

\subsubsection{Transfer to Other Simulated Datasets}

The WAIR-D-based dataset~\cite{huangfu2022wair} contains 100,000 channel samples generated from 1,000 environments in Scenario~1 at 2.6~GHz. It uses a 30~kHz base subcarrier spacing, a nominal bandwidth of 98.28~MHz, and 64 frequency points obtained by sampling every 48 subcarriers. The receiver employs an $8\mathrm{H}\times4\mathrm{V}$ UPA, while each sample contains 40 time slots separated by 0.5~ms, with user velocities uniformly distributed between 0 and 120~km/h. To prevent environment leakage, 900 environments containing 90,000 samples are used for testing, whereas the remaining 100 environments are used for fine-tuning. AirFM-DDA and WiFo-Large are evaluated under zero-shot transfer and after fine-tuning with 1,000, 2,000, 5,000, and 10,000 target-domain samples. As shown in Fig.~\ref{fig:WAIR-D Generalization}, the two AirFM-DDA variants average approximately $-13$~dB and $-15$~dB NMSE on TP and FP, respectively, without target-domain fine-tuning. Fine-tuning with only 1,000 samples further improves TP and FP by 4.38~dB and 5.17~dB on average, demonstrating effective adaptation to previously unseen simulated channel distributions.

The LAMBDA dataset~\cite{zhou2026lambda} contains 29,407 validated path-parameter samples at 3.5~GHz, each comprising 25 valid propagation paths. We use the Urban Area--Block, Scene~1, sunny-weather configuration with one BS and one UAV following the predefined Z-shaped trajectory (\texttt{1\_bs\_1\_uav\_z\_traj}). Its receive-array configuration, frequency sampling scheme, number of frequency points, and temporal settings are consistent with those used for WAIR-D, while a $1\mathrm{H}\times1\mathrm{V}$ transmit array is employed. All samples are reserved exclusively for strict zero-shot evaluation. For both TP and FP, the first 50\% of the corresponding axis is observed, and the remaining portion is predicted at an SNR of 10~dB. As reported in Table~\ref{tab:lambda_zero_shot_full}, all AirFM-DDA variants substantially outperform WiFo-Large. AirFM-DDA-Large achieves $-19.19$~dB on TP and $-19.48$~dB on FP, corresponding to improvements of 4.73~dB and 11.10~dB, respectively. These results confirm that the learned representation can be directly transferred to an independently constructed ray-tracing dataset without access to target-domain training samples.

\subsubsection{Transfer to Measured Channels}

The LuViRA dataset \cite{yaman2024luvira} is a publicly available real-world multimodal dataset collected indoors using synchronized vision, radio, and audio sensors. Its radio measurements were acquired using a 5G massive-MIMO testbed over 89 robot trajectories. We use its measured CSI at 3.7 GHz to evaluate transfer from the DeepMIMO-based simulated pre-training corpus to real-world measured channels. The original 15 kHz OFDM grid is sampled every 12 subcarriers, yielding an effective spacing of 180 kHz; the central 64 frequency points and $8\mathrm{H}\times4\mathrm{V}$ receive subarray are retained. Each sample contains 40 snapshots separated by 10 ms. After excluding two predefined trajectories, the remaining 87 trajectories yield 9,472 non-overlapping windows. To avoid trajectory leakage, 17 trajectories provide 1,650 training and 244 validation windows, while the other 70 are held out for evaluation. As reported in Table~\ref{tab:luvira_transfer_full}, AirFM-DDA-Large achieves the best TP and FP performance, reaching $-10.00$~dB and $-11.11$~dB, respectively. It outperforms WiFo-Large by 2.83~dB on TP and 3.66~dB on FP. 

Together with the cross-scenario results in Figs.~\ref{fig:TP Task Performance}--\ref{fig:CE Task Performance}, these experiments establish two complementary levels of zero-shot generalization: transfer from 18 pre-training cities to four unseen layouts without target-domain fine-tuning, and direct evaluation of the same pre-trained model on the independently constructed WAIR-D and LAMBDA datasets. The LuViRA results further demonstrate efficient simulation-to-measurement adaptation with limited target data. Thus, AirFM-DDA is not tied to specific training environments or dataset distributions.
\end{revisionblue}

\begin{revisionblue}
\subsection{Adaptation to Different UPA Configurations}

Although AirFM-DDA is pre-trained with a specific UPA configuration, the proposed framework can be adapted to antenna arrays with different sizes and geometries. For each target geometry, we construct a geometry-specific complex principal component analysis (PCA) interface~\cite{jolliffe2016principal}. Specifically, PCA is fitted to the native DDA angular coefficients and projects them onto the principal subspace spanned by the dominant eigenvectors of their complex covariance matrix, thereby mapping different antenna dimensions to the backbone's fixed 32-complex-channel interface while preserving the dominant channel variations. The corresponding PCA decoder projects the backbone output back to the native angular space. To compensate for the semantic shift introduced by this mapping, a single shared low-rank adaptation (LoRA) adapter is jointly trained across different array geometries while the original AirFM-DDA backbone remains frozen. Specifically, LoRA is inserted into the query, key, and value projections, the attention-output projection, and the two linear layers of the MLP in all window-based attention blocks.

To evaluate this adaptation strategy, we use AirFM-DDA-Large as a representative backbone and consider the $4\mathrm{H}\times4\mathrm{V}$, $16\mathrm{H}\times2\mathrm{V}$, and $8\mathrm{H}\times6\mathrm{V}$ UPA configurations. We evaluate TP and FP with $x=0.5$ on the test set and compare the adapted AirFM-DDA-Large with WiFo. As shown in Table XI, AirFM-DDA-Large can be effectively adapted to these antenna-array configurations and achieves competitive or superior performance over WiFo. Moreover, the shared LoRA adapter consistently provides substantial performance gains across all evaluated array configurations.

\end{revisionblue}

\FloatBarrier

\section{Conclusion}


{In this paper, we proposed AirFM-DDA, a wireless-native FM that maps multipath-superimposed STF-domain CSI to the resolvable DDA domain. Its window-based backbone captures local multipath structures with near-linear attention complexity, while FS-PE accommodates heterogeneous frame configurations. Experiments demonstrate strong zero-shot channel reconstruction, few-shot downstream adaptation, cross-dataset and simulation-to-measurement transfer, and adaptation to different antenna-array configurations. Future work will extend AirFM-DDA to broader downstream tasks and over-the-air deployments while further improving its efficiency.}

\begingroup
\renewcommand{\IEEEbibitemsep}{0pt plus .5pt}
\apptocmd{\thebibliography}{\setlength{\baselineskip}{1.05\baselineskip}}
  {}{\PackageError{AirFM-DDA}{Reference spacing patch failed}{}}

\par
\endgroup

\begingroup
\raggedbottom
\makeatletter
\def\@IEEEBIOskipN{0.75\baselineskip}
\expandafter\patchcmd\csname\string\IEEEbiography\endcsname
  {\vskip \@IEEEBIOskipN plus 1fil minus 0\baselineskip}
  {\vskip \@IEEEBIOskipN}
  {}{\PackageError{AirFM-DDA}{Biography spacing patch failed}{}}
\makeatother

\begin{IEEEbiography}[{\includegraphics[width=1in,height=1.25in,keepaspectratio]{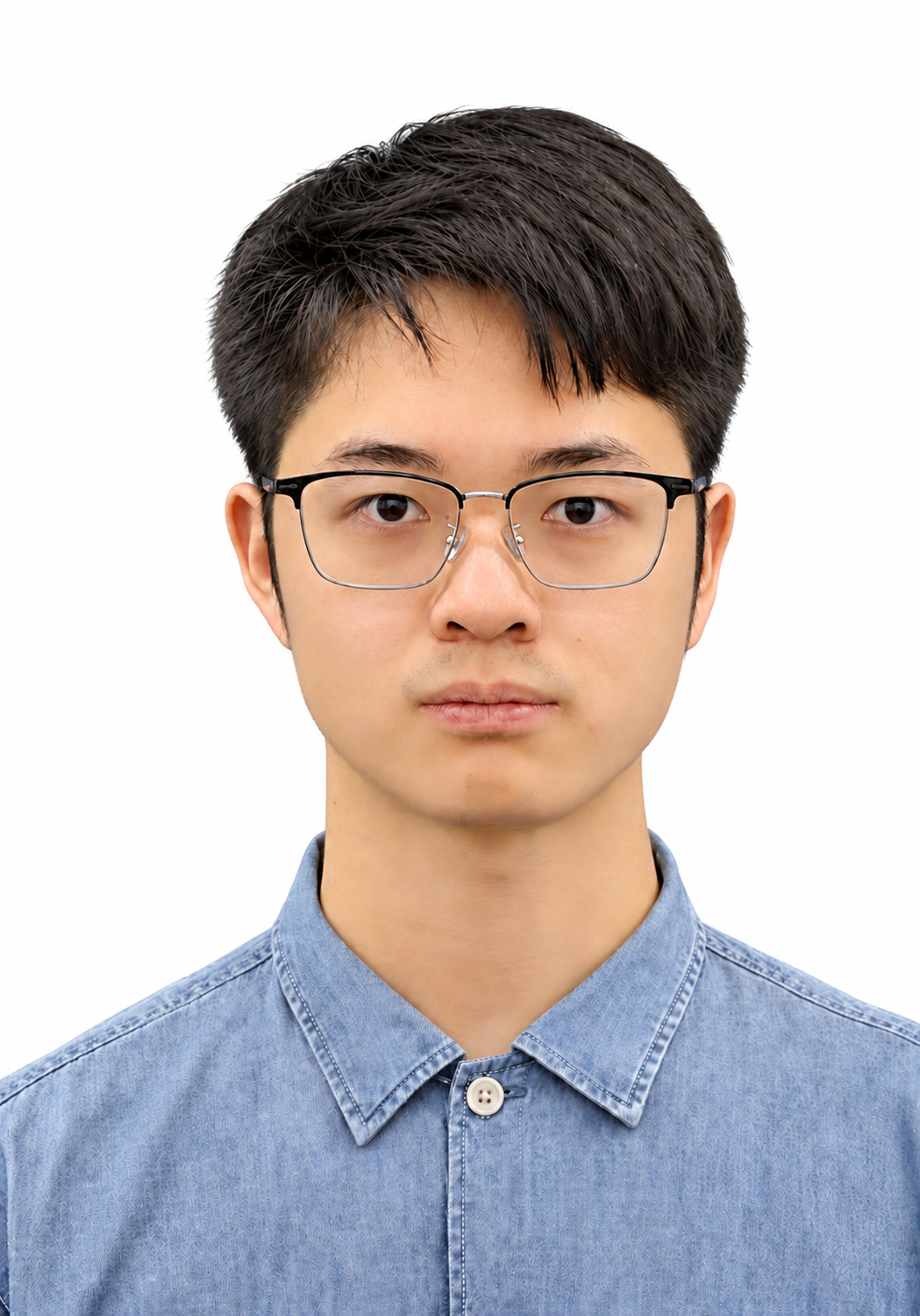}}]{Kejia Bian}
received the B.S. degree from Xi'an Jiaotong University, Xi'an, China, in 2024. He is currently pursuing the Ph.D. degree at Shanghai Jiao Tong University, Shanghai, China. His research interests include wireless foundation models and intelligent channel modeling.
\end{IEEEbiography}

\begin{IEEEbiography}[{\includegraphics[width=1in,height=1.25in,keepaspectratio]{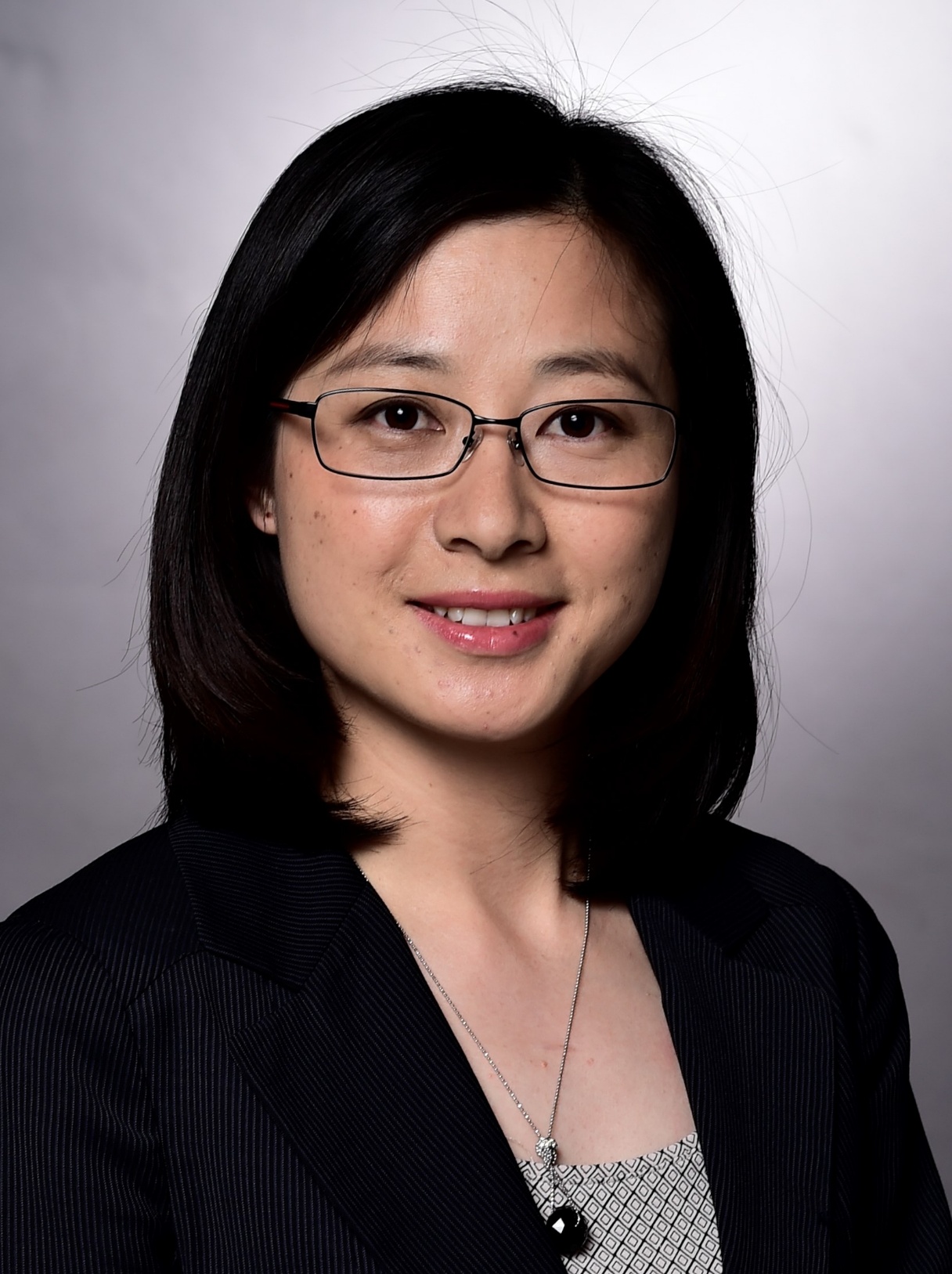}}]{Meixia Tao}
(Fellow, IEEE) is a Professor with the School of Information Science and Electronic Engineering at Shanghai Jiao Tong University, China. She received the B.S. degree in electronic engineering from Fudan University, Shanghai, China, in 1999, and the Ph.D. degree in electrical and electronic engineering from the Hong Kong University of Science and Technology in 2003. Her current research interests include wireless edge learning, semantic communications, integrated communication-computing-sensing, and AI-based channel modeling and beamforming. She has published over 300 peer-reviewed research papers in the related field. Prof. Tao has received several awards and recognitions, including the 2020 First Prize in Natural Science from the Shanghai Municipality, China, the 2019 IEEE Marconi Prize Paper Award, and the 2013 IEEE Heinrich Hertz Award for Best Communications Letters. Prof. Tao is currently a Vice-Chair of the Information Theory Society of the Chinese Institute of Electronics and a member of the IEEE ComSoc GITC committee. She was on the Steering Committee of IEEE Transactions on Wireless Communications during 2022-2025. Previously, she served on the editorial boards of a number of journals, including IEEE Transactions on Wireless Communications, IEEE Transactions on Communications, and IEEE Transactions on Information Theory. She also served as the TPC Co-Chair of IEEE ICC 2023 and IEEE WCNC 2026.
\end{IEEEbiography}

\begin{IEEEbiography}[{\includegraphics[width=1in,height=1.25in,clip,keepaspectratio]{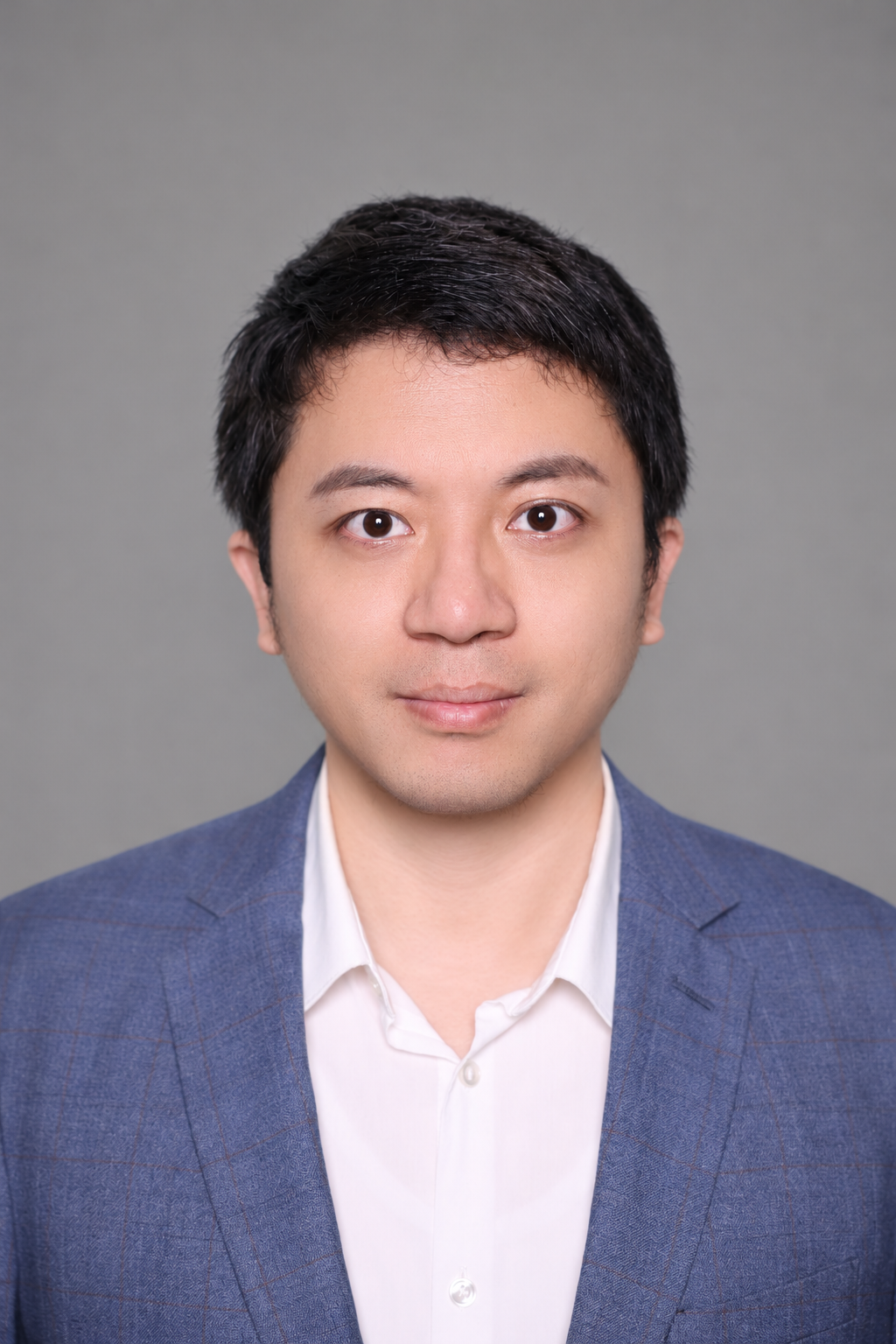}}]{Jianhua Mo}
(Senior Member, IEEE) received the B.S. and M.S. degrees in electronic engineering from Shanghai Jiao Tong University in 2010 and 2013, respectively, and the Ph.D. degree in electrical and computer engineering from The University of Texas at Austin in 2017. Since December 2024, he has been a tenure-track Associate Professor with Shanghai Jiao Tong University. Previously, he was a Senior Staff Engineer with Samsung Research America, Plano, TX, USA, where his patents on mmWave beam management have been adopted in commercial mmWave systems. He has filed over 60 international patent applications, with 50 patents granted. His research interests include MIMO communications with low-resolution ADCs, mmWave beam management, phase-time arrays and integrated sensing and communication. His awards and honors include the Stephen O. Rice Prize in 2019, the Heinrich Hertz Award in 2013, Samsung Outstanding Inventor Award in 2023, Samsung Excellent Achievement Award in 2021, the Best Wi-Fi Innovation Award by Wireless Broadband Alliance (WBA) in 2019, and the Finalist for Qualcomm Innovation Fellowship in 2014.
\end{IEEEbiography}

\begin{IEEEbiography}[{\includegraphics[width=1in,height=1.25in,keepaspectratio]{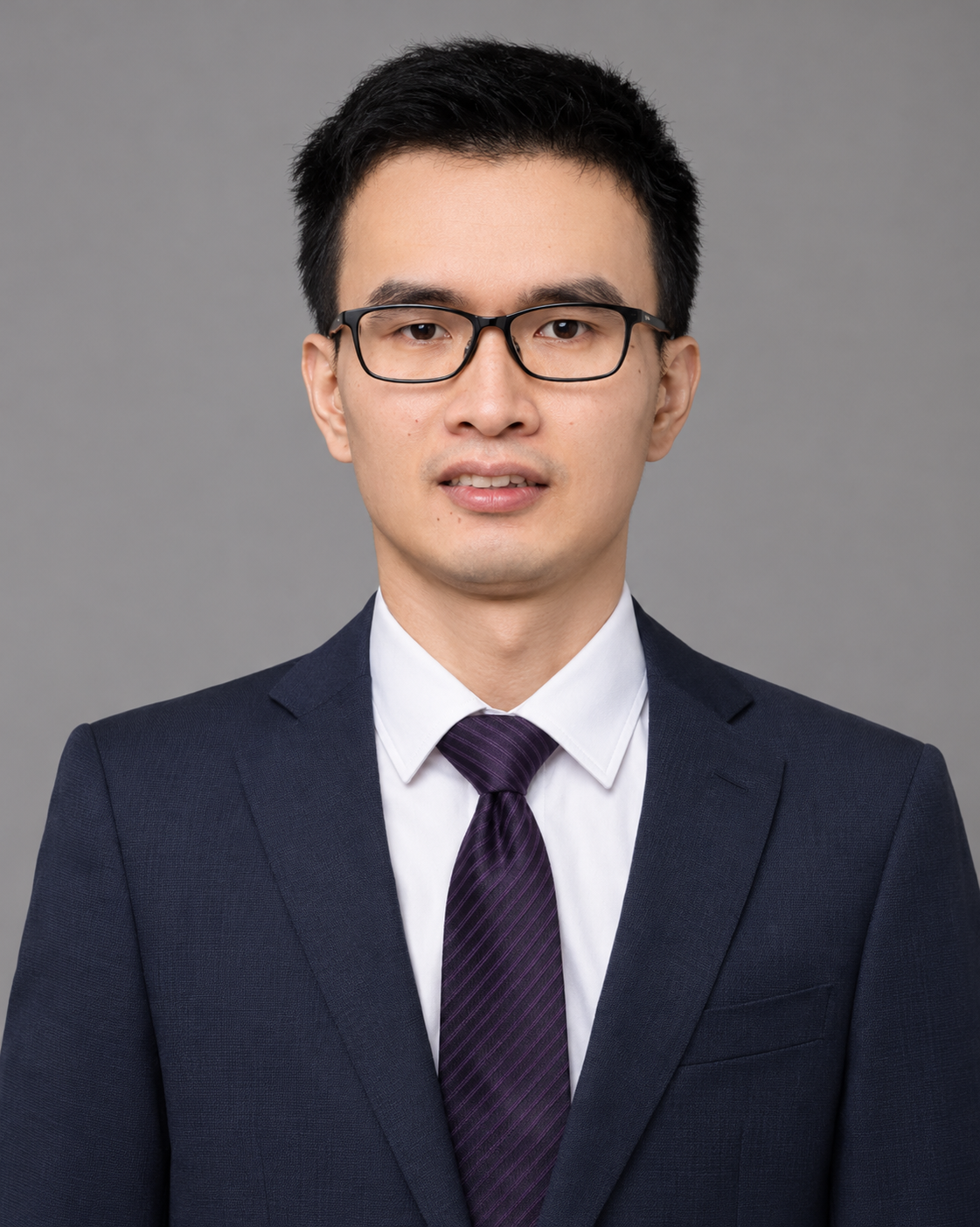}}]{Zhiyong Chen} (Senior Member, IEEE)
received the Ph.D. degree from the School of Information and Communication Engineering, Beijing University of Posts and Telecommunications (BUPT), Beijing, China, in 2011. From 2009 to 2011, he was a visiting Ph.D. student at the Department of Electronic Engineering, University of Washington, Seattle, USA. He is currently a Professor with the Cooperative Medianet Innovation Center, Shanghai Jiao Tong University (SJTU), Shanghai, China. His research interests include mobile communications-computing-caching (3C) networks and mobile AI systems. He served as the Student Volunteer Chair for IEEE ICC 2019, the Publicity Chair for IEEE/CIC ICCC 2014, and a technical program committee (TPC) member for major international conferences. He received the IEEE Asia-Pacific Outstanding Paper Award in 2019.
\end{IEEEbiography}

\begin{IEEEbiography}[{\includegraphics[width=1in,height=1.25in,keepaspectratio]{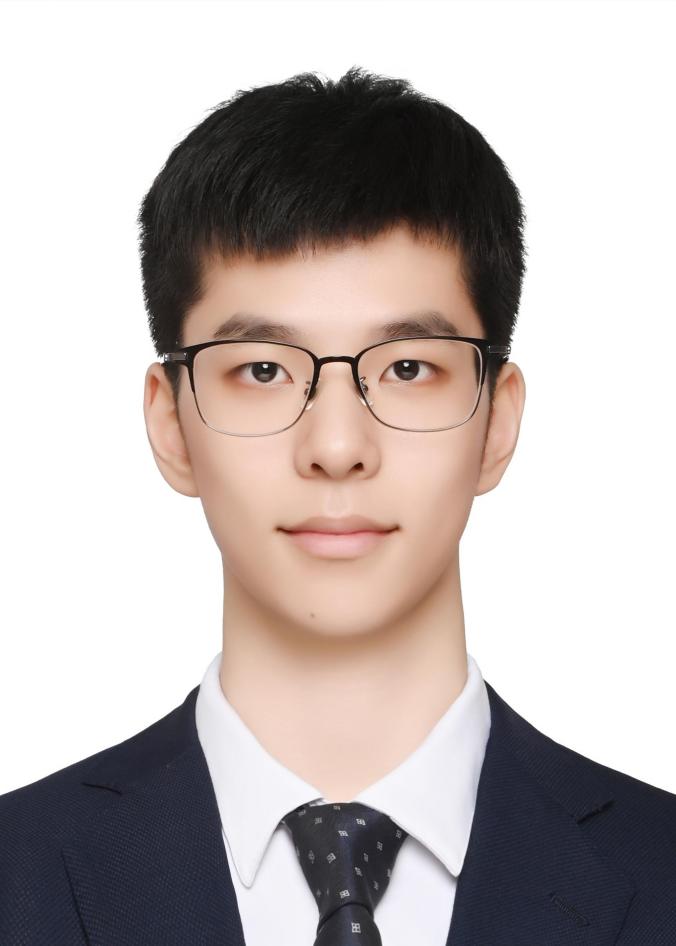}}]{Leyan Chen}
received the B.S. degree in communication engineering from Shanghai Jiao Tong University, Shanghai, China, in 2026. He is currently pursuing the M.S. degree in Information and Communication Engineering with the Department of Electronic Engineering, Shanghai Jiao Tong University, Shanghai, China. His research interests include semantic communication and wireless foundation models.
\end{IEEEbiography}

\par
\endgroup
\vfill


\begin{thebibliography}{10}
\providecommand{\url}[1]{#1}
\csname url@samestyle\endcsname
\providecommand{\newblock}{\relax}
\providecommand{\bibinfo}[2]{#2}
\providecommand{\BIBentrySTDinterwordspacing}{\spaceskip=0pt\relax}
\providecommand{\BIBentryALTinterwordstretchfactor}{4}
\providecommand{\BIBentryALTinterwordspacing}{\spaceskip=\fontdimen2\font plus
\BIBentryALTinterwordstretchfactor\fontdimen3\font minus
  \fontdimen4\font\relax}
\providecommand{\BIBforeignlanguage}[2]{{%
\expandafter\ifx\csname l@#1\endcsname\relax
\typeout{** WARNING: IEEEtran.bst: No hyphenation pattern has been}%
\typeout{** loaded for the language `#1'. Using the pattern for}%
\typeout{** the default language instead.}%
\else
\language=\csname l@#1\endcsname
\fi
#2}}
\providecommand{\BIBdecl}{\relax}
\BIBdecl

\bibitem{conferenceWork}
K.~Bian \emph{et~al.}, ``{AirMIND}: Air-interface foundation model in the
  delay–{Doppler}–angle domain,'' in \emph{Proc. IEEE Int. Conf. Commun.
  Workshops (ICC Workshops)}, 2026, pp. 1--6.

\bibitem{bommasani2021opportunities}
R.~Bommasani \emph{et~al.}, ``On the opportunities and risks of foundation
  models,'' \emph{arXiv preprint arXiv:2108.07258}, 2021.

\bibitem{radford2019language}
A.~Radford \emph{et~al.}, ``Language models are unsupervised multitask
  learners,'' OpenAI, Technical Report, 2019.

\bibitem{kirillov2023segment}
A.~Kirillov \emph{et~al.}, ``Segment anything,'' in \emph{Proc. IEEE/CVF Int.
  Conf. Comput. Vis. (ICCV)}, 2023, pp. 4015--4026.

\bibitem{11180834}
L.~Yu \emph{et~al.}, ``{ChannelGPT}: A large model toward real-world channel
  foundation model for {6G} environment intelligence communication,''
  \emph{IEEE Commun. Mag.}, vol.~63, no.~10, pp. 68--74, 2025.

\bibitem{jiang2025towards}
J.~Jiang \emph{et~al.}, ``Towards channel foundation models ({CFMs}):
  Motivations, methodologies and opportunities,'' \emph{arXiv preprint
  arXiv:2507.13637}, 2025.

\bibitem{chen2025towards}
Z.~Chen \emph{et~al.}, ``Towards wireless native big {AI} model: the mission
  and approach differ from large language model,'' \emph{Sci. China Inf. Sci.},
  vol.~68, no.~7, 2025.

\bibitem{jiang2026comprehensive}
F.~Jiang \emph{et~al.}, ``A comprehensive survey of large {AI} models for
  future communications: Foundations, applications, and challenges,''
  \emph{IEEE Commun. Surv. Tutor.}, vol.~28, pp. 4731--4764, 2026.

\bibitem{11074348}
X.~Cheng \emph{et~al.}, ``Foundation model empowered synesthesia of machines
  ({SoM}): {AI}-native intelligent multi-modal sensing-communication
  integration,'' \emph{IEEE Trans. Netw. Sci. Eng.}, vol.~13, pp. 762--782,
  2026.

\bibitem{11370176}
F.~Jiang \emph{et~al.}, ``From large {AI} models to agentic {AI}: A tutorial on
  future intelligent communications,'' \emph{IEEE J. Sel. Areas Commun.},
  vol.~44, pp. 3507--3540, 2026.

\bibitem{11449577}
L.~Liang \emph{et~al.}, ``Large language models for wireless communications:
  From adaptation to autonomy,'' \emph{IEEE Commun. Mag.}, vol.~64, no.~5, pp.
  140--147, May 2026.

\bibitem{liu2025llm4wm}
X.~Liu \emph{et~al.}, ``{LLM4WM}: Adapting {LLM} for wireless multi-tasking,''
  \emph{IEEE Trans. Mach. Learn. Commun. Networking}, vol.~3, pp. 835--847,
  2025.

\bibitem{zheng2025large}
T.~Zheng and L.~Dai, ``Large language model enabled multi-task physical layer
  network,'' \emph{IEEE Trans. Commun.}, vol.~74, pp. 307--321, 2026.

\bibitem{liu2024llm4cp}
B.~Liu \emph{et~al.}, ``{LLM4CP}: Adapting large language models for channel
  prediction,'' \emph{J. Commun. Inf. Netw.}, vol.~9, no.~2, pp. 113--125,
  2024.

\bibitem{xie20252dlam}
D.~Xie \emph{et~al.}, ``{2DLAM}: Joint {Delay-Doppler} estimation in {UAV}
  {mmWave} system via large {AI} model,'' in \emph{AAAI workshop on AI4WCN},
  Mar. 2025, pp. 1--9.

\bibitem{cui2025exploring}
Y.~Cui \emph{et~al.}, ``Exploring the potential of large language models for
  massive {MIMO} {CSI} feedback,'' in \emph{Proc. IEEE Global Commun. Conf.
  (GLOBECOM)}, Dec. 2025, pp. 3511--3516.

\bibitem{guo2025lvm4csi}
J.~Guo \emph{et~al.}, ``{LVM4CSI}: Enabling direct application of pre-trained
  large vision models for wireless channel tasks,'' \emph{arXiv preprint
  arXiv:2507.05121}, 2025.

\bibitem{alikhani2024large}
S.~Alikhani \emph{et~al.}, ``{LWM}: A pre-trained wireless foundation model for
  universal feature extraction,'' in \emph{Proc. IEEE Int. Conf. Mach. Learn.
  Commun. Netw. (ICMLCN)}, Barcelona, Spain, May 2025, pp. 1--6.

\bibitem{catak2025bert4mimo}
F.~O. Catak \emph{et~al.}, ``{BERT4MIMO}: A transformer-based model using
  {BERT} architecture for massive {MIMO} channel state information
  reconstruction,'' \emph{Digit. Signal Process.}, vol. 184, Nov. 2026, art.
  no. 106480.

\bibitem{liu2025wifo}
B.~Liu \emph{et~al.}, ``{WiFo}: Wireless foundation model for channel
  prediction,'' \emph{Sci. China Inf. Sci.}, vol.~68, no.~6, 2025.

\bibitem{yang2025wirelessgpt}
T.~Yang \emph{et~al.}, ``{WirelessGPT}: A generative foundation model for
  multi-task integrated sensing and communication,'' \emph{IEEE J. Sel. Areas
  Commun.}, vol.~44, pp. 2259--2273, 2026.

\bibitem{guler2025multi}
B.~Guler \emph{et~al.}, ``A multi-task foundation model for wireless channel
  representation using contrastive and masked autoencoder learning,''
  \emph{IEEE J. Sel. Areas Commun.}, vol.~44, pp. 4489--4504, 2026.

\bibitem{10971878}
T.~Jiao \emph{et~al.}, ``Addressing the curse of scenario and task
  generalization in {AI-6G}: A multi-modal paradigm,'' \emph{IEEE Trans.
  Wireless Commun.}, vol.~24, no.~9, pp. 7377--7391, 2025.

\bibitem{jiang2026csi}
J.~Jiang \emph{et~al.}, ``{CSI-MAE}: A masked autoencoder-based channel
  foundation model,'' \emph{arXiv preprint arXiv:2601.03789}, 2026.

\bibitem{sheng2025wireless}
Y.~Sheng \emph{et~al.}, ``A wireless foundation model for multi-task
  prediction,'' \emph{arXiv preprint arXiv:2507.05938}, 2025.

\bibitem{liu2025foundation}
B.~Liu \emph{et~al.}, ``{WiFo-2}: A generalist foundation model unifies
  heterogeneous wireless system design,'' \emph{Natl. Sci. Rev.}, vol.~13,
  no.~18, Sep. 2026.

\bibitem{zhang2026hetercsi}
C.~Zhang \emph{et~al.}, ``{HeterCSI}: Channel-adaptive heterogeneous {CSI}
  pretraining framework for generalized wireless foundation models,''
  \emph{arXiv preprint arXiv:2601.18200}, 2026.

\bibitem{wang2026consisformer}
Y.~Wang \emph{et~al.}, ``{ConsisFormer}: Compute-efficient transformer for
  wireless foundation models based on channel consistency,'' \emph{arXiv
  preprint arXiv:2606.19953}, 2026.

\bibitem{zhang2026adaptive}
C.~Zhang \emph{et~al.}, ``{Adaptive 3D-RoPE}: Physics-aligned rotary positional
  encoding for wireless foundation models,'' \emph{arXiv preprint
  arXiv:2605.00968}, 2026.

\bibitem{jiang2025mimo}
J.~Jiang \emph{et~al.}, ``A {MIMO} wireless channel foundation model via
  {CIR-CSI} consistency,'' in \emph{Proc. IEEE Int. Conf. Mach. Learn. Commun.
  Netw. (ICMLCN)}, 2025, pp. 1--6.

\bibitem{pan2025large}
G.~Pan \emph{et~al.}, ``Large wireless localization model ({LWLM}): A
  foundation model for positioning in {6G} networks,'' \emph{IEEE Trans.
  Wireless Commun.}, vol.~25, pp. 21\,449--21\,464, 2026.

\bibitem{wen2026icwlm}
Y.~Wen \emph{et~al.}, ``{ICWLM}: A multi-task wireless large model via
  in-context learning,'' \emph{IEEE Trans. Commun.}, vol.~74, pp. 3646--3658,
  2026.

\bibitem{zheng2025muse}
T.~Zheng \emph{et~al.}, ``{MUSE-FM}: Multi-task environment-aware foundation
  model for wireless communications,'' \emph{IEEE Trans. Wireless Commun.},
  vol.~25, pp. 19\,791--19\,806, 2026.

\bibitem{alkhateeb2019deepmimo}
A.~Alkhateeb, ``{DeepMIMO}: A generic deep learning dataset for millimeter wave
  and massive {MIMO} applications,'' in \emph{Proc. Inf. Theory Appl. Workshop
  (ITA)}, San Diego, CA, USA, Feb. 2019, pp. 1--8.

\bibitem{3gpp38901}
\BIBentryALTinterwordspacing
{3rd Generation Partnership Project (3GPP)}, ``{Study on channel model for
  frequencies from 0.5 to 100 {GHz}},'' {3rd Generation Partnership Project
  (3GPP)}, {Technical Report} {TR 38.901}, Jan. 2026, {Version 19.2.0, Release
  19}. [Online]. Available: \url{https://www.3gpp.org/dynareport/38901.htm}
\BIBentrySTDinterwordspacing

\bibitem{huangfu2022wair}
Y.~Huangfu \emph{et~al.}, ``{WAIR-D}: Wireless {AI} research dataset,''
  \emph{arXiv preprint arXiv:2212.02159}, 2022.

\bibitem{Sionna}
J.~Hoydis \emph{et~al.}, ``Sionna {RT}: Differentiable ray tracing for radio
  propagation modeling,'' in \emph{Proc. IEEE Globecom Workshops (GC Wkshps)},
  Dec. 2023, pp. 317--321.

\bibitem{luo2023sensiverse}
J.~Luo \emph{et~al.}, ``Sensiverse: A dataset for {ISAC} study,'' \emph{arXiv
  preprint arXiv:2308.13789}, 2023.

\bibitem{Barlacchi2015}
G.~Barlacchi \emph{et~al.}, ``A multi-source dataset of urban life in the city
  of {Milan} and the province of {Trentino},'' \emph{Sci. Data}, vol.~2, Oct.
  2015.

\bibitem{liu2021swin}
Z.~Liu \emph{et~al.}, ``{Swin Transformer}: Hierarchical vision {Transformer}
  using shifted windows,'' in \emph{Proc. IEEE/CVF Int. Conf. Comput. Vis.
  (ICCV)}, Oct. 2021, pp. 10\,012--10\,022.

\bibitem{liang2021swinir}
J.~Liang \emph{et~al.}, ``{SwinIR}: Image restoration using {Swin
  Transformer},'' in \emph{Proc. IEEE/CVF Int. Conf. Comput. Vis. Workshops
  (ICCVW)}, Oct. 2021, pp. 1833--1844.

\bibitem{zhu2021deep}
W.~Zhu \emph{et~al.}, ``Deep-learned approximate message passing for
  asynchronous massive connectivity,'' \emph{IEEE Trans. Wireless Commun.},
  vol.~20, no.~8, pp. 5434--5448, Aug. 2021.

\bibitem{9392296}
X.~Wang \emph{et~al.}, ``A survey on curriculum learning,'' \emph{IEEE Trans.
  Pattern Anal. Mach. Intell.}, vol.~44, no.~9, pp. 4555--4576, 2022.

\bibitem{9121328}
M.~Alrabeiah and A.~Alkhateeb, ``Deep learning for {mmWave} beam and blockage
  prediction using sub-6 {GHz} channels,'' \emph{IEEE Trans. Commun.}, vol.~68,
  no.~9, pp. 5504--5518, 2020.

\bibitem{3gpp38211}
\BIBentryALTinterwordspacing
3GPP, ``{NR}; physical channels and modulation,'' 3rd Generation Partnership
  Project (3GPP), TS 38.211 V15.4.0, Dec. 2018. [Online]. Available:
  \url{https://www.3gpp.org/ftp/Specs}
\BIBentrySTDinterwordspacing

\bibitem{jiang2022transformer}
H.~Jiang \emph{et~al.}, ``Accurate channel prediction based on {Transformer}:
  Making mobility negligible,'' \emph{IEEE J. Sel. Areas Commun.}, vol.~40,
  no.~9, pp. 2717--2732, Sep. 2022.

\bibitem{barbieri2009arma}
A.~Barbieri \emph{et~al.}, ``On the {ARMA} approximation for fading channels
  described by the {Clarke} model with applications to {Kalman}-based
  receivers,'' \emph{IEEE Trans. Wireless Commun.}, vol.~8, no.~2, pp.
  535--540, Feb. 2009.

\bibitem{edfors1998ofdm}
O.~Edfors \emph{et~al.}, ``{OFDM} channel estimation by singular value
  decomposition,'' \emph{IEEE Trans. Commun.}, vol.~46, no.~7, pp. 931--939,
  Jul. 1998.

\bibitem{sun2023channel}
Z.~Sun \emph{et~al.}, ``Channel state identification in complex indoor
  environments with {ST-CNN} and transfer learning,'' \emph{IEEE Commun.
  Lett.}, vol.~27, no.~2, pp. 546--550, 2023.

\bibitem{zhou2026lambda}
L.~Zhou \emph{et~al.}, ``{LAMBDA}: A low-altitude multimodal base dataset for
  {UAV} sensing and communication,'' \emph{arXiv preprint arXiv:2607.03826},
  2026.

\bibitem{yaman2024luvira}
I.~Yaman \emph{et~al.}, ``The {LuViRA} dataset: Synchronized vision, radio, and
  audio sensors for indoor localization,'' in \emph{Proc. IEEE Int. Conf.
  Robot. Autom. (ICRA)}, 2024, pp. 11\,920--11\,926.

\bibitem{li2026open}
J.~Li \emph{et~al.}, ``Open-set recognition of communication jamming using raw
  {I/Q} data with domain adaptation,'' \emph{IEEE Trans. Commun.}, vol.~74, pp.
  6508--6522, 2026.

\bibitem{jolliffe2016principal}
I.~T. Jolliffe and J.~Cadima, ``Principal component analysis: A review and
  recent developments,'' \emph{Philos. Trans. R. Soc. A: Math. Phys. Eng.
  Sci.}, vol. 374, no. 2065, 2016.

\end{thebibliography}
\end{document}